\documentclass{article}

\usepackage{microtype}
\usepackage{graphicx}
\usepackage{booktabs} 

\usepackage{algorithm}
\usepackage{algorithmic}
\usepackage{amsmath}
\usepackage{amssymb}
\usepackage{bm}

\usepackage{subcaption}
\usepackage{diagbox}
\usepackage{multirow}
\usepackage{multicol}
\usepackage{bm}

\usepackage{xcolor}
\usepackage{makecell}

\def\DAI{\texttt{DouZero}}

\DeclareMathOperator*{\argmax}{arg\,max}

\renewcommand\algorithmiccomment[1]{%
  \hfill$\triangleright$\ \emph{#1}%
}

\usepackage{hyperref}

\usepackage[accepted]{icml2021}

\icmltitlerunning{\DAI: Mastering DouDizhu with Self-Play Deep Reinforcement Learning}

\begin{document}

\twocolumn[
\icmltitle{\DAI: Mastering DouDizhu with Self-Play Deep Reinforcement Learning}




\begin{icmlauthorlist}
\icmlauthor{Daochen Zha}{to}
\icmlauthor{Jingru Xie}{goo}
\icmlauthor{Wenye Ma}{goo}
\icmlauthor{Sheng Zhang}{ed}
\icmlauthor{Xiangru Lian}{goo}
\icmlauthor{Xia Hu}{to}
\icmlauthor{Ji Liu}{goo}
\end{icmlauthorlist}

\icmlaffiliation{to}{Department of Computer Science and Engineering, Texas A\&M University}
\icmlaffiliation{goo}{AI Platform, Kwai Inc.}
\icmlaffiliation{ed}{Georgia Institute of Technology}

\icmlcorrespondingauthor{Daochen Zha}{daochen.zha@tamu.edu}

\icmlkeywords{Machine Learning, ICML}

\vskip 0.3in
]



\printAffiliationsAndNotice 

\begin{abstract}
Games are abstractions of the real world, where artificial agents learn to compete and cooperate with other agents. While significant achievements have been made in various perfect- and imperfect-information games, DouDizhu (a.k.a. Fighting the Landlord), a three-player card game, is still unsolved. DouDizhu is a very challenging domain with competition, collaboration, imperfect information, large state space, and particularly a massive set of possible actions where the legal actions vary significantly from turn to turn. Unfortunately, modern reinforcement learning algorithms mainly focus on simple and small action spaces, and not surprisingly, are shown not to make satisfactory progress in DouDizhu. In this work, we propose a conceptually simple yet effective DouDizhu AI system, namely $\DAI$, which enhances traditional Monte-Carlo methods with deep neural networks, action encoding, and parallel actors. Starting from scratch in a single server with four GPUs, $\DAI$ outperformed all the existing DouDizhu AI programs in days of training and was ranked the first in the Botzone leaderboard among 344 AI agents. Through building $\DAI$, we show that classic Monte-Carlo methods can be made to deliver strong results in a hard domain with a complex action space. The code and an online demo are released\footnote{{\url{https://github.com/kwai/DouZero}}} with the hope that this insight could motivate future work.
\end{abstract}

\section{Introduction}
Games often serve as benchmarks of AI since they are abstractions of many real-world problems. Significant achievements have been made in perfect-information games. For example, AlphaGo~\cite{silver2016mastering}, AlphaZero~\cite{silver2018general} and MuZero~\cite{schrittwieser2020mastering} have established state-of-the-art performance on Go game. Recent research has evolved to more challenging imperfect-information games, where the agents compete or cooperate with others in a partially observable environment. Encouraging progress has been made from two-player games, such as simple Leduc Hold'em and limit/no-limit Texas Hold’em~\cite{zinkevich2008regret,heinrich2016deep,moravvcik2017deepstack,brown2018superhuman}, to multi-player games, such as multi-player Texas hold’em~\cite{brown2019superhuman}, Starcraft~\cite{vinyals2019grandmaster}, DOTA~\cite{berner2019dota}, Hanabi~\cite{lerer2020improving}, Mahjong~\cite{li2020suphx}, Honor of Kings~\cite{ye2020mastering,ye2020towards}, and No-Press Diplomacy~\cite{gray2020human}.

This work aims at building AI programs for DouDizhu\footnote{\url{https://en.wikipedia.org/wiki/Dou_dizhu}} (a.k.a. Fighting the Landlord), the most popular card game in China with hundreds of millions of daily active players. DouDizhu has two interesting properties that pose great challenges for AI systems. First, the players in DouDizhu need to both compete and cooperate with others in a partially observable environment with limited communication. Specifically, two Peasants players will play as a team to fight against the Landlord player. Popular algorithms for poker games, such as Counterfactual Regret Minimization (CFR)~\cite{zinkevich2008regret}) and its variants, are often not sound in this complex three-player setting. Second, DouDizhu has a large number of information sets with a very large average size and has a very complex and large action space of up to $10^4$ possible actions due to combinations of cards~\cite{zha2019rlcard}. Unlike Texas Hold'em, the actions in DouDizhu can not be easily abstracted, which makes search computationally expensive and commonly used reinforcement learning algorithms less effective. Deep Q-Learning~(DQN)~\cite{mnih2015human} is problematic in very large action space due to overestimating issue~\cite{zahavy2018learn}; policy gradient methods, such as A3C~\cite{mnih2016asynchronous}, cannot leverage the action features in DouDizhu, and thus cannot generalize over unseen actions as naturally as DQN~\cite{dulac2015deep}. Not surprisingly, previous work shows that DQN and A3C can not make satisfactory progress in DouDizhu. In \cite{you2019combinational}, DQN and A3C are shown to have less than 20\% winning percentage against simple rule-based agents even with twenty days of training; the DQN in \cite{zha2019rlcard} is only slightly better than random agents that sample legal moves uniformly.

Some previous efforts have been made to build DouDizhu AI by combining human heuristics with learning and search. Combination Q-Network~(CQN)~\cite{you2019combinational} proposes to reduce the action space by decoupling the actions into decomposition selection and final move selection. However, decomposition relies on human heuristics and is extremely slow. In practice, CQN can not even beat simple heuristic rules after twenty days of training. DeltaDou~\cite{jiang2019deltadou} is the first AI program that reaches human-level performance compared with top human players. It enables an AlphaZero-like algorithm by using Bayesian methods to infer hidden information and sampling the other players' actions based on their own policy networks. To abstract the action space, DeltaDou pre-trains a kicker network based on heuristic rules. However, the kicker plays an important role in DouDizhu and can not be easily abstracted. A bad selection of the kicker may directly result in losing a game since it may break some other card categories, e.g., a Chain of Solo. Moreover, the Bayesian inference and the search are computationally expensive. It takes more than two months to train DeltaDou even when initializing the networks with supervised regression to heuristics~\cite{jiang2019deltadou}. Therefore, the existing DouDizhu AI programs are computationally expensive and could be sub-optimal since they highly rely on abstractions with human knowledge.


In this work, we present $\DAI$, a conceptually simple yet effective AI system for DouDizhu without the abstraction of the state/action space or any human knowledge. $\DAI$ enhances traditional Monte-Carlo methods~\cite{sutton2018reinforcement} with deep neural networks, action encoding, and parallel actors. $\DAI$ has two desirable properties. First, unlike DQN, it is not susceptible to overestimation bias. Second, by encoding the actions into card matrices, it can naturally generalize over the actions that are not frequently seen throughout the training process. Both of these two properties are crucial in dealing with the huge and complex action space of DouDizhu. Unlike many tree search algorithms, $\DAI$ is based on sampling, which allows us to use complex neural architectures and generate much more data per second, given the same computational resources. Unlike many prior poker AI studies that rely on domain-specific abstractions, $\DAI$ does not require any domain knowledge or knowledge of the underlying dynamics. Trained from scratch in a single server with only 48 cores and four 1080Ti GPUs, $\DAI$ outperforms CQN and the heuristic rules in half a day, beats our internal supervised agents in two days, and surpasses DeltaDou in ten days. Extensive evaluations suggest that $\DAI$ is the strongest DouDizhu AI system up to date.

Through building $\DAI$ system, we demonstrate that classical Monte-Carlo methods can be made to deliver strong results in large-scale and complex card games that need to reason about both competing and cooperation over huge state and action spaces. We note that some work also discovers that Monte-Carlo methods can achieve competitive performance~\cite{mania2018simple,zha2021simplifying} and help in sparse rewards settings~\cite{guo2018generative,zha2021rank}. Unlike these studies that focus on simple and small environments, we demonstrate the strong performance of Monte-Carlo methods on a large-scale card game. With the hope that this insight could facilitate future research on tackling multi-agent learning, sparse reward, complex action spaces, and imperfect information, we have released our environment and the training code. Unlike many Poker AI systems that require thousands of CPUs in training, e.g., DeepStack~\cite{moravvcik2017deepstack} and Libratus~\cite{brown2018superhuman}, $\DAI$ enables a reasonable experimental pipeline, which only requires days of training on a single GPU server that is affordable for most research labs. We hope that it could motivate future research in this domain and serve as a strong baseline.

\vspace{-5pt}
\section{Background of DouDizhu}
\vspace{-5pt}

DouDizhu is a popular three-player card game that is easy to learn but difficult to master. It has attracted hundreds of millions of players in China, with many tournaments held every year. It is a shedding-type game where the player's objective is to empty one's hand of all cards before other players. Two of the Peasants players play as a team to fight against the other Landlord player. The Peasants win if either of the Peasants players is the first to have no cards left. Each game has a bidding phase, where the players bid for the Landlord based on the strengths of the hand cards, and a card-playing phase, where the players play cards in turn. We provide a detailed introduction in Appendix~\ref{sec:A}.

\begin{figure}[t]
  \centering
    \includegraphics[width=0.46\textwidth]{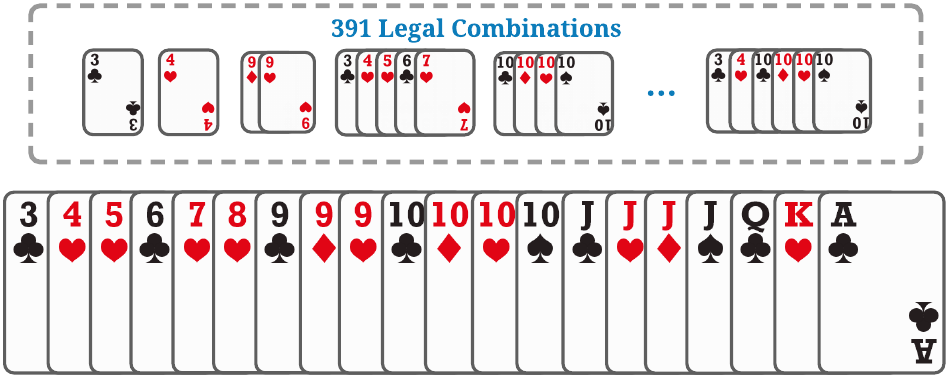}
\vspace{-10pt}
  \caption{A hand and its corresponding legal moves.}
  \label{fig:examplehand}
  \vspace{-15pt}
\end{figure}

DouDizhu is still an unsolved benchmark for multi-agent reinforcement learning~\cite{zha2019rlcard,terry2020pettingzoo}. Two interesting properties make DouDizhu particularly challenging to solve. First, the Peasants need to cooperate in fighting against the Landlord. For example, Figure~\ref{fig:casestudymain} shows a typical situation where the bottom Peasant can choose to play a small Solo to help the Peasant on the right-hand side to win. Second, DouDizhu has a complex and large action space due to the combination of cards. There are $27,472$ possible combinations, where different subsets of these combinations will be legal for different hands. Figure~\ref{fig:examplehand} shows an example of the hand, which has $391$ legal combinations, including Solo, Pair, Trio, Bomb, Plane, Quad, etc. The action space can not be easily abstracted since improperly playing a card may break other categories and directly result in losing a game. Thus, building DouDizhu AI is challenging since the players in DouDizhu need to reason about both competing and cooperation over a huge action space.

\vspace{-5pt}
\section{Deep Monte-Carlo}
In this section, we revisit Monte-Carlo (MC) methods and introduce Deep Monte-Carlo~(DMC), which generalizes MC with deep neural networks for function approximation. Then we discuss and compare DMC with policy gradient methods (e.g., A3C) and DQN, which are shown to fail in DouDizhu~\cite{you2019combinational,zha2019rlcard}.

\begin{figure}[t]
  \centering
    \includegraphics[width=0.45\textwidth]{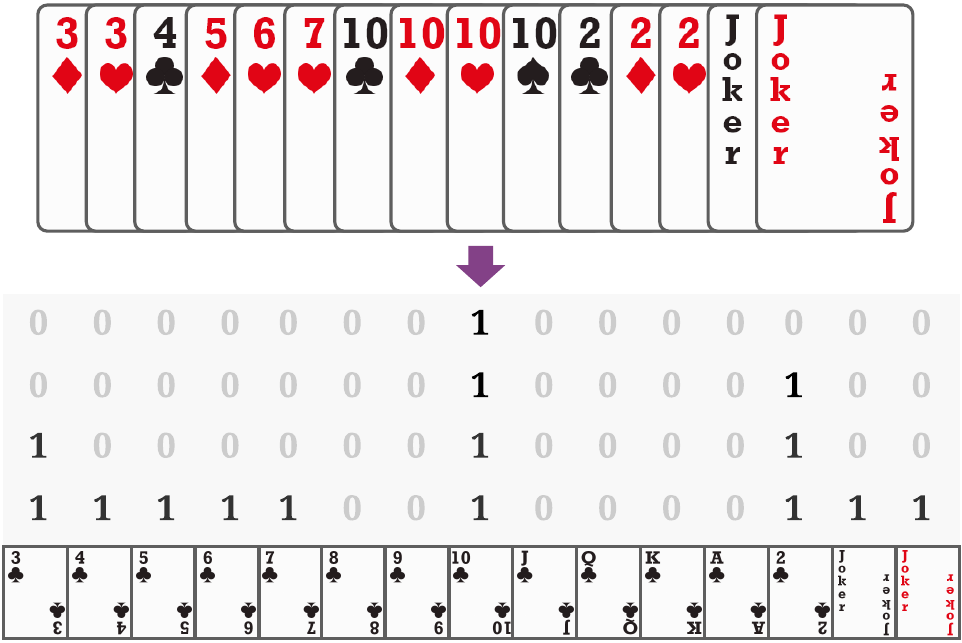}
    \vspace{-12pt}
  \caption{Cards for both states and actions are encoded into a $4 \times 15$ one-hot matrix, where columns correspond to the 13 ranks and the jokers, and each row corresponds to the number of cards of a specific rank or joker. More examples are provided in Appendix~\ref{sec:B}.}
  \label{fig:cardrepresentation}
  \vspace{-10pt}
\end{figure}

\vspace{-5pt}
\subsection{Monte-Carlo Methods with Deep Neural Networks}
Monte-Carlo (MC) methods are traditional reinforcement learning algorithms based on averaging sample returns~\cite{sutton2018reinforcement}. MC methods are designed for episodic tasks, where experiences can be divided into episodes and all the episodes eventually terminate. To optimize a policy $\pi$, every-visit MC can be used to estimate Q-table $Q(s,a)$ by iteratively executing the following procedure:
\vspace{-7pt}
\begin{enumerate}
    \item Generate an episode using $\pi$.
    \vspace{-7pt}
    \item For each $s, a$ appeared in the episode, calculate and update  $Q(s,a)$ with the return averaged over all the samples concerning $s, a$.
    \vspace{-7pt}
    \item For each $s$ in the episode, $\pi(s) \leftarrow \argmax_a Q(s,a)$.
    \vspace{-7pt}
\end{enumerate}
The average return in Step 2 is usually obtained by the discounted cumulative reward. Different from Q-learning that relies on bootstrapping, MC methods directly approximate the target Q-value. In step 1, we can use epsilon-greedy to balance exploration and exploitation. The above procedure can be naturally combined with deep neural networks, which leads to Deep Monte-Carlo~(DMC). Specifically, we can replace the Q-table with a neural network and use mean-square-error (MSE) to update the Q-network in Step 2.

While MC methods are criticized not to be able to deal with incomplete episodes and believed to be inefficient due to the high variance~\cite{sutton2018reinforcement}, DMC is very suitable for DouDizhu. First, DouDizhu is an episodic task so that we do not need to handle incomplete episodes. Second, DMC can be easily parallelized to efficiently generate many samples per second to alleviate the high variance issue.


\begin{figure}[t]
  \centering
    \includegraphics[width=0.47\textwidth]{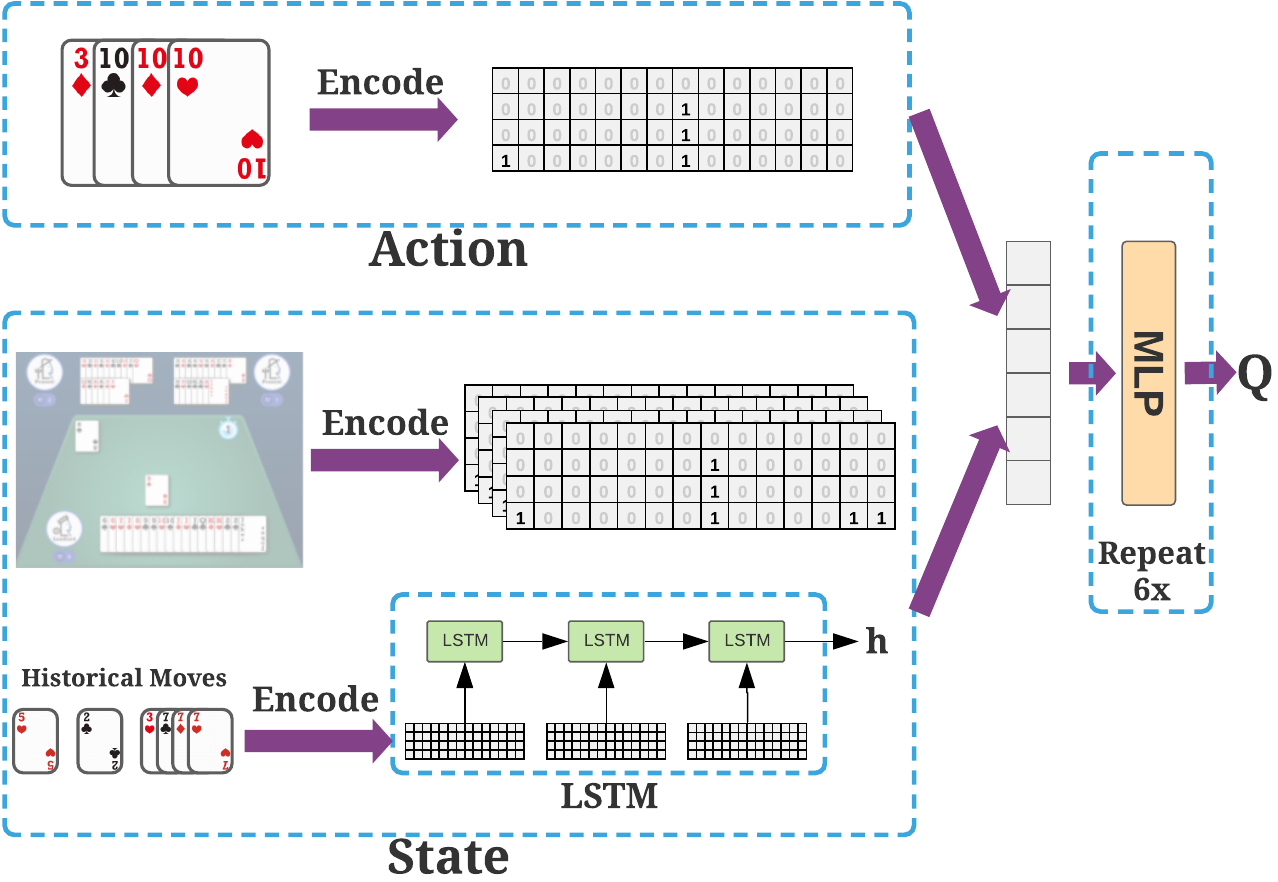}
  \caption{The Q-network of $\DAI$ consists of an LSTM to encode historical moves and six layers of MLP with hidden dimension of 512. The network predicts a value for a given state-action pair based on the concatenated representation of action and state. More details are provided in Appendix~\ref{sec:C1}.}
  \label{fig:overview}
  \vspace{-9pt}
\end{figure}

\vspace{-3pt}
\subsection{Comparison with Policy Gradient Methods}
Policy gradients methods, such as REINFORCE~\cite{williams1992simple}, A3C~\cite{mnih2016asynchronous}, PPO~\cite{schulman2017proximal}, and IMPALA~\cite{espeholt2018impala}, are very popular for reinforcement learning. They target modeling and optimizing the policy directly with gradient descent. In policy gradient methods, we often use a classifier-like function approximator, where the output scales linearly with the number of actions. While policy gradients methods work well in large action space, they cannot use the action features
to reason about previously unseen actions~\cite{dulac2015deep}. In practice, the actions in DouDizhu can be naturally encoded into card matrices, which are crucial for reasoning. For example, if the agent is rewarded by the action \texttt{3KKK} because it chooses a nice kicker, it could also generalize this knowledge to unseen actions in the future, such as \texttt{3JJJ}. This property is crucial in dealing with very large action spaces and accelerating the learning since many of the actions are not frequently seen in the simulated data.

DMC can naturally leverage the action features to generalize over unseen actions by taking as input the action features. While it might have high execution complexity if the action size is large, in most states of DouDizhu, only a subset of the actions is legal, so that we do not need to iterate over all the actions. Thus, DMC is overall an efficient algorithm for DouDizhu. While it is possible to introduce action features into an actor-critic framework (e.g., by using a Q-network as the critic), the classifier-like actor will still suffer from the large action space. Our preliminary experiments confirm that this strategy is not very effective (see Figure~\ref{fig:sarsaac}).

\subsection{Comparison with Deep Q-Learning}
The most popular value-based algorithm is Deep Q-Learning~(DQN)~\cite{mnih2015human}, which is a bootstrapping method that updates the Q-value based on the Q-values in the next step. While both DMC and DQN approximate the Q-values, DMC has several advantages in DouDizhu.

First, the overestimation bias caused by approximating the maximum action value in DQN is difficult to control when using function approximation ~\cite{thrun1993issues,hasselt2010double} and becomes more pronounced with very large action space~\cite{zahavy2018learn}. While some techniques, such as double Q-learning~\cite{van2016deep} and experience replay~\cite{lin1992self}, might alleviate this issue, we find in practice that DQN is very unstable and often diverges in DouDizhu. Whereas, Monte-Carlo estimation is not susceptible to bias since it directly approximates the true values without bootstrapping~\cite{sutton2018reinforcement}.

Second, DouDizhu is a task with long horizons and sparse reward, i.e., the agent will need to go though a long chain of states without feedback, and the only time a nonzero reward is incurred is at the end of a game. This may slow down the convergence of Q-learning because estimating the Q-value in the current state needs to wait until the value in the next state gets close to its true value~\cite{szepesvari2009algorithms,beleznay1999comparing}. Unlike DQN, the convergence of Monte-Carlo estimation is not impacted by the episode length since it directly approximates the true target values.

Third, it is inconvenient to efficiently implement DQN in DouDizhu due to the large and variable action space. Specifically, the max operation of DQN in every update step will cause high computation cost since it requires iterating across all the legal actions on a very costly deep Q-network. Moreover, the legal moves differ in different states, which makes it inconvenient to do batch learning. As a result, we find DQN is too slow in terms of wall-clock time. While Monte-Carlo methods might suffer from high variance~\cite{sutton2018reinforcement}, which means it might require more samples to converge, it can be easily parallelized to generate thousands of samples per second to alleviate the high variance issue and accelerate training. We find that the high variance of DMC is greatly outweighed by the scalability it provides, and DMC is very efficient in wall-clock time.

\section{$\DAI$ System}
\vspace{-3pt}
In this section, we introduce $\DAI$ system by first describing the state/action representations and neural architecture and then elaborating on how we parallelize DMC with multiple processes to stabilize and accelerate training.

\vspace{-3pt}
\subsection{Card Representation and Neural Architecture}
We encode each card combination with a one-hot $4 \times 15$ matrix (Figure~\ref{fig:cardrepresentation}). Since suits are irrelevant in DouDizhu, we use each row to represent the number of cards of a specific rank or joker. Figure~\ref{fig:overview} shows the architecture of the Q-network. For the state, we extract several card matrices to represent the hand cards, the union of the other players’ hand cards and the most recent moves, and some one-hot vectors to represent the number of cards of the other players and the number of bombs played so far. Similarly, we use one card matrix to encode the action. For the neural architecture, LSTM is used to encode historical moves, and the output is concatenated with the other state/action features. Finally, we use six layers of MLP with a hidden size of 512 to produce Q-values. We provide more details in Appendix~\ref{sec:C1}.

\vspace{-3pt}
\subsection{Parallel Actors}
We denote Landlord as L, the player that moves before the Landlord as U, and the player that moves after the Landlord as D. We parallelize DMC with multiple actor processes and one learner process, summarized in Algorithm~\ref{alg:1} and Algorithm~\ref{alg:2}, respectively. The learner maintains three global Q-networks for the three positions and updates the networks with MSE loss to approximate the target values based on the data provided by the actor processes. Each actor maintains three local Q-networks, which are synchronized with the global networks periodically. The actor will repeatedly sample trajectories from the game engine and calculate cumulative reward for each state-action pair. The communication of learner and actors are implemented with three shared buffers. Each buffer is divided into several entries, where each entry consists of several data instances.

\begin{algorithm}[t]
\caption{Actor Process of $\DAI$}
\label{alg:1}
\setlength{\arraycolsep}{2pt}
\setlength{\intextsep}{0pt} 
\begin{algorithmic}[1]
\STATE \textbf{Input:} Shared buffers $\mathcal{B}_L$, $\mathcal{B}_U$ and $\mathcal{B}_D$ with $B$ entries and size $S$ for each entry, exploration hyperparameter $\epsilon$, discount factor $\gamma$
\STATE Initialize local Q-networks $Q_\text{L}$, $Q_\text{U}$ and $Q_\text{D}$, and local buffers $\mathcal{D}_L$, $\mathcal{D}_U$ and $\mathcal{D}_D$
\FOR{iteration = $1$, $2$, ...}
    \STATE Synchronize $Q_\text{L}$, $Q_\text{U}$ and $Q_\text{D}$ with the learner process
    \FOR[Generate an episode]{t = $1$, $2$, ... T} 
        \STATE $Q \leftarrow \text{one of } Q_\text{L}, Q_\text{U}$, $Q_\text{D}$ based on position
        \STATE  $a_t \leftarrow \left\{
\begin{array}{rcl}
\argmax_a Q(s_t, a)       &      & \text{with prob }(1-\epsilon)\\
\text{random action}    &      & \text{with prob }\epsilon
\end{array} \right.$
        \STATE Perform $a_t$, observe $s_{t+1}$ and reward $r_t$
        \STATE Store $\{s_t, a_t, r_t\}$ to $\mathcal{D}_L$, $\mathcal{D}_U$, or $\mathcal{D}_D$ accordingly
    \ENDFOR
    \FOR[Obtain cumulative reward]{t = T-1, T-2, ... 1}
        \STATE $r_t \leftarrow r_t + \gamma r_{t+1}$ and update $r_t$ in $\mathcal{D}_L$, $\mathcal{D}_U$, or $\mathcal{D}_D$
    \ENDFOR
    \FOR[Optimized by multi-thread]{$p \in \{L, U, D\}$}
    \IF{$\mathcal{D}_p.\text{length} \ge L$}
        \STATE Request and wait for an empty entry in $\mathcal{B}_p$
        \STATE Move $\{s_t, a_t, r_t\}$ of size $L$ from $\mathcal{D}_p$ to $\mathcal{B}_p$
    \ENDIF
    
    \ENDFOR
\ENDFOR
\end{algorithmic}
\end{algorithm}

\begin{algorithm}[t]
\caption{Learner Process of $\DAI$}
\label{alg:2}
\setlength{\intextsep}{0pt} 
\begin{algorithmic}[1]
\STATE \textbf{Input:} Shared buffers $\mathcal{B}_L$, $\mathcal{B}_U$ and $\mathcal{B}_D$ with $B$ entries and size $S$ for each entry, batch size $M$, learning rate $\psi$
\STATE Initialize global Q-networks $Q^g_\text{L}$, $Q^g_\text{U}$ and $Q^g_\text{D}$
\FOR{iteration = $1$, $2$, ... until convergence}
    \FOR[Optimized by multi-thread]{$p \in \{L, U, D\}$}
    \IF{the number of full entries in $\mathcal{B}_p \ge M$}
        \STATE Sample a batch of $\{s_t, a_t, r_t\}$ with $M \times S$ instances from $\mathcal{B}_p$ and free the entries
        \STATE Update $Q^g_\text{p}$ with MSE loss and learning rate $\psi$
    \ENDIF
    
    \ENDFOR
\ENDFOR
\end{algorithmic}
\end{algorithm}


\section{Experiments}
The experiments are designed to answer the following research questions. \textbf{RQ1:} How does $\DAI$ compare with existing DouDizhu programs, such as rule-based strategies, supervised learning, RL-based methods, and MCTS-based solutions~(Section~\ref{sec:2})? \textbf{RQ2:} How will $\DAI$ perform if we consider bidding phase~(Section~\ref{sec:3})? \textbf{RQ3:} How efficient is the training of $\DAI$~(Section~\ref{sec:4})? \textbf{RQ4:} How does $\DAI$ compare with bootstrapping and actor critic methods~(Section~\ref{sec:5})? \textbf{RQ5:} Does the learned card playing strategies of $\DAI$ align with human knowledge~(Section~\ref{sec:6})? \textbf{RQ6:} Is $\DAI$ computationally efficient in inference compared with existing programs~(Section~\ref{sec:7})?  \textbf{RQ7:} Can the two Peasants of $\DAI$ learn to cooperate with each other~(Section~\ref{sec:8})?


\subsection{Experimental Setup}
\label{sec:1}

A commonly used measure of strategy strength in poker games is exploitability~\cite{johanson2011accelerating}. However, in DouDizhu, calculating exploitability itself is intractable since DouDizhu has huge state/action spaces, and there are three players. To evaluate the performance, following~\cite{jiang2019deltadou}, we launch tournaments that include the two opponent sides of Landlord and Peasants. We reduce the variance by playing each deck twice. Specifically, for two competing algorithms \texttt{A} and \texttt{B}, they will first play as Landlord and Peasants positions, respectively, for a given deck. Then they switch sides, i.e., \texttt{A} takes Peasants position, and \texttt{B} takes Landlord position, and play the same deck again. To simulate the real environment, in Section~\ref{sec:3}, we further train a bidding network with supervised learning, and the agents will bid the Landlord in each game based on the strengths of the hand cards~(more details in Appendix~\ref{sec:C2}). We consider the following competing algorithms.

\begin{itemize}
    \item \textbf{DeltaDou:} A strong AI program which uses Bayesian methods to infer hidden information and searches the moves with MCTS~\cite{jiang2019deltadou}. We use the code and the pre-trained model provided by the authors. The model is trained for two months and is shown to have on par performance with top human players.
    \item \textbf{CQN:} Combinational Q-Learning~\cite{you2019combinational} is a program based on card decomposition and Deep Q-Learning. We use the open-sourced code and the pre-trained model provided by the authors\footnote{\url{https://github.com/qq456cvb/doudizhu-C}}.
    \item \textbf{SL:} A supervised learning baseline. We internally collect $226,230$ human expert matches from the players of the highest level in league in our DouDizhu game mobile app. Then we use the same state representation and neural architecture as $\DAI$ to train supervised agents with $49,990,075$ samples generated from these data. See Appendix~\ref{sec:C2} for more details.
    \item \textbf{Rule-Based Programs:} We collect some open-sourced heuristic-based programs, including \textbf{RHCP}\footnote{\url{https://blog.csdn.net/sm9sun/article/details/70787814}}, an improved version called \textbf{RHCP-v2}\footnote{\url{https://github.com/deecamp2019-group20/RuleBasedModelV2}}, and the rule model in \textbf{RLCard} package\footnote{\url{https://github.com/datamllab/rlcard}}~\cite{zha2019rlcard}. In addition, we consider a \textbf{Random} program that samples legal moves uniformly.
\end{itemize}

\begin{table*}[t]
    \centering
    \small
    \caption{Performance of $\DAI$ against existing DouDizhu programs by playing 10,000 randomly sampled decks. Algorithm \texttt{A} outperforms \texttt{B} if WP is larger than 0.5 or ADP is larger than 0 (highlighted in boldface). The algorithms are ranked according to the number of the other algorithms that they beat. The full results of each position are provided in Appendix~\ref{sec:D1}.}.
    \vspace{-7pt}
    \label{tab:mainperformance}
    \setlength{\tabcolsep}{2.0pt}
    \begin{tabular}{c||l|cc|cc|cc|cc|cc|cc|cc|cc}
    \toprule
     
\multirow{2}{*}{Rank} & \multirow{2}{*}{\diagbox [width=5em,trim=l] {\texttt{A}}{\texttt{B}}} & \multicolumn{2}{c|}{$\DAI$} & \multicolumn{2}{c|}{DeltaDou} & \multicolumn{2}{c|}{SL} & \multicolumn{2}{c|}{RHCP-v2} & \multicolumn{2}{c|}{RHCP} & \multicolumn{2}{c|}{RLCard} & \multicolumn{2}{c|}{CQN} &\multicolumn{2}{c}{Random}\\
\cline{3-18}
& & WP & ADP & WP & ADP & WP & ADP & WP & ADP & WP & ADP & WP & ADP & WP & ADP & WP & ADP\\
    \hline
    \midrule
     1 & $\DAI$ & - & - & \textbf{0.586} & \textbf{0.258} & \textbf{0.659} & \textbf{0.700} & \textbf{0.757} & \textbf{1.662} & \textbf{0.764} & \textbf{1.671} & \textbf{0.889} & \textbf{2.288} & \textbf{0.810} & \textbf{1.685} & \textbf{0.989} & \textbf{3.036}\\
     2 & DeltaDou & 0.414 & -0.258 & - & - & \textbf{0.617} & \textbf{0.653} & \textbf{0.745} & \textbf{1.500} & \textbf{0.747} & \textbf{1.514} & \textbf{0.876} & \textbf{2.459} & \textbf{0.784} & \textbf{1.534} & \textbf{0.992} & \textbf{3.099}\\
     3 & SL & 0.341 & -0.700 & 0.396 & -0.653 & - & - & \textbf{0.611} & \textbf{0.853} & \textbf{0.632} & \textbf{0.886} & \textbf{0.813} & \textbf{1.821} & \textbf{0.694} & \textbf{1.037} & \textbf{0.976} & \textbf{2.721}\\
     4 & RHCP-v2 & 0.243 & -1.662 & 0.257 & -1.500 & 0.389 & -0.853 & - & - & \textbf{0.515} & \textbf{0.052} & \textbf{0.692} & \textbf{1.121} & \textbf{0.621} & \textbf{0.714} & \textbf{0.967} & \textbf{2.631}\\
     5 & RHCP & 0.236 & -1.671 & 0.253 & -1.514 & 0.369 & -0.886 & 0.485 & -0.052 & - & - & \textbf{0.682} & \textbf{1.259} & \textbf{0.603} & \textbf{0.248} & \textbf{0.941} & \textbf{2.720}\\
     6 & RLCard & 0.111 & -2.288 & 0.124 & -2.459 & 0.187 & -1.821 & 0.309 & -1.121 & 0.318 & -1.259 & - & - & \textbf{0.522} & \textbf{0.168} & \textbf{0.943} & \textbf{2.471}\\
     7 & CQN & 0.190 & -1.685 & 0.216 & -1.534 & 0.306 & -1.037 & 0.379 & -0.714 & 0.397 & -0.248 & 0.478 & -0.168 & - & - & \textbf{0.889} & \textbf{1.912}\\
     8 & Random & 0.011 & -3.036 & 0.008 & -3.099 & 0.024 & -2.721 & 0.033 & -2.631 & 0.059 & -2.720 & 0.057 & -2.471 & 0.111 & -1.912 & - & -\\
   
     \bottomrule
    \end{tabular}
    \vspace{-7pt}
\end{table*}

\textbf{Metrics.} Following~\cite{jiang2019deltadou}, given an algorithm \texttt{A} and an opponent \texttt{B}, we use two metrics to compare the performance of \texttt{A} and \texttt{B}:
\begin{itemize}
    \item \textbf{WP}~(Winning Percentage): The number of the games won by \texttt{A} divided by the total number of games.
    \item \textbf{ADP}~(Average Difference in Points): The average difference of points scored per game between \texttt{A} and \texttt{B}. The base point is 1. Each bomb will double the score.
\end{itemize}
We find in practice that these two metrics encourage different styles of strategies. For example, if using ADP as reward, the agent tends to be very cautious about playing bombs since playing a bomb is risky and may lead to larger ADP loss. In contrast, with WP as objective, the agent tends to aggressively play bombs even if it will lose because a bomb will not affect WP. We observe that the agent trained with ADP performs slightly better than the agent trained with WP in terms of ADP and vice versa. In what follows, we train and report the results of two $\DAI$ agents with ADP and WP as objectives, respectively\footnote{For WP, we give a +1 or -1 reward to the final timestep based on whether the agent wins or loses a game. For ADP, we directly use ADP as the rewards. DeltaDou and CQN were trained with ADP and WP as objectives, respectively.}. More discussions of the two objectives are provided in Appendix~\ref{sec:D2}.

We first launch a preliminary tournament by letting each pair of the algorithms play 10,000 decks. We then compute the Elo rating score for the top 3 algorithms for a more reliable comparison, i.e., $\DAI$, DeltaDou, and SL, by playing 100,000 decks. An algorithm wins a deck if it achieves higher WP or ADP summed over the two games played on this deck. We repeat this process five times with different randomly sampled decks and report the mean and standard deviation of the Elo scores. For the evaluation with the bidding phase, each deck is played six times with different perturbations of $\DAI$, DeltaDou, and SL in different positions. We report the result with 100,000 decks.

\textbf{Implementation Details.} We run all the experiments on a single server with 48 processors of Intel(R) Xeon(R) Silver 4214R CPU @ 2.40GHz and four 1080 Ti GPUs. We use 45 actors, which are allocated across three GPUs. We run a learner in the remaining GPU to train the Q-networks. Our implementation is based on TorchBeast framework~\cite{kuttler2019torchbeast}. The detailed training curves are provided in Appendix~\ref{sec:D5}. Each shared buffer has $B=50$ entries with size $S=100$, batch size $M=32$, and $\epsilon = 0.01$. We set discount factor $\gamma=1$ since DouDizhu only has a non-zero reward in the last timestep and early moves are very important. We use ReLU as the activation function for each layer of MLP. We adopt RMSprop optimizer with a learning rate $\psi=0.0001$, smoothing constant $0.99$ and $\epsilon=10^{-5}$. We train $\DAI$ for $30$ days.

\begin{figure}[t]
  \centering
  \begin{subfigure}[b]{0.230\textwidth}
    \centering
    \includegraphics[width=1.00\textwidth]{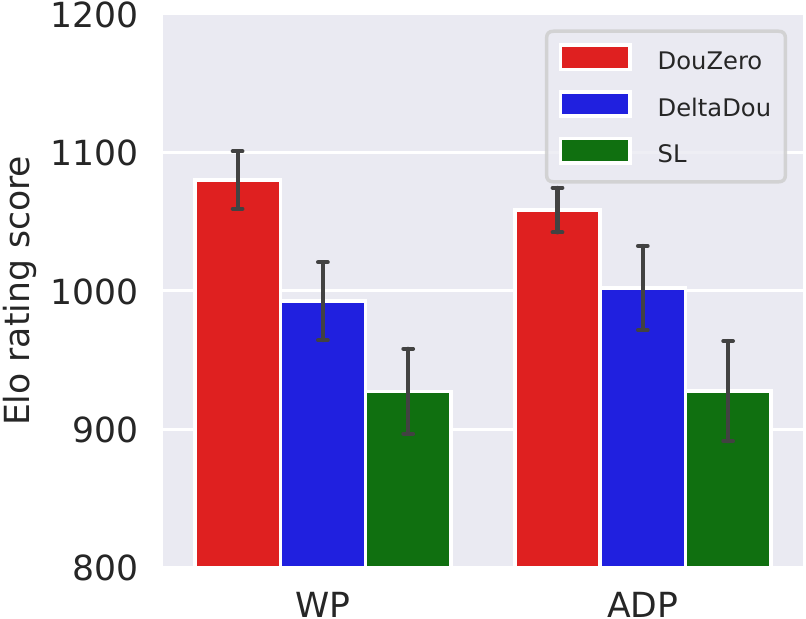}
  \end{subfigure}%
  \begin{subfigure}[b]{0.240\textwidth}
    \centering
    \includegraphics[width=1.00\textwidth]{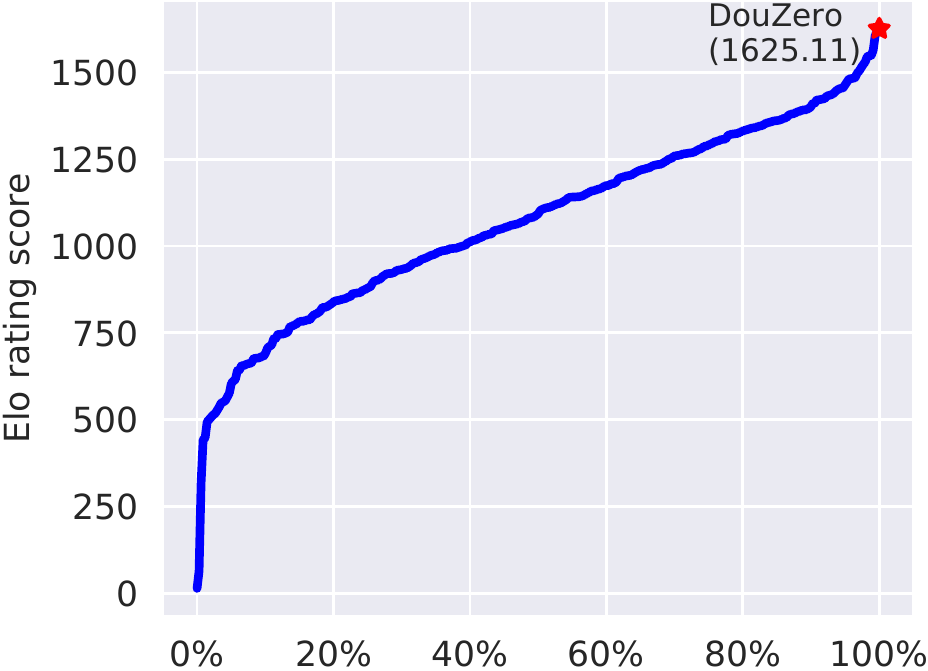}
  \end{subfigure}%
  \vspace{-7pt}
  \caption{\textbf{Left:} Elo rating scores of $\DAI$, DeltaDou, and SL by playing 100,000 randomly sampled decks. We report the mean and standard deviation across $5$ different random seeds. \textbf{Right:} Elo rating scores on Botzone, an online platform for DouDizhu competition. $\DAI$ ranked the first among the 344 bots, achieving an Elo rating score of 1625.11 as of October 30, 2020.}
  \label{fig:eloscores}
  \vspace{-7pt}
\end{figure}

\begin{figure*}[t]
  \centering
  \begin{subfigure}[b]{0.40\textwidth}
    \includegraphics[width=1.0\textwidth]{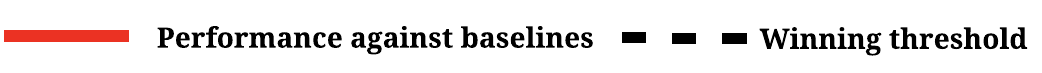}
  \end{subfigure}

  \begin{subfigure}[b]{0.25\textwidth}
    \centering
    \includegraphics[width=0.95\textwidth]{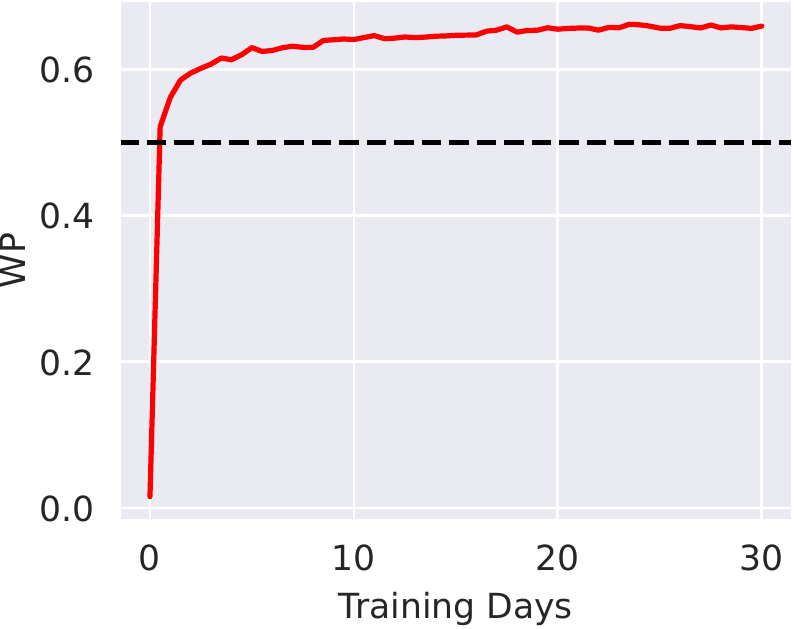}
    \caption{WP against SL}
  \end{subfigure}%
  \begin{subfigure}[b]{0.25\textwidth}
    \centering
    \includegraphics[width=0.95\textwidth]{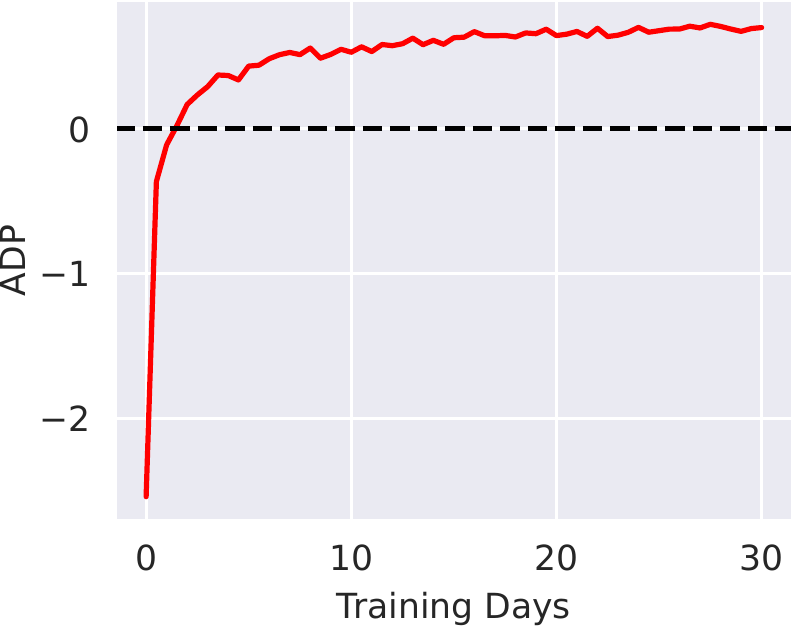}
    \caption{ADP against SL}
  \end{subfigure}%
  \begin{subfigure}[b]{0.25\textwidth}
    \centering
    \includegraphics[width=0.95\textwidth]{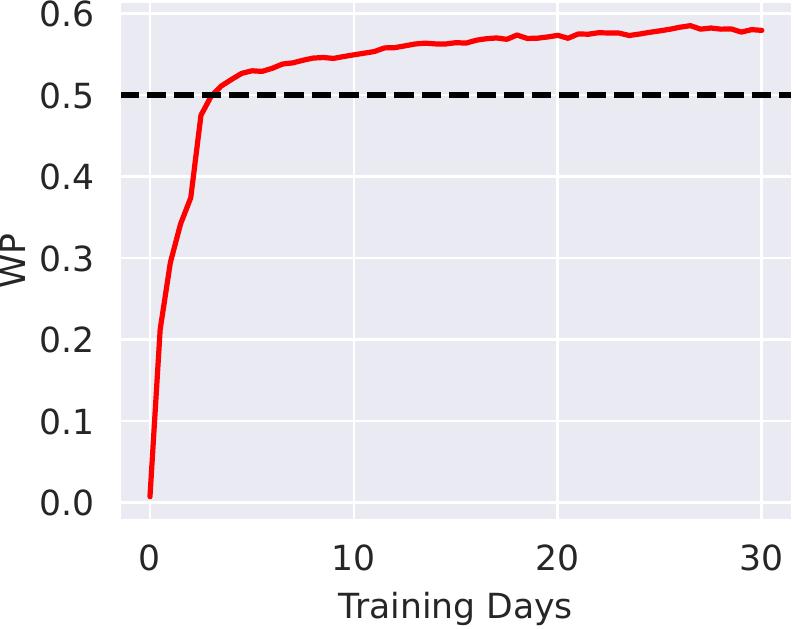}
    \caption{WP against DeltaDou}
  \end{subfigure}%
  \begin{subfigure}[b]{0.25\textwidth}
    \centering
    \includegraphics[width=0.95\textwidth]{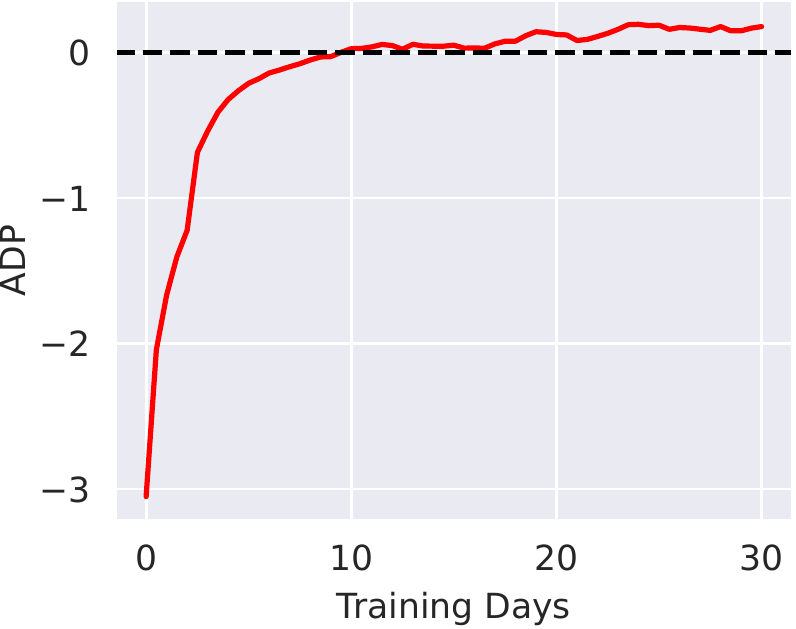}
    \caption{ADP against DeltaDou}
  \end{subfigure}%
  \vspace{-8pt}
  \caption{WP and ADP of $\DAI$ against SL and DeltaDou w.r.t. the number of training days. $\DAI$ outperforms SL with 2 days of training, i.e., the overall WP is larger than the threshold of 0.5 and the overall ADP is larger than the threshold of 0, and surpasses DeltaDou within 10 days, using a single server with four 1080 Ti GPUs and 48 processors. We provide the full curves for each position and the curves w.r.t. timesteps in Appendix~\ref{sec:D3}.}
  \vspace{-15pt}
  \label{fig:learningprogress}
\end{figure*}

\subsection{Performance against Existing Programs}
\label{sec:2}
To answer \textbf{RQ1}, we compare $\DAI$ with the baselines offline and report its result on Botzone~\cite{zhou2018botzone}, an online platform for DouDizhu competition (more details are provided in Appendix~\ref{sec:E}).

Table~\ref{tab:mainperformance} summarizes the WP and ADP of head-to-head completions among $\DAI$ and all the baselines. We make three observationss. First, $\DAI$ dominates all the rule-based strategies and supervised learning, which demonstrates the effectiveness of adopting reinforcement learning in DouDizhu. Second, $\DAI$ achieves significantly better performance than CQN. Recall that CQN similarly trains the Q-networks with action decomposition and DQN. The superiority of $\DAI$ suggests that DMC is indeed an effective way to train the Q-networks in DouDizhu. Third, $\DAI$ outperforms DeltaDou, the strongest DouDizhu AI in the literature. We note that DouDizhu has very high variance, i.e., to win a game relies on the strength of the initial hand card, which is highly dependent on luck. Thus, a WP of $0.586$ and an ADP of $0.258$ suggest a significant improvement over DeltaDou. Moreover, DeltaDou requires searching at both training and testing time. Whereas, $\DAI$ does not do the searching, which verifies that the Q-networks learned by $\DAI$ are very strong.

The left-hand side of Figure~\ref{fig:eloscores} shows the Elo rating scores of $\DAI$, DeltaDou, and SL by playing $100,000$ decks. We observe that $\DAI$ outperforms DeltaDou and SL in terms of both WP and ADP significantly. This again demonstrates the strong performance of $\DAI$.

The right-hand side of Figure~\ref{fig:eloscores} illustrates the performance of $\DAI$ on Botzone leaderboard. We note that Botzone adopts a different scoring mechanism. In addition to WP, it gives additional bonuses to some specific card categories, such as Chain of Pair and Rocket~(detailed in Appendix~\ref{sec:E}). While it is very likely that $\DAI$ can achieve better performance if using the scoring mechanism of Botzone as the objective, we directly upload the pre-trained model of $\DAI$ that is trained with WP as objective. We observe that this model is strong enough to beat the other bots.

\begin{table}[t]
    \centering
    \small
    \caption{Comparison of $\DAI$, DeltaDou and SL with the bidding phase by playing 100,000 randomly sampled decks.}
    \label{tab:douzerobidding}
    \begin{tabular}{l|c|c|c}
    \toprule
     
    & $\DAI$ & DeltaDou & SL \\
    \midrule
    WP & \textbf{0.580} & 0.461 & 0.381 \\
    ADP & \textbf{0.323} & -0.004 & -0.320\\
     \bottomrule
    \end{tabular}
    \vspace{-15pt}
\end{table}

\subsection{Comparison with Bidding Phase}
\label{sec:3}
To investigate \textbf{RQ2}, we train a bidding network with supervised learning using human expert data. We place the top-3 algorithms, i.e., $\DAI$, DeltaDou, and SL, into the three seats of a DouDizhu game. In each game, we randomly choose the first bidder and simulate the bidding phase with the pre-trained bidding network. The same bidding network is used for all three algorithms for a fair comparison. The results are summarized in Table~\ref{tab:douzerobidding}. Although $\DAI$ is trained on randomly generated decks without the bidding network, we observe that $\DAI$ dominates the other two algorithms in both WP and ADP. This demonstrates the applicability of $\DAI$ in real-world competitions where the bidding phase needs to be considered.

\begin{figure}[t]
  \centering
  \begin{subfigure}[b]{0.240\textwidth}
    \centering
    \includegraphics[width=1.00\textwidth]{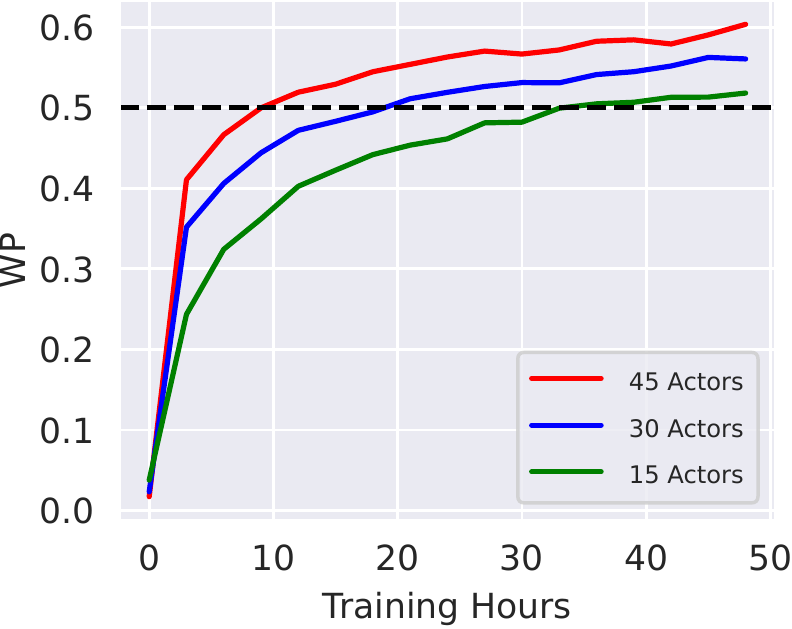}
  \end{subfigure}%
  \begin{subfigure}[b]{0.240\textwidth}
    \centering
    \includegraphics[width=1.00\textwidth]{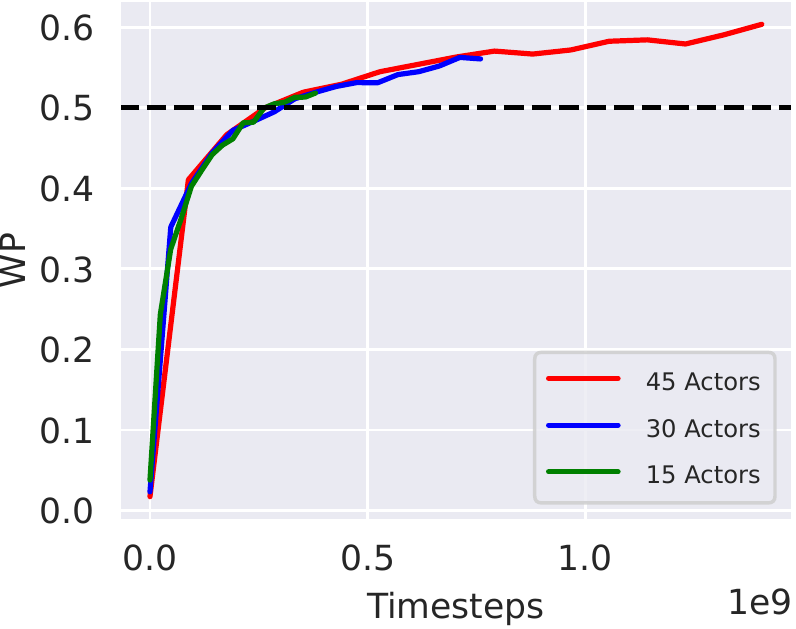}
  \end{subfigure}%
  \vspace{-10pt}
  \caption{\textbf{Left:} The WP against SL w.r.t. training time using different number of actors \textbf{Right:} The WP against SL w.r.t. timesteps using different number of actors.}
  \label{fig:lessactors}
  \vspace{-8pt}
\end{figure}

\subsection{Analysis of Learning Progress}
\label{sec:4}

To study \textbf{RQ3}, we visualize the learning progress of $\DAI$ in Figure~\ref{fig:learningprogress}. We use SL and DeltaDou as opponents to draw the curves of WP and ADP w.r.t. the number of training days. We make two observations as follows. First, $\DAI$ outperforms SL in one day and two days of training in terms of WP and ADP, respectively. We note that $\DAI$ and SL use the exactly same neural architecture for training. Thus, we attribute the superiority of $\DAI$ to self-play reinforcement learning. While SL also performs well, it relies on a large amount of data, which is not flexible and could limit its performance. Second, $\DAI$ outperforms DeltaDou in three days and ten days of training in terms of WP and ADP, respectively. We note that DeltaDou is initialized with supervised learning on heuristics and is trained for more than two months. Whereas, $\DAI$ starts from scratch and only needs days of training to beat DeltaDou. This suggests that model-free reinforcement learning without search is indeed effective in DouDizhu.

We further analyze the learning speed when using different numbers of actors. Figure~\ref{fig:lessactors} reports the performance against SL when using 15, 30, and 45 actors. We observe that using more actors can accelerate the training in wall-clock time. We also find that all three settings show similar sample efficiency. In the future, we will explore the possibility of using more actors across multiple servers to further improve the training efficiency.

\begin{figure}[t]
  \centering
  \begin{subfigure}[b]{0.240\textwidth}
    \centering
    \includegraphics[width=1.00\textwidth]{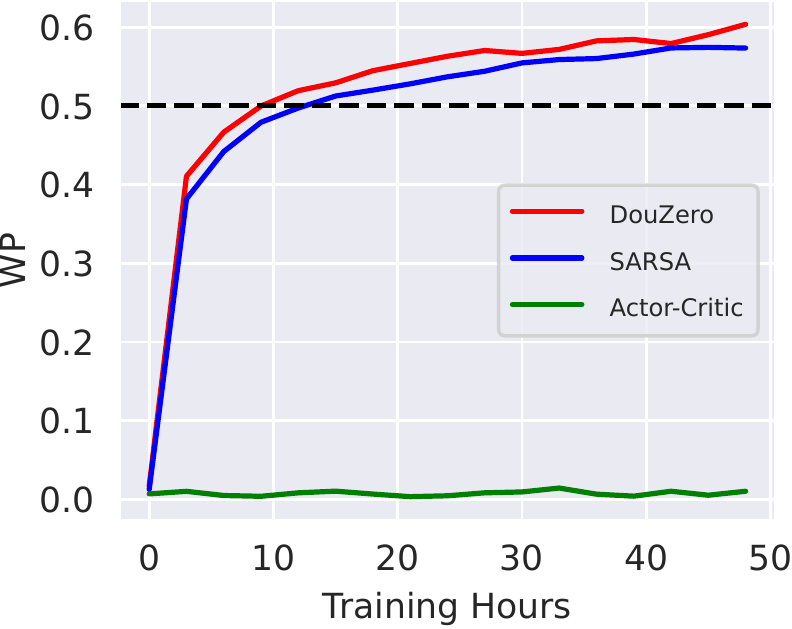}
  \end{subfigure}%
  \begin{subfigure}[b]{0.240\textwidth}
    \centering
    \includegraphics[width=1.00\textwidth]{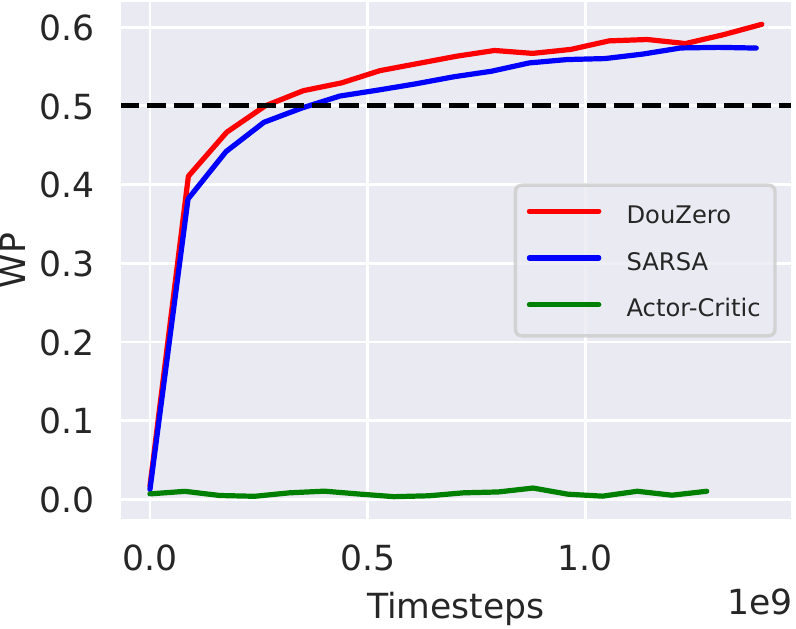}
  \end{subfigure}%
  \vspace{-10pt}
  \caption{\textbf{Left:} The WP against SL w.r.t. training time for SARSA and Actor-Critic \textbf{Right:} The WP against SL w.r.t. timesteps for SARSA and Actor-Critic.}
  \label{fig:sarsaac}
\end{figure}

\begin{figure}[t]
  \centering
    \includegraphics[width=0.40\textwidth]{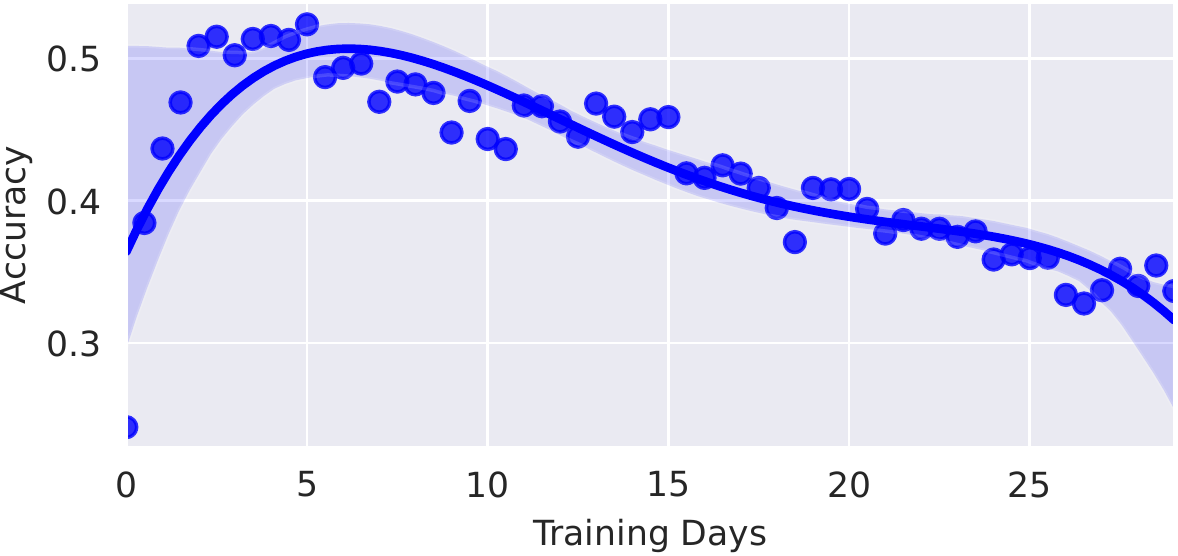}
    \vspace{-5pt}
  \caption{Average accuracy across the three positions on the human data w.r.t. the number of training days for $\DAI$. We fit the data points with a polynomial with four terms for better visualizing the trend. The accuracy for SL is $84.2\%$. $\DAI$ aligns with human expertise in the first five days of training but discovers novel strategies beyond human knowledge in the later training stages. The curves for all the three positions are provided in Appendix~\ref{sec:D4}.}
  \label{fig:humancurve}
  \vspace{-5pt}
\end{figure}

\subsection{Comparison with SARSA and Actor-Critic}
\label{sec:5}
To answer \textbf{RQ4}, we implement two variants based on $\DAI$. First, we replace the DMC objective with the Temporal-Difference (TD) objective. This leads to a deep version of SARSA. Second, we implement an Actor-Critic variant with action features. Specifically, we use Q-network as a critic with action features and train policy as an actor with action masks to remove illegal actions.

Figure~\ref{fig:sarsaac} shows the results of SARSA and Actor-Critic with a single run. First, we do not observe a clear benefit of using TD learning. We observe that DMC learns slightly faster than SARSA in wall-clock time and sample efficiency. The possible reason is that TD learning will not help much in the sparse reward setting. We believe more studies are needed to understand when TD learning will help. Second, we observe the Actor-Critic fails. This suggests that simply adding action features to the critic may not be enough to resolve the complex action space issue. In the future, we will investigate whether we can effectively incorporate action features into the actor-critic framework.

\subsection{Analysis of $\DAI$ on Expert Data}
\label{sec:6}

For \textbf{RQ5}, we calculate the accuracy of $\DAI$ on the human data throughout the training process. We report the model trained with ADP as objective since the game app from which the human data is collected also adopts ADP. Figure~\ref{fig:humancurve} shows the results. We make two interesting observations as follows. First, at the early stages, i.e., the first five days of training, the accuracy keeps improving. This suggests that the agents may have learned some strategies that align with human expertise with purely self-play. Second, after five days of training, the accuracy decreases dramatically. We note that the ADP against SL is still improving after five days. This suggests that the agents may have discovered some novel and stronger strategies that humans can not easily discover, which again verifies the effectiveness of self-play reinforcement learning.

\subsection{Comparison of Inference Time}
\label{sec:7}
\begin{figure}[t]
  \centering
    \includegraphics[width=0.45\textwidth]{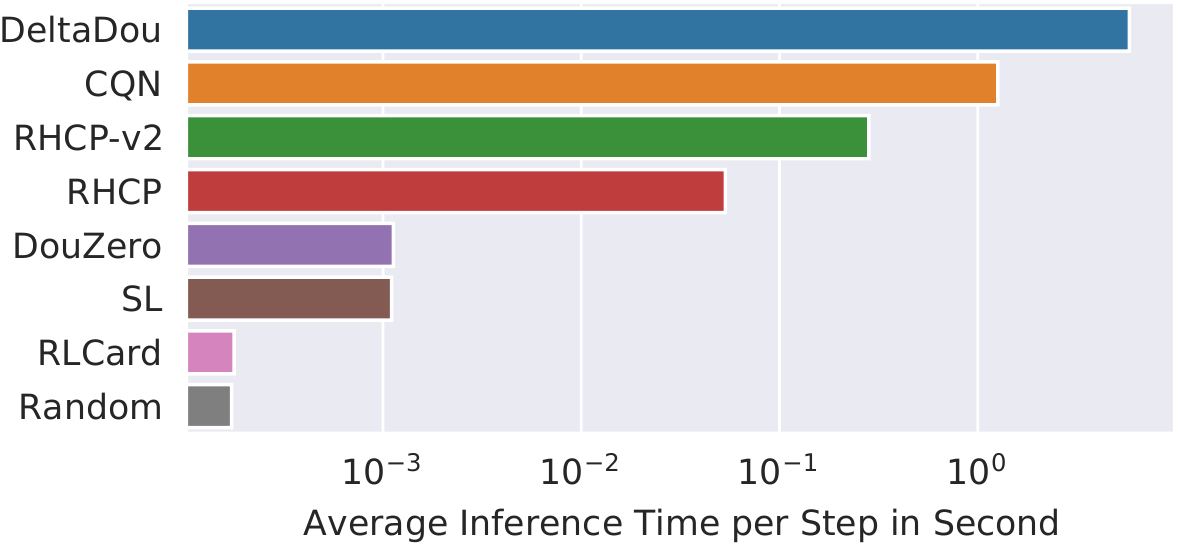}
  \vspace{-10pt}
  \caption{Comparison of inference time.}
  \label{fig:inferencetime}
  \vspace{-18pt}
\end{figure}

To answer \textbf{RQ6}, we report the average inference time per step in Figure~\ref{fig:inferencetime}. For a fair comparison, we evaluate all the algorithms on the CPU. We observe that $\DAI$ is orders of magnitude faster than DeltaDou, CQN, RHCP, and RHCP-v2. This is expected since DeltaDou needs to perform a large number of Monte Carlo simulations, and CQN, RHCP, and RHCP-v2 require expensive card decomposition. Whereas, $\DAI$ only performs one forward pass of neural networks in each step. The efficient inference of $\DAI$ enables us to generate a large number of samples per second for reinforcement learning. It also makes it affordable to deploy the models in real-world applications.

\subsection{Case Study}
\label{sec:8}
To investigate \textbf{RQ7}, we conduct case studies to understand the decisions made by $\DAI$. We dump the logs of the competitions from Botzone and visualize the top actions with their predicted Q-values. We provide most of the case studies, including both good and bad cases, in Appendix~\ref{sec:F}.

Figure~\ref{fig:casestudymain} shows a typical case when the two Peasants can cooperate to beat the Landlord. The Peasant on the right-hand side only has one card left. Here, the Peasant at the bottom can play a small Solo to help the other Peasant win. When looking into the top three actions predicted by $\DAI$, we make two interesting observations. First, we find that all the top actions outputted by $\DAI$ are small Solos with high confidence to win, suggesting that the two Peasants of $\DAI$ may have learned to cooperate. Second, the predicted Q-value of action \texttt{4} (0.808) is much lower than that of action \texttt{3} (0.971). A  possible explanation is that there is still a \texttt{4} out there, so that playing \texttt{4} may not necessarily help the Peasant win. In practice, in this specific case, the other Peasant's only card is not higher than \texttt{4} in rank. Overall, action \texttt{3} is indeed the best move in this case.

\vspace{-5pt}
\section{Related Work}
\textbf{Search for Imperfect-Information Games.} Counterfactual Regret Minimization~(CFR)~\cite{zinkevich2008regret} is a leading iterative algorithm for poker games, with many variants~\cite{lanctot2009monte,gibson2012generalized,bowling2015heads,moravvcik2017deepstack,brown2018superhuman,brown2019solving,brown2019deep,lanctot2019openspiel,li2020openholdem}. However, traversing the game tree of DouDizhu is computationally intensive since it has a huge tree with a large branching factor. Moreover, most of the prior studies focus on zero-sum settings. While some efforts have been devoted to addressing the cooperative settings, e.g., with blueprint policy~\cite{lerer2020improving}, it remains challenging to reason about both competing and cooperation. Thus, DouDizhu has not seen an effective CFR-like solution.

\begin{figure}[t]
  \centering
    \includegraphics[width=0.47\textwidth]{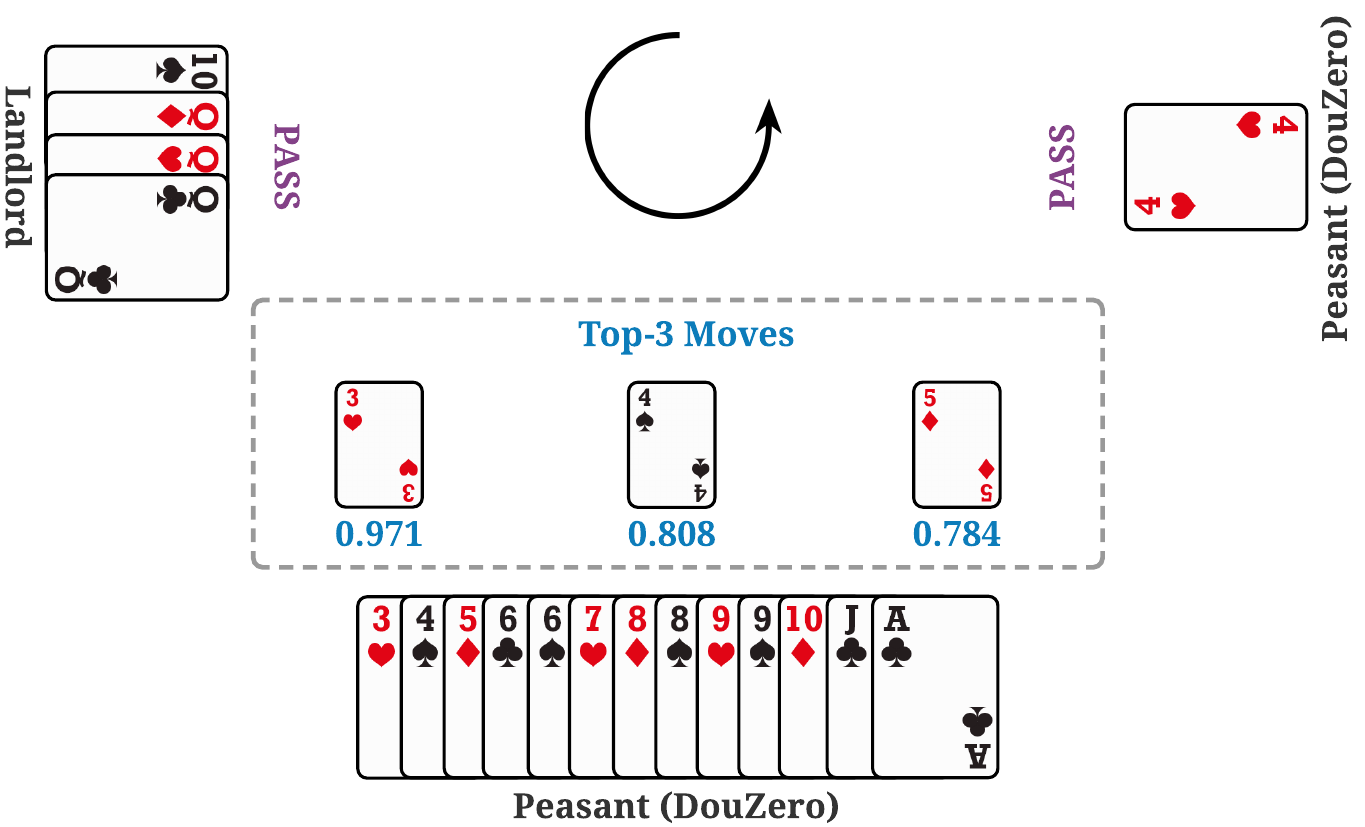}
    \vspace{-10pt}
  \caption{A case study dumped from Botzone, where the three players play cards in counter-clockwise order. The Peasant agent learns to play small Solo to cooperate with the other Peasant to win the game. Note that the other players' hands are showed face up solely for better visualization but are hidden in the real game. More case studies are provided in Appendix~\ref{sec:F}.}
  \label{fig:casestudymain}
  \vspace{-8pt}
\end{figure}

\textbf{RL for Imperfect-Information Games.} Recent studies show that Reinforcement Learning~(RL) can achieve competitive performance in poker games~\cite{heinrich2015fictitious,heinrich2016deep,lanctot2017unified}. Unlike CFR, RL is based on sampling so that it can easily generalize to large-scale games. RL has been successfully applied in some complex imperfect-information games, such as Starcraft~\cite{vinyals2019grandmaster}, DOTA~\cite{berner2019dota} and Mahjong~\cite{li2020suphx}. More recently, RL+search is explored and shown to be effective in poker games~\cite{brown2020combining}. DeltaDou adopts a similar idea, which first infers the hidden information and then uses MCTS to combine RL with search in DouDizhu~\cite{jiang2019deltadou}. However, DeltaDou is computationally expensive and heavily relies on human expertise. In practice, even without search, our $\DAI$ outperforms DeltaDou in days of training. 

\vspace{-5pt}
\section{Conclusions and Future Work}
This work presents a strong AI system for DouDizhu. Some unique properties make DouDizhu particularly challenging to solve, e.g., huge state/action space and reasoning about both competing and cooperation. To address these challenges, we enhance classic Monte-Carlo methods with deep neural networks, action encoding, and parallel actors. This leads to a pure RL solution, namely $\DAI$, which is conceptually simple yet effective and efficient. Extensive evaluations demonstrate that $\DAI$ is the strongest AI program for DouDizhu up to date. We hope the insight that simple Monte-Carlo methods can lead to strong policies in such a hard domain will motivate future research.

For future work, we will explore the following directions. First, we plan to try other neural architectures, such as convolutional neural networks and ResNet~\cite{he2016deep}. Second, we will involve bidding in the loop for reinforcement learning. Third, we will combine $\DAI$ with search at training and/or test time as in~\cite{brown2020combining}, and study how to balance RL and search. Fourth, we will explore off-policy learning to improve the training efficiency. Specifically, we will study whether and how we can improve the wall-clock time and the sample efficiency with experience replay~\citep{lin1992self,zhang2017deeper,zha2019experience,fedus2020revisiting}. Fifth, we will try explicitly modeling collaboration of the Peasants~\cite{panait2005cooperative,foerster2016learning,raileanu2018modeling,lai2020dual}. Sixth, we plan to try scalable frameworks, such as SEED RL~\cite{espeholt2019seed}. Last but not least, we will test the applicability of Monte-Carlo methods on other tasks.

\vspace{-5pt}
\section*{Acknowledgements}
We thank our colleagues in Kuai Inc. for building the DouDizhu environment and the helpful discussions, Qiqi Jiang from DeltaDou team for helping us set up DeltaDou models, and Songyi Huang\footnote{\url{https://github.com/hsywhu}} from RLCard team for developing demo. We would also like to thank the anonymous reviewers and the meta-reviewer for the insightful feedback.

\bibliography{ref}
\bibliographystyle{icml2021}

\appendix

\cleardoublepage
\onecolumn

\section{Introduction of DouDizhu}
\label{sec:A}

As the most popular card game in China, DouDizhu has attracted hundreds of millions of players with many tournaments held every year. DouDizhu is known to be easy to learn but challenging to master. It requires careful planning and strategic thinking. DouDizhu is played among three players. In each game, the players will first bid for the Landlord position. After the bidding phase, one player will become the Landlord, and the other two players will become the Peasants. The two Peasants play as a team to fight against the Landlord. The objective of the game is to be the first player to have no cards left. In addition to the huge state/action spaces and incomplete information, the two Peasants need to cooperate to beat the Landlord. Thus, existing algorithms for poker games, which usually operate on small games and are only designed for two players, are not applicable in DouDizhu. In what follows, we first give an overview of the game rule of DouDizhu and then analyze the state/action spaces of DouDizhu. Readers who are familiar with the game may skip Section~\ref{sec:A1}. Readers who are not familiar with DouDizhu may also refer to Wikipedia~\footnote{\url{https://en.wikipedia.org/wiki/Dou_dizhu}} for more introduction.

\subsection{Rules}
\label{sec:A1}
DouDizhu is played with one pack of cards, including the two jokers. Suits are irrelevant in DouDizhu. The cards are ranked by Red Joker, Black Joker, 2, A, K, Q, J, 10, 9, 8, 7, 6, 5, 4, 3. Each game has three phases as follows.
\begin{itemize}
    \item \textbf{Dealing:} A shuffled pack of 54 cards will be dealt to the three players. Each player will be dealt 17 cards, and the last three leftover cards will be kept on the deck, face down. These three cards will be dealt to the Landlord, which are decided in the bidding phase.
    \item \textbf{Bidding:} The three players will analyze their own cards without showing to other players. The players decide whether they would like to bid the Landlord based on their hand cards' strength. There are many versions of bidding rules. In this paper, we consider a version adopted in most online DouDizhu app. The first bidder will be randomly chosen. The first bidder will then decide whether she bids. If the first bidder does not bid, the other players will become the bidder in turn until someone bids. If no one bids, a new pack of cards will be dealt to the players. If one chooses to bid, the other players will decide whether she accepts the bid or she wants to outbid. Each player only has one chance to outbid. The last player who bids or outbids will become the Landlord. Once the Landlord is settled, the Landlord will be dealt with the three cards on the deck. The other two players will play as the Peasants to fight against the Landlord.
    \item \textbf{Card-Playing:} In this phase, the players will play cards in turn starting from the Landlord. The first player can choose either category of cards such as Solo, Pair, etc. (detailed in the next paragraph). Then the next player must play the cards in the same category with a higher rank. The next player can also choose ``PASS" if she does not have a higher rank in hand or she does not want to follow the category. If all the other players choose ``PASS," the player who first plays the category can freely play cards in other categories. The players will play cards in turn until one player has no cards left. The Landlord wins if she has no cards left. The Peasant team wins if either of the peasants has no cards left. The two Peasants need to cooperate to increase the possibility of winning. A Peasant may still win a game by helping the other Peasant win even if she has terrible hand cards.
\end{itemize}

One challenge of DouDizhu is the rich categories, which consist of various combinations of cards. For some categories, the player can choose a kicker card, which can be any card in hand. One will usually choose a useless card as a kicker card so that she can more easily go out of hand. As a result, the player needs to carefully plan how to play the cards to win a game. The categories in DouDizhu are listed as follows. Note that Bomb and Rocket defy the category rules and can dominate all the other categories.
\vspace{-5pt}
\begin{itemize}
    \item \textbf{Solo:} Any single card.
    \item \textbf{Pair:} Two matching cards of equal rank.
    \item \textbf{Trio:} Three individual cards of equal rank.
    \item \textbf{Trio with Solo:} Three individual cards of equal rank with a Solo as the kicker.
    \item \textbf{Trio with Pair:} Three individual cards of equal rank with a Pair as the kicker.
    \item \textbf{Chain of Solo:} $\ge$Five consecutive individual cards.
    \item \textbf{Chain of Pair:} $\ge$Three consecutive Pairs.
    \item \textbf{Chain of Trio:} $\ge$Two consecutive Trios.
    \item \textbf{Plane with Solo:} $\ge$Two consecutive trios with each has a distinct individual kicker card.
    \item \textbf{Quad with Pair:} Four-of-a-kind with two sets of Pair as the kicker.
    \item \textbf{Bomb:} Four-of-a-kind.
    \item \textbf{Rocket:} Red and Black jokers.
\end{itemize}

\vspace{-5pt}
\subsection{State and Action Space of DouDizhu}
According to the estimation in RLCard~\cite{zha2019rlcard}, the number of information sets in DouDizhu is up to $10^{83}$ and the average size of each information set is up to $10^{23}$. While the number of information sets is smaller than that of No-limit Texas Hold'em ($10^{126}$), the average size of each information set is much larger than that of No-limit Texas Hold'em ($10^4$). Different from Hold'em games, the state space of DouDizhu can not be easily abstracted. Specifically, every card matters in DouDizhu towards winning. For example, the number of cards of rank 2 in the historical moves is crucial since the players need to decide whether their cards will be dominated by other players with a 2. Thus, a very slight difference in the state representation could significantly impact the strategy. While the size of the state space of DouDizhu is not as large as that of No-limit Texas Hold'em, learning an effective strategy is very challenging since the agents need to distinguish different states accurately. $\DAI$ approaches this problem by extracting representations and learning the strategy automatically with deep neural networks.

DouDizhu suffers from an explosion of action space due to the combinations of cards. We summarize the action space in Table~\ref{tab:sumactionspace}. The size of the action space of DouDizhu is $27,472$, which is much larger than Mahjong ($10^2$). It is also much more complicated than No-limit Texas Hold'em, whose action space can be easily abstracted. Specifically, in DouDizhu, every card matters. For example, for the action type \texttt{Trio with Solo}, wrongly choosing the kicker may directly result in a loss since it could potentially break a chain. Thus, it is difficult to abstract the action space. This poses challenges for reinforcement learning since most of the algorithms only work well on small action space. In contrast to the previous work that abstracts the action space with heuristics~\cite{jiang2019deltadou}, $\DAI$ approaches this issue with Monte-Carlo methods, which allow flexible exploration of the action space to potentially discover better moves.
\vspace{-5pt}
\begin{table}[h!]
    \centering
    \caption{Summary of the action space of DouDizhu. We follow the summary provided in RLCard~\cite{zha2019rlcard}.}
    \label{tab:sumactionspace}
    \begin{tabular}{l|l}
    \toprule
    \textbf{Action Type} & \textbf{Number of Actions} \\
    \midrule
     Solo    &  $15$\\
     Pair    &  $13$\\
     Trio    &  $13$\\
     Trio with Solo & $182$\\
     Trio with Pair & $156$\\
     Chain of Solo & $36$\\
     Chain of Pair & $52$\\
     Chain of Trio & $45$\\
     Plane with Solo & $21,822$\\
     Plane with Pair & $2,939$\\
     Quad with Solo & $1,326$\\
     Quad with Pair & $858$\\
     Bomb & $13$\\
     Rocket & $1$\\
     Pass & $1$\\
     
     \midrule
     Total & $27,472$\\
     \bottomrule
    \end{tabular}
\end{table}

\cleardoublepage

\section{Additional Examples of Card Representations}
\label{sec:B}

\begin{figure}[H]
  \centering

  \begin{subfigure}[b]{0.5\textwidth}
    \centering
    \includegraphics[width=0.8\textwidth]{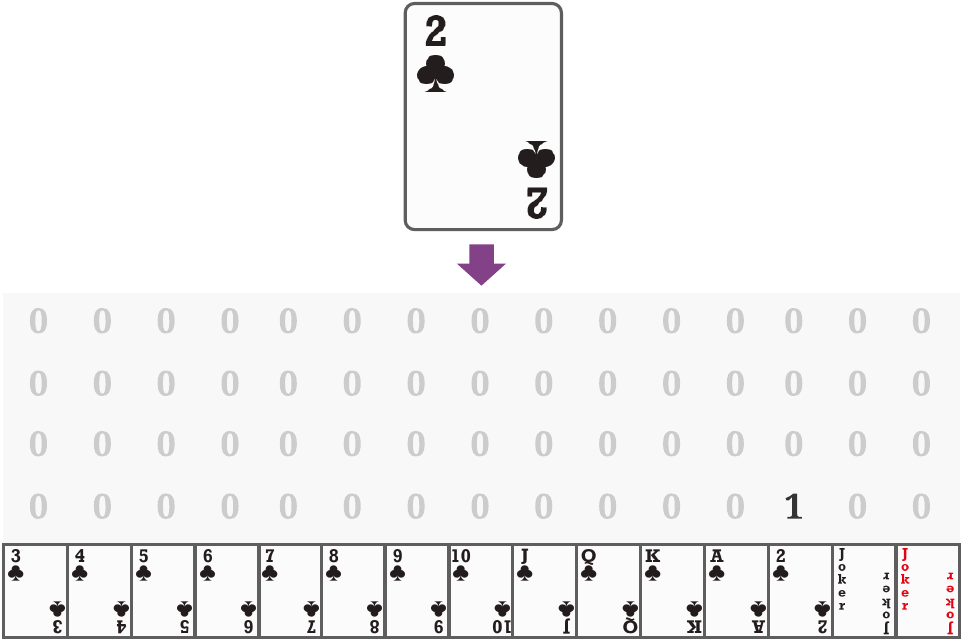}
    \caption{Solo}
  \end{subfigure}%
  \begin{subfigure}[b]{0.5\textwidth}
    \centering
    \includegraphics[width=0.8\textwidth]{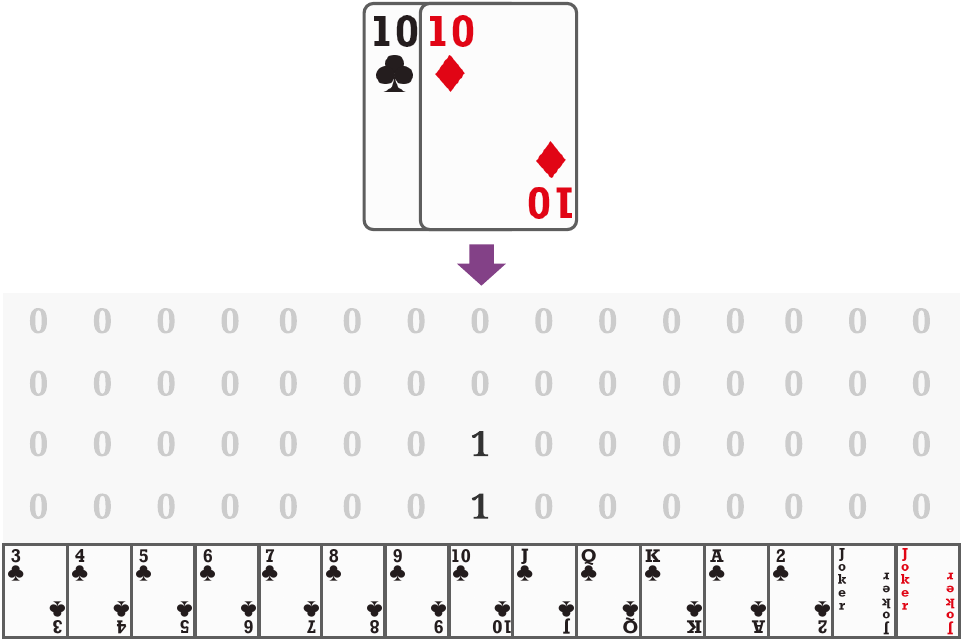}
    \caption{Pair}
  \end{subfigure}%
  
  \begin{subfigure}[b]{0.5\textwidth}
    \centering
    \includegraphics[width=0.8\textwidth]{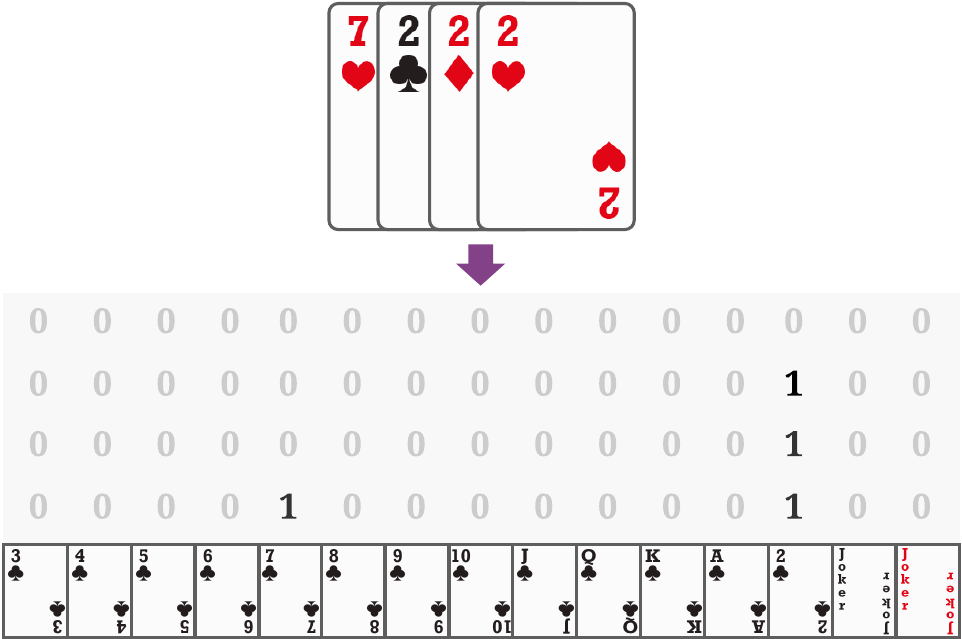}
    \caption{Trio with Solo}
  \end{subfigure}%
  \begin{subfigure}[b]{0.5\textwidth}
    \centering
    \includegraphics[width=0.8\textwidth]{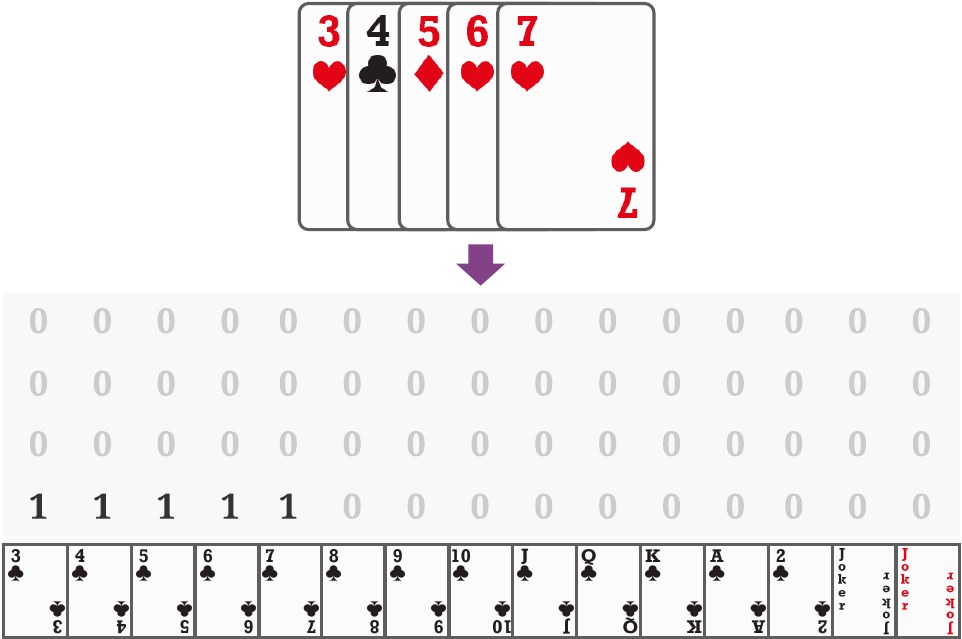}
    \caption{Chain of solo}
  \end{subfigure}%
 
  \begin{subfigure}[b]{0.5\textwidth}
    \centering
    \includegraphics[width=0.8\textwidth]{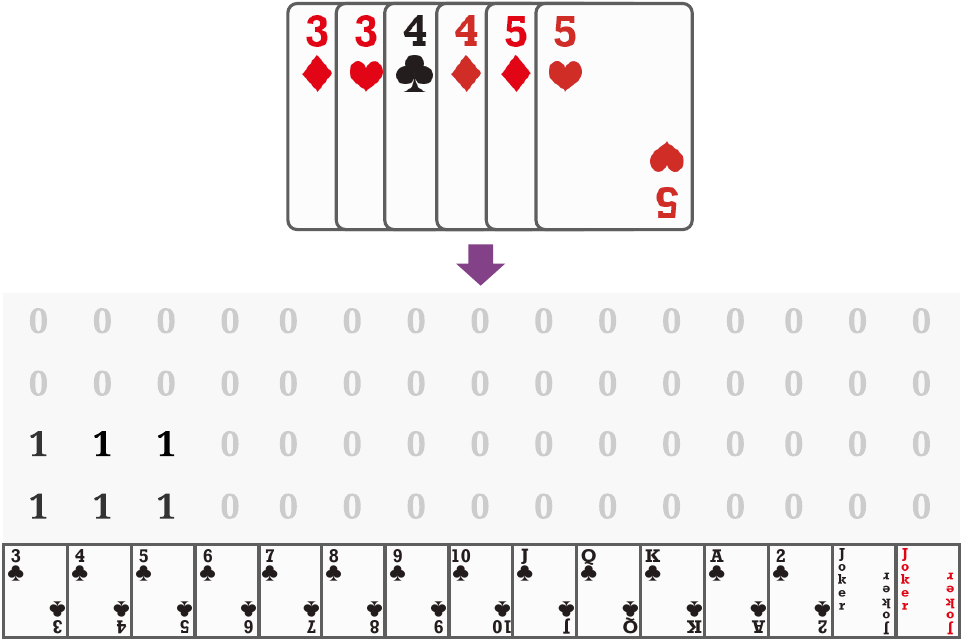}
    \caption{Chain of pair}
  \end{subfigure}%
  \begin{subfigure}[b]{0.5\textwidth}
    \centering
    \includegraphics[width=0.8\textwidth]{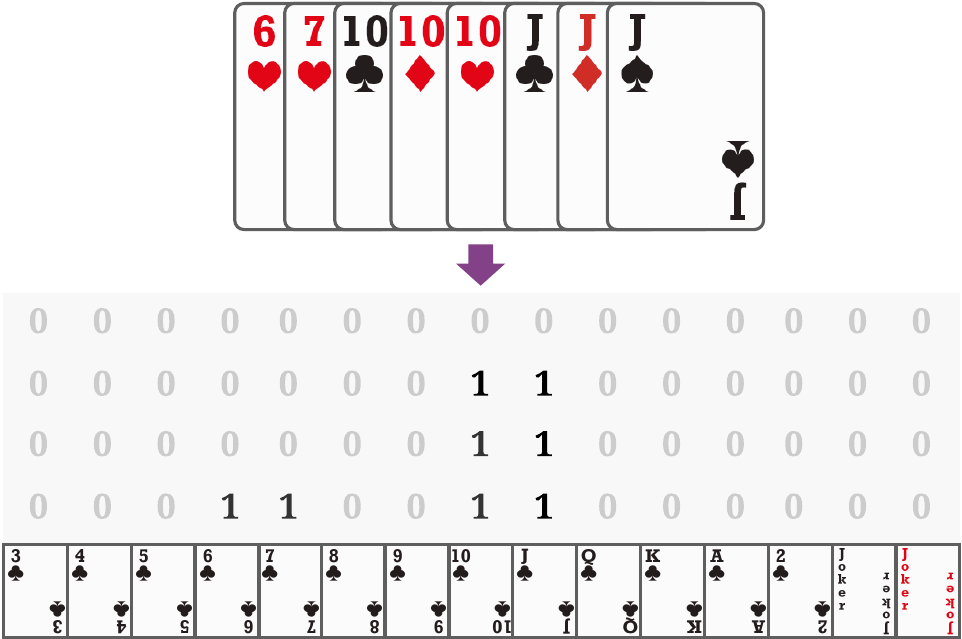}
    \caption{Plane with solo}
  \end{subfigure}%
  
  \begin{subfigure}[b]{0.5\textwidth}
    \centering
    \includegraphics[width=0.8\textwidth]{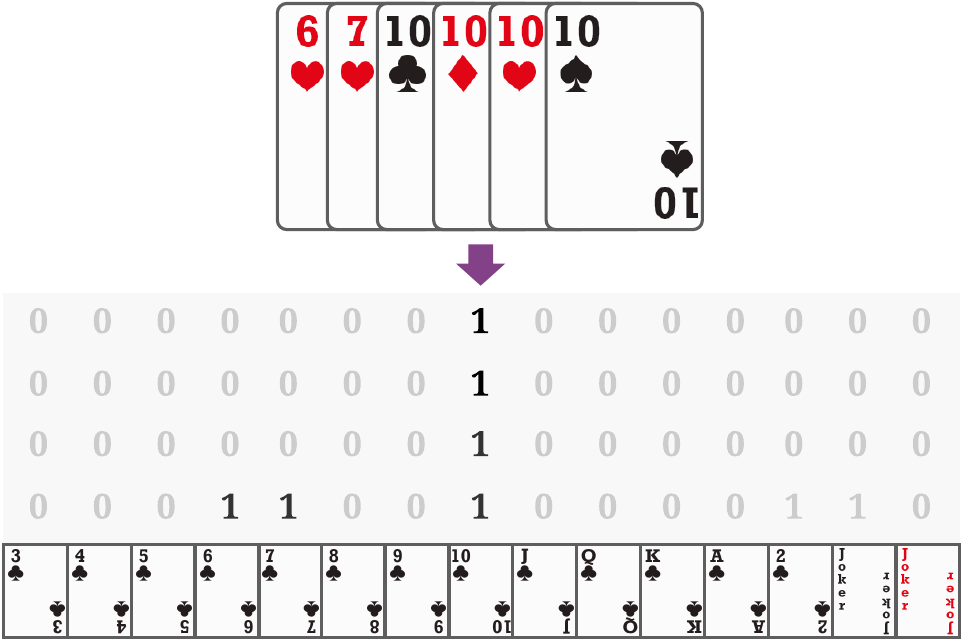}
    \caption{Quad with solo}
  \end{subfigure}%
  \begin{subfigure}[b]{0.5\textwidth}
    \centering
    \includegraphics[width=0.8\textwidth]{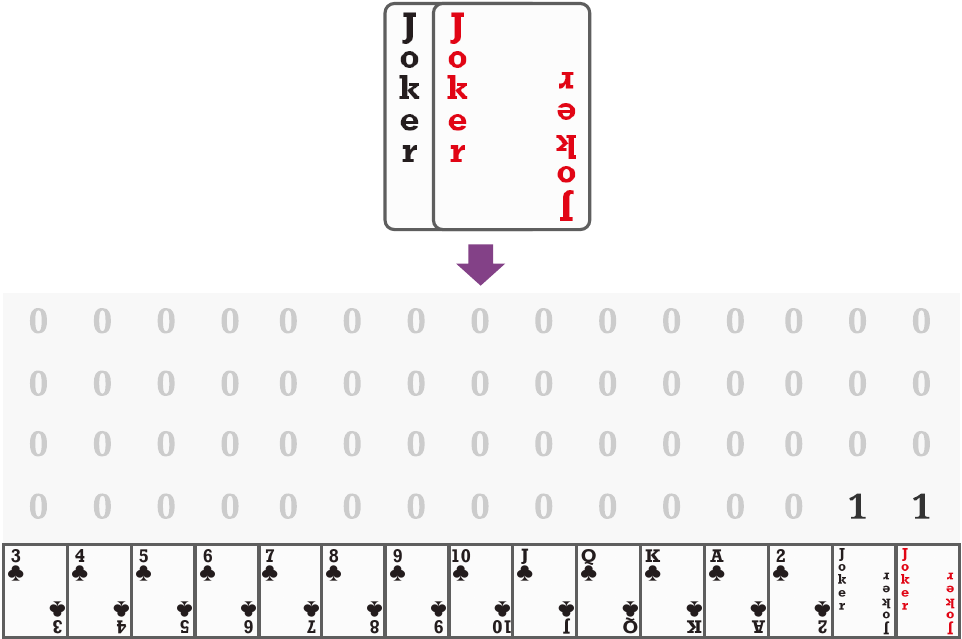}
    \caption{Rocket}
  \end{subfigure}%
  \caption{Additional examples of encoding different types of cards.}
  \label{fig:cardrepresentationexamples}
\end{figure}

\cleardoublepage

\section{Additional Details of Feature Representation and Neural Architecture}
\label{sec:C}

\subsection{Action and State Representation}
\label{sec:C1}
The input of the neural network is the concatenated representation of state and action. For each $15 \times 4$ card matrix, we first flatten the matrix into a 1-dimensional vector of size 60. Then we remove six entries that are always zero since there is only one black or red joker. In other words, each card matrix is transformed into a one-hot vector of size 54. In addition to card matrices, we further use a one-hot vector to represent the other two players' current hand cards. For example, for Peasant, we use a vector of size 17, where each entry corresponds to the number of hand cards in the current state. For the Landlord, the vector's size is 20 since the Landlord can have at most 20 cards in hand. Similarly, we use a 15-dimension vector to represent the number of bombs in the current state. For historical moves, we consider the most recent 15 moves and concatenate the representations of every three consecutive moves; that is, the historical moves are encoded into a $5 \times 162$ matrix. The historical moves are fed into an LSTM, and we use the hidden representation in the last cell to represent the historical moves. If there are less than 15 moves historically, we use zero matrices for the missing moves. We summarize the encoded features of Landlord and each Peasant in Table~\ref{tab:landlordfeature} and Table~\ref{tab:peasantfeature}, respectively.

\begin{table}[h!]
    \centering
    \caption{Features of the Landlord.}
    \label{tab:landlordfeature}
    \begin{tabular}{c|l|l}
    \toprule
     & \textbf{Feature} & \textbf{Size} \\
    \midrule
    \midrule
     Action & Card matrix of the action    &  $54$\\
     
     \midrule
     \multirow{2}{*}{State} & Card matrix of hand cards & $54$\\
      & Card matrix of the union of the other two players' hand cards & $54$\\
      & Card matrix of the most recent move & $54$\\
      & Card matrix of the the played cards of the first Peasant & $54$\\
      & Card matrix of the the played cards of the second Peasant & $54$\\
      & One-hot vector representing the number cards left of the first Peasant & $17$\\
      & One-hot vector representing the number cards left of the second Peasant & $17$\\
      & One-hot vector representing the number bombs in the current state & $15$\\
      & Concatenated matrix of the most recent 15 moves & $5 \times 162$\\
     \bottomrule
    \end{tabular}
    
\end{table}

\begin{table}[h!]
    \centering
    \caption{Features of the Peasants.}
    \label{tab:peasantfeature}
    \begin{tabular}{c|l|l}
    \toprule
     & \textbf{Feature} & \textbf{Size} \\
    \midrule
    \midrule
     Action & Card matrix of the action    &  $54$\\
     
     \midrule
     \multirow{2}{*}{State} & Card matrix of hand cards & $54$\\
      & Card matrix of the union of the other two players' hand cards & $54$\\
      & Card matrix of the most recent move & $54$\\
      & Card matrix of the most recent move performed by the Landlord & $54$\\
      & Card matrix of the most recent move performed by the other Peasant & $54$\\
      & Card matrix of the the played cards of the Landlord & $54$\\
      & Card matrix of the the played cards of the other Peasant & $54$\\
      & One-hot vector representing the number cards left of the Landlord & $20$\\
      & One-hot vector representing the number cards left of the other Peasant & $17$\\
      & One-hot vector representing the number bombs in the current state & $15$\\
      & Concatenated matrix of the most recent 15 moves & $5 \times 162$\\
     \bottomrule
    \end{tabular}
    
\end{table}

\subsection{Data Collection and Neural Architecture of Supervised Learning}
\label{sec:C2}
\begin{figure}[h!]
  \centering
    \includegraphics[width=0.70\textwidth]{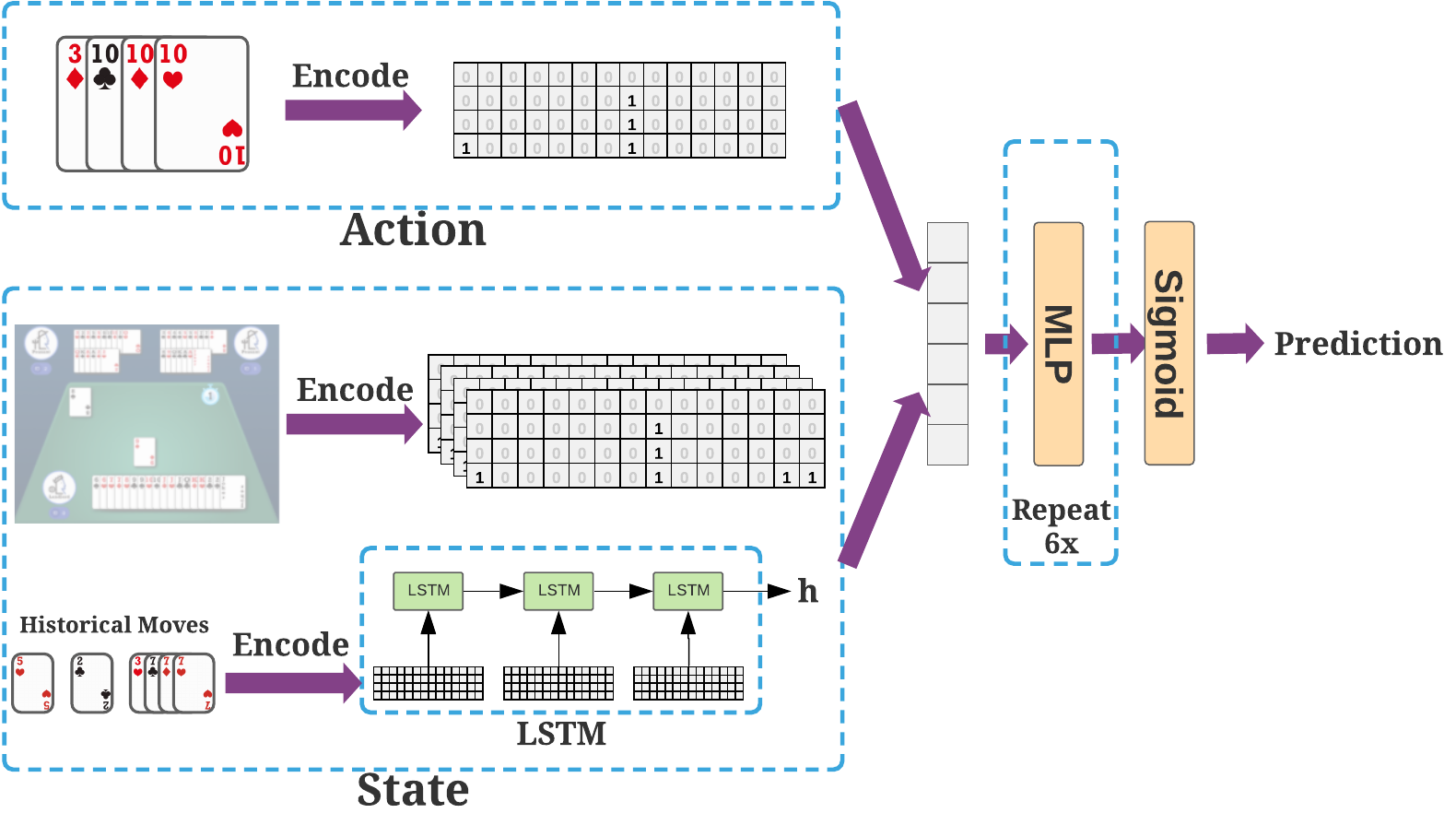}
  \caption{We use the same neural architecture as $\DAI$ for SL. We add a Sigmoid function to the output and transform the problem into a binary classification task. The agent will perform the action that leads to the highest prediction score.}
  \label{fig:supervisedoverview}
\end{figure}

\begin{figure*}[h!]
  \centering

  \begin{subfigure}[b]{0.33\textwidth}
    \centering
    \includegraphics[width=0.95\textwidth]{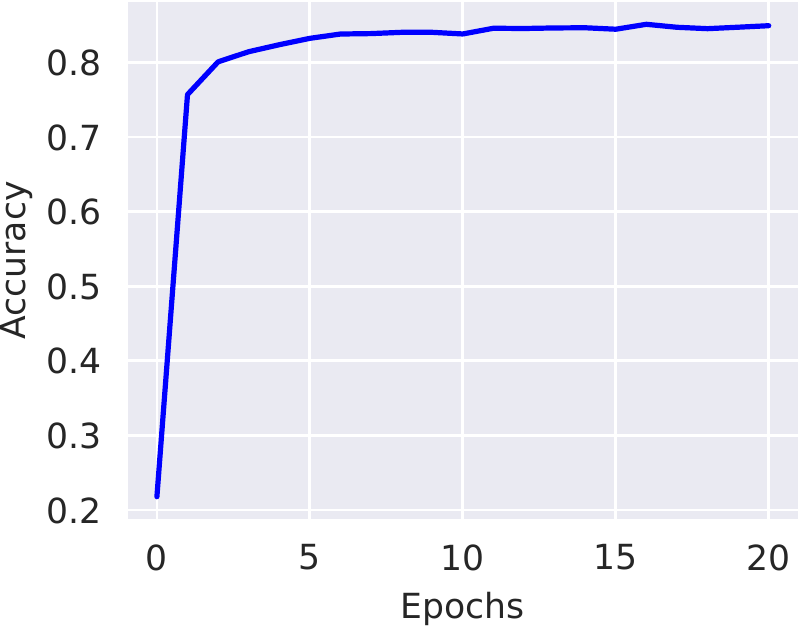}
    \caption{Landlord}
  \end{subfigure}%
  \begin{subfigure}[b]{0.33\textwidth}
    \centering
    \includegraphics[width=0.95\textwidth]{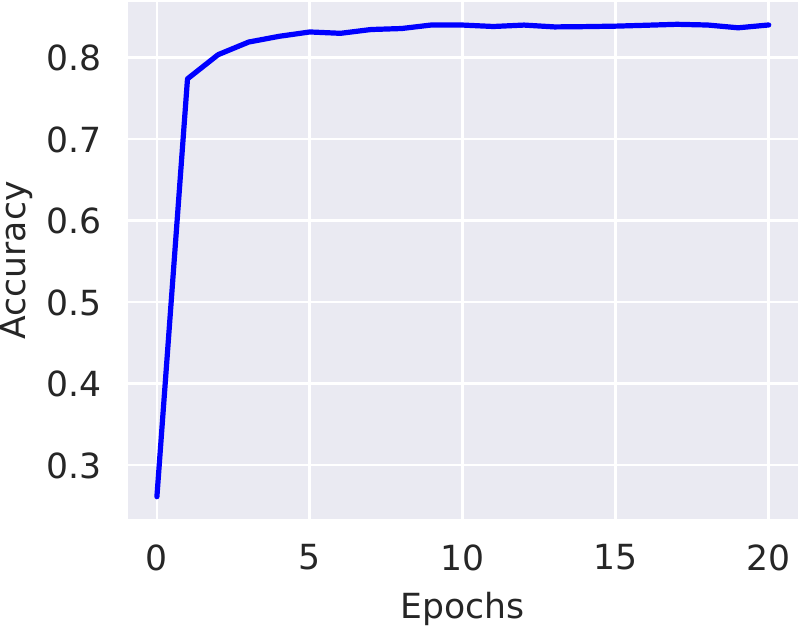}
    \caption{LandlordUp}
  \end{subfigure}%
  \begin{subfigure}[b]{0.33\textwidth}
    \centering
    \includegraphics[width=0.95\textwidth]{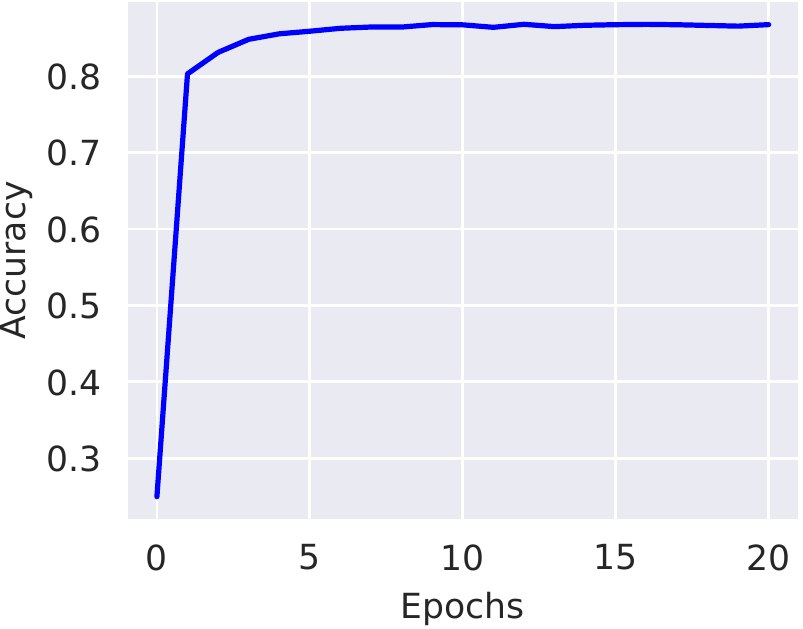}
    \caption{LandlordDown}
  \end{subfigure}%
  \caption{Accuracy w.r.t. the number of training epochs of SL for the three positions. LandlordUp stands for the Peasant that moves before the Landlord. LandlordDown stands for the Peasant that moves after the Landlord.}
  \label{fig:sllearningprogress}
\end{figure*}

In order to train an agent with supervised learning, we collect user data internally from a popular DouDizhu game mobile app. The users in the app have different leagues, which represent the strengths of the users. We filter out the raw data by only keeping the data generated by the players of the highest league to ensure the quality of the data. After filtering, we obtain 226,230 human expert matches. We treat each move as an instance and use a supervised loss to train the networks. The problem can be formulated as a classification problem, where we aim at predicting the action based on a given state, with a total of $27,472$ classes. However, we find in practice that most of the actions are illegal, and it is expensive to iterate over all the classes. Motivated by Q-network's design, we transform the problem into a binary classification task, as shown in Figure~\ref{fig:supervisedoverview}. Specifically, we use the same neural architecture as $\DAI$ and add a Sigmoid function to the output. We then use binary cross-entropy loss to train the network. We randomly sample $10\%$ of the data for validation purposes and use the rest for training. We transform the user data into positive instances and generate negative instances based on the legal moves that are not selected. Eventually, the training data consists of $49,990,075$ instances. We further find that the data is imbalanced, where the number of negative instances is much larger than that of positive instances. Thus, we adopt a re-weighted cross-entropy loss based on the distribution of positive and negative instances. We find in practice that the re-weighted loss can improve the performance. We set the batch size to be 8096 and train 20 epochs. The prediction is made by choosing the action that leads to the highest score. We output the model that has the highest accuracy on the validation data. We do this process three times for the three positions, respectively. We plot the validation accuracy w.r.t. the number epochs in Figure~\ref{fig:sllearningprogress}. The network can achieve around $84\%$ accuracy for all positions.

\subsection{Neural Architecture and Training Details of Bidding Network}
\label{sec:C3}
\begin{figure}[h!]
  \centering
    \includegraphics[width=0.45\textwidth]{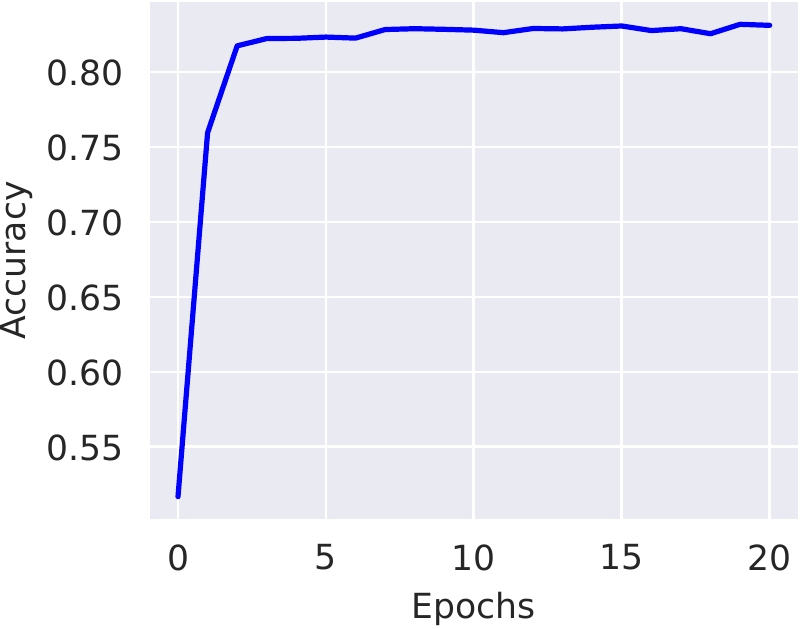}
  \caption{Accuracy w.r.t. the number of training epochs of the bidding network.}
  \label{fig:sllearningprogressbidding}
\end{figure}

The bidding phase's goal is to determine whether a player should become the landlord based on the strengths of the hand cards. This decision is much simpler than card-playing since the agent only needs to consider the hand cards and the other players' decisions, and we only need to make a binary prediction, i.e., whether we bid. At the beginning of the bidding phase, a randomly chosen player will decide whether to bid or not. Then the other two players will also choose whether to bid. If only one player bids, then that player will become the landlord. Suppose two or more players bid, the player who bids first will have the priority to decide whether she wants to become the landlord. We extract 128 features to represent hand cards and the players' moves, as summarized in Table~\ref{tab:sumbiddingfeature}. For the network architecture, we use a (512, 256, 128, 64, 32, 16) MLP.  Like the supervised card playing agent, we add a Sigmoid function to the output and train the network with binary cross-entropy loss. We plot the validation accuracy w.r.t. the number epochs in Figure~\ref{fig:sllearningprogressbidding}. The network can achieve $83.1\%$ accuracy.

\begin{table}[h!]
    \centering
    \caption{Features of the bidding network.}
    \label{tab:sumbiddingfeature}
    \begin{tabular}{l|l}
    \toprule
    \textbf{Feature} & \textbf{Size} \\
    \midrule
    Card matrix of hand cards    &  $54$\\
    A vector representing solos of ranks 3 to A  &  $12$\\
    A vector representing pairs of ranks 3 to 2  &  $13$\\
    A vector representing trios of ranks 3 to 2  &  $13$\\
    A vector representing bombs of ranks 3 to 2 and the rocket  &  $14$\\
    The number of cards of rank 2 and the jokers  &  $10$\\
    A vector encoding historical bidding moves  &  $12$\\
     \midrule
     Total & $128$\\
     \bottomrule
    \end{tabular}
\end{table}

\cleardoublepage

\section{Additional Results of $\DAI$}
\label{sec:D}

\subsection{Full WP and ADP Results for Landlord and Peasants}
\label{sec:D1}

We report the results for Landlord and Peasants in Table~\ref{tab:douzerofullwp} and Table~\ref{tab:douzerofulladp} for WP and ADP, respectively. We observe that the advantage of $\DAI$ for Peasants tends to be larger than that of Landlord. A possible explanation is that the two Peasants agents in $\DAI$ have learned cooperation skills, which could be hardly covered by the heuristics and other algorithms.

\begin{table}[h!]
    \centering
    \small
    \caption{WP of $\DAI$ and the baselines. L: WP of \texttt{A} as Landlord; P: WP of \texttt{A} as Peasants. If the average WP of L and P is higher than 0.5, we conclude that \texttt{A} outperforms \texttt{B} and highlight both L and P in boldface. The algorithms are ranked according to the number of the other algorithms that they beat.}.
    \label{tab:douzerofullwp}
    \setlength{\tabcolsep}{2.3pt}
    \begin{tabular}{c||l|cc|cc|cc|cc|cc|cc|cc|cc}
    \toprule
     
\multirow{2}{*}{Rank} & \multirow{2}{*}{\diagbox [width=5em,trim=l] {\texttt{A}}{\texttt{B}}} & \multicolumn{2}{c|}{$\DAI$} & \multicolumn{2}{c|}{DeltaDou} & \multicolumn{2}{c|}{SL} & \multicolumn{2}{c|}{RHCP-v2} & \multicolumn{2}{c|}{RHCP} & \multicolumn{2}{c|}{RLCard} & \multicolumn{2}{c|}{CQN} &\multicolumn{2}{c}{Random}\\
\cline{3-18}
& & L & P & L & P & L & P & L & P & L & P & L & P & L & P & L & P\\
    \hline
    \midrule
     1 & $\DAI$ & .4159 & .5841 & \textbf{.4870} & \textbf{.6843} & \textbf{.5692} & \textbf{.7494} & \textbf{.6844} & \textbf{.8303} & \textbf{.7253} & \textbf{.8033} & \textbf{.8695} & \textbf{.9089} & \textbf{.7686} & \textbf{.8513} & \textbf{.9858} & \textbf{.9920}\\
     2 & DeltaDou & .3166 & .5130 & .4120 & .5880 & \textbf{.5130} & \textbf{.7211} & \textbf{.6701} & \textbf{.8165} & \textbf{.7048} & \textbf{.7899} & \textbf{.8563} & \textbf{.8955} & \textbf{.7326} & \textbf{.8351} & \textbf{.9871} & \textbf{.9960}\\
     3 & SL & .2506 & .4308 & .2789 & .5130 & .4072 & .5928 & \textbf{.5370} & \textbf{.6857} & \textbf{.5831} & \textbf{.6810} & \textbf{.7605} & \textbf{.8650} & \textbf{.6450} & \textbf{.7428} & \textbf{.9599} & \textbf{.9927}\\
     4 & RHCP-v2 & .1697 & .3156 & .1835 & .3299 & .3143 & .4630 & .4595 & .5405 & \textbf{.5134} & \textbf{.5165} & \textbf{.6813} & \textbf{.7018} & \textbf{.6313} & \textbf{.6116} & \textbf{.9519} & \textbf{.9821}\\
     5 & RHCP & .1967 & .2747 & .2101 & .2952 & .3190 & .4179 & .4835 & .4866 & .4971 & .5029 & \textbf{.6718} & \textbf{.6913} & \textbf{.6416} & \textbf{.5640} & \textbf{.9092} & \textbf{.9725}\\
     6 & RLCard & .0911 & .1305 & .1045 & .1437 & .1350 & .2395 & .2982 & .3187 & .3087 & .3282 & .4465 & .5535 & \textbf{.5839} & \textbf{.4603} & \textbf{.9314} & \textbf{.9539}\\
     7 & CQN & .1487 & .2314 & .1649 & .2674 & .2572 & .3550 & .3884 & .3687 & .4360 & .3584 & .5397 & .4161 & .5238 & .4762 & \textbf{.8566} & \textbf{.9213}\\
     8 & Random & .0080 & .0142 & .0040 & .0129 & .0073 & .0401 & .0179 & .0481 & .0025 & .0908 & .0461 & .0686 & .0787 & .1434 & .3461 & .6539\\
   
     \bottomrule
    \end{tabular}

\end{table}

\begin{table}[h!]
    \centering
    \small
    \caption{ADP of $\DAI$ and the baselines. L: ADP of \texttt{A} as Landlord; P: ADP of \texttt{A} as Peasants. If the average ADP of L and P is higher than 0, we conclude that \texttt{A} outperforms \texttt{B} and highlight both L and P in boldface. The algorithms are ranked according to the number of the other algorithms that they beat.}.
    \label{tab:douzerofulladp}
    \setlength{\tabcolsep}{1.2pt}
    \begin{tabular}{c||l|cc|cc|cc|cc|cc|cc|cc|cc}
    \toprule
     
\multirow{2}{*}{Rank} & \multirow{2}{*}{\diagbox [width=5em,trim=l] {\texttt{A}}{\texttt{B}}} & \multicolumn{2}{c|}{$\DAI$} & \multicolumn{2}{c|}{DeltaDou} & \multicolumn{2}{c|}{SL} & \multicolumn{2}{c|}{RHCP-v2} & \multicolumn{2}{c|}{RHCP} & \multicolumn{2}{c|}{RLCard} & \multicolumn{2}{c|}{CQN} &\multicolumn{2}{c}{Random}\\
\cline{3-18}
& & L & P & L & P & L & P & L & P & L & P & L & P & L & P & L & P\\
    \hline
    \midrule
     1 & $\DAI$ & -0.435 & 0.435 & \textbf{-0.342} & \textbf{0.858} & \textbf{0.287} & \textbf{1.112} & \textbf{1.436} & \textbf{1.888} & \textbf{1.492} & \textbf{1.850} & \textbf{2.222} & \textbf{2.354} & \textbf{1.368} & \textbf{2.001} & \textbf{3.254} & \textbf{2.818}\\
     2 & DeltaDou & 0.342 & -0.858 & -0.476 & 0.476 & \textbf{0.268} & \textbf{1.038} & \textbf{1.297} & \textbf{1.703} & \textbf{1.312} & \textbf{1.715} & \textbf{2.270} & \textbf{2.648} & \textbf{1.218} & \textbf{1.849} & \textbf{3.268} & \textbf{2.930}\\
     3 & SL & -1.112 & -0.287 & -1.038 & -0.268 & -0.364 & 0.364 & \textbf{0.564} & \textbf{1.142} & \textbf{0.658} & \textbf{1.114} & \textbf{1.652} & \textbf{1.990} & \textbf{0.878} & \textbf{1.196} & \textbf{3.026} & \textbf{2.415}\\
     4 & RHCP-v2 & -1.888 & -1.436 & -1.703 & -1.297 & -1.142 & -0.564 & -0.209 & 0.209 & \textbf{0.074} & \textbf{0.029} & \textbf{1.011} & \textbf{1.230} & \textbf{0.750} & \textbf{0.677} & \textbf{2.638} & \textbf{2.624}\\
     5 & RHCP & -1.850 & -1.492 & -1.715 & -1.312 & -1.114 & -0.658 & -0.029 & -0.074 & -0.007 & 0.007 & \textbf{1.190} & \textbf{1.328} & \textbf{0.927} & \textbf{-0.432} & \textbf{2.722} & \textbf{2.717}\\
     6 & RLCard & -2.354 & -2.222 & -2.648 & -2.270 & -1.990 & -1.652 & -1.230 & -1.011 & -1.328 & -1.190 & -0.266 & 0.266 & \textbf{0.474} & \textbf{-0.138} & \textbf{2.630} & \textbf{2.312}\\
     7 & CQN & -2.001 & -1.368 & -1.849 & -1.218 & -1.196 & -0.878 & -0.677 & -0.750 & 0.432 & -0.927 & 0.138 & -0.474 & 0.056 & -0.056 & \textbf{1.832} & \textbf{1.992}\\
     8 & Random & -2.818 & -3.254 & -2.930 & -3.268 & -2.415 & -3.026 & -2.624 & -2.638 & -2.717 & -2.722 & -2.312 & -2.629 & -1.991 & -1.832 & -0.883 & 0.883\\
   
     \bottomrule
    \end{tabular}

    \vspace{-10pt}
\end{table}

\subsection{Comparison of Using WP and ADP as Objectives}
\label{sec:D2}

In our experiments, we find that the agents will learn different styles of card playing strategies when using WP and ADP as objectives. Specifically, we observe that the agents trained with WP play more aggressively about bombs even if it will lose. We visualize this phenomenon in Appendix~\ref{sec:f4}. The possible explanation is that a bomb will not double the points so that playing a bomb or rocket will not harm WP. Aggressively playing bombs may benefit WP since they will dominate other payers, which allows them to play hand cards freely. In contrast, the agents trained with ADP tend to be very cautious of playing bombs since improperly playing a bomb may double the ADP loss if the agents lose the game in the end.

To better interpret the differences between WP and ADP, we show the results of the agents trained with ADP and WP against the baselines. In Table~\ref{tab:douzeroadp}, we report the results of $\DAI$ trained with ADP using WP as the metric. We observe that the performance is slightly worse than that in Table~\ref{tab:douzerofullwp}. In Table~\ref{tab:douzerowp}, we show the results of $\DAI$ trained with WP using ADP as the metric. Similarly, the ADP result is slightly worse than thate in Table~\ref{tab:douzerofulladp}. We observe similar results if considering the bidding phase (see Table~\ref{tab:douzeroadpbidding} and Table~\ref{tab:douzerowpbidding}). Finally, we launch a head-to-head competition of these two agents in Table~\ref{tab:douzeroheadtohead}. The results again verify that the agents trained with WP are better in terms of WP and vice versa. The above results suggest that WP and ADP are indeed different and encourage different card playing strategies.

In addition to WP and ADP, some other metrics could also be adopted in real-word DouDizhu completions. For example, some apps allow users to double the base score at the beginning of a game. We argue that we should adjust the objectives to achieve the best performance according to different scenarios.

\begin{table}[h!]
    \centering
    \small
    \caption{WP of $\DAI$ against baselines when using ADP as the reward. L: WP of \texttt{A} as Landlord; P: WP of \texttt{A} as Peasants. If the average WP of L and P is higher than 0.5, we conclude that \texttt{A} outperforms \texttt{B} and highlight both L and P in boldface.}.
    \label{tab:douzeroadp}
    \setlength{\tabcolsep}{2.3pt}
    \begin{tabular}{l|cc|cc|cc|cc|cc|cc|cc|cc}
    \toprule
     
\multirow{2}{*}{\diagbox [width=5em,trim=l] {\texttt{A}}{\texttt{B}}} & \multicolumn{2}{c|}{$\DAI$ (ADP)} & \multicolumn{2}{c|}{DeltaDou} & \multicolumn{2}{c|}{SL} & \multicolumn{2}{c|}{RHCP-v2} & \multicolumn{2}{c|}{RHCP} & \multicolumn{2}{c|}{RLCard} & \multicolumn{2}{c|}{CQN} &\multicolumn{2}{c}{Random}\\
\cline{2-17}
& L & P & L & P & L & P & L & P & L & P & L & P & L & P & L & P\\
    \hline
    \midrule
     $\DAI$ (ADP) & .4281 & .5719 & \textbf{.4177} & \textbf{.6319} & \textbf{.5039} & \textbf{.6815} & \textbf{.6615} & \textbf{.7543} & \textbf{.6950} & \textbf{.7628} & \textbf{.8416} & \textbf{.8668} & \textbf{.7198} & \textbf{.8280} & \textbf{.9801} & \textbf{.9895}\\
   
     \bottomrule
    \end{tabular}

\end{table}

\begin{table}[h!]
    \centering
    \small
    \caption{ADP of $\DAI$ against baselines when using WP as the reward. L: ADP of \texttt{A} as Landlord; P: ADP of \texttt{A} as Peasants. If the average ADP of L and P is higher than 0, we conclude that \texttt{A} outperforms \texttt{B} and highlight both L and P in boldface.}.
    \label{tab:douzerowp}
    \setlength{\tabcolsep}{2.3pt}
    \begin{tabular}{l|cc|cc|cc|cc|cc|cc|cc|cc}
    \toprule
     
\multirow{2}{*}{\diagbox [width=5em,trim=l] {\texttt{A}}{\texttt{B}}} & \multicolumn{2}{c|}{$\DAI$ (WP)} & \multicolumn{2}{c|}{DeltaDou} & \multicolumn{2}{c|}{SL} & \multicolumn{2}{c|}{RHCP-v2} & \multicolumn{2}{c|}{RHCP} & \multicolumn{2}{c|}{RLCard} & \multicolumn{2}{c|}{CQN} &\multicolumn{2}{c}{Random}\\
\cline{2-17}
& L & P & L & P & L & P & L & P & L & P & L & P & L & P & L & P\\
    \hline
    \midrule
     $\DAI$ (WP) & -0.411 & 0.411 & \textbf{-0.360} & \textbf{0.664} & \textbf{0.224} & \textbf{1.001} & \textbf{1.252} & \textbf{1.880} & \textbf{1.378} & \textbf{1.794} & \textbf{2.094} & \textbf{2.298} & \textbf{1.418} & \textbf{1.872} & \textbf{2.947} & \textbf{2.518}\\
   
     \bottomrule
    \end{tabular}

\end{table}

\begin{table}[h!]
    \centering
    \small
    \caption{$\DAI$ against DeltaDou and SL when using ADP as reward with bidding network.}
    \label{tab:douzeroadpbidding}
    \begin{tabular}{l|c|c|c}
    \toprule
     
    & $\DAI$ (ADP) & DeltaDou & SL \\
    \midrule
    WP & \textbf{0.535} & 0.477 & 0.407 \\
    ADP & \textbf{0.323} & -0.004 & -0.320\\
   
     \bottomrule
    \end{tabular}

\end{table}

\begin{table}[h!]
    \centering
    \small
    \caption{$\DAI$ against DeltaDou and SL when using WP as reward with bidding network.}
    \label{tab:douzerowpbidding}
    \begin{tabular}{l|c|c|c}
    \toprule
     
    & $\DAI$ (WP) & DeltaDou & SL \\
    \midrule
    WP & \textbf{0.580} & 0.461 & 0.381 \\
    ADP & \textbf{0.315} & 0.075 & -0.390\\
   
     \bottomrule
    \end{tabular}

\end{table}

\begin{table}[h!]
    \centering
    \small
    \caption{Head-to-head comparison between using ADP and WP as objectives. $\DAI$ (ADP) outperforms $\DAI$ (WP) in terms of ADP but is worse than $\DAI$ (WP) in terms of WP. The agents tend to learn different skills with different objectives.}
    \label{tab:douzeroheadtohead}
    \setlength{\tabcolsep}{2.3pt}
    \begin{tabular}{l|cc|cc}
    \toprule
     
  & \multicolumn{2}{c|}{ADP} & \multicolumn{2}{c}{WP} \\
\cline{2-5}
& L & P & L & P\\
    \hline
    \midrule
     $\DAI$ (ADP) vs $\DAI$ (WP) & \textbf{-0.3101} & \textbf{0.4476} & .3617 & .5151 \\
   
     \bottomrule
    \end{tabular}

\end{table}

\subsection{Additional Results of Learning Progress}
\label{sec:D3}

\begin{figure}[H]
  \centering
  \begin{subfigure}[b]{0.40\textwidth}
    \includegraphics[width=1.0\textwidth]{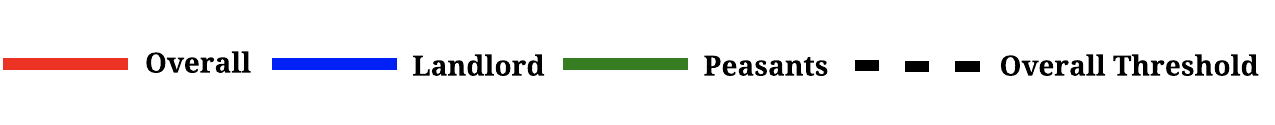}
  \end{subfigure}

  \begin{subfigure}[b]{0.25\textwidth}
    \centering
    \includegraphics[width=0.95\textwidth]{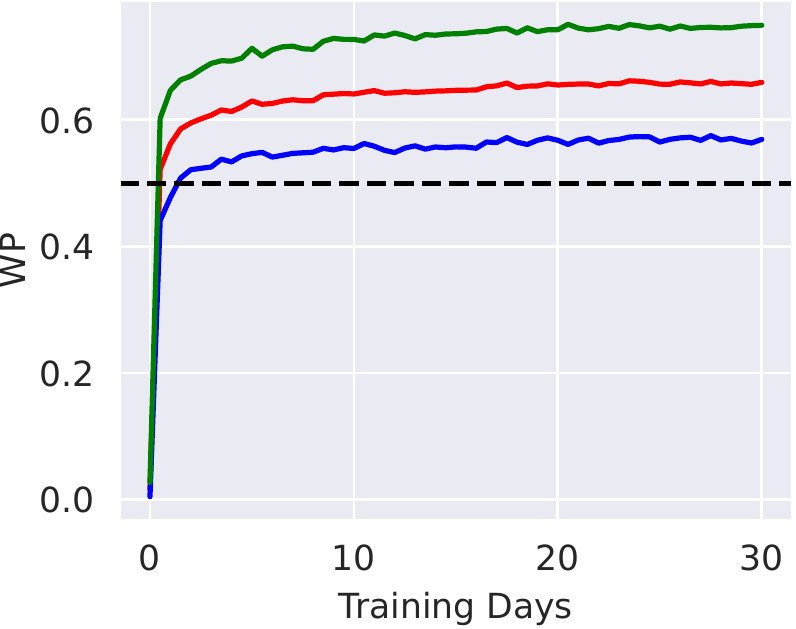}
    \caption{WP against SL}
  \end{subfigure}%
  \begin{subfigure}[b]{0.25\textwidth}
    \centering
    \includegraphics[width=0.95\textwidth]{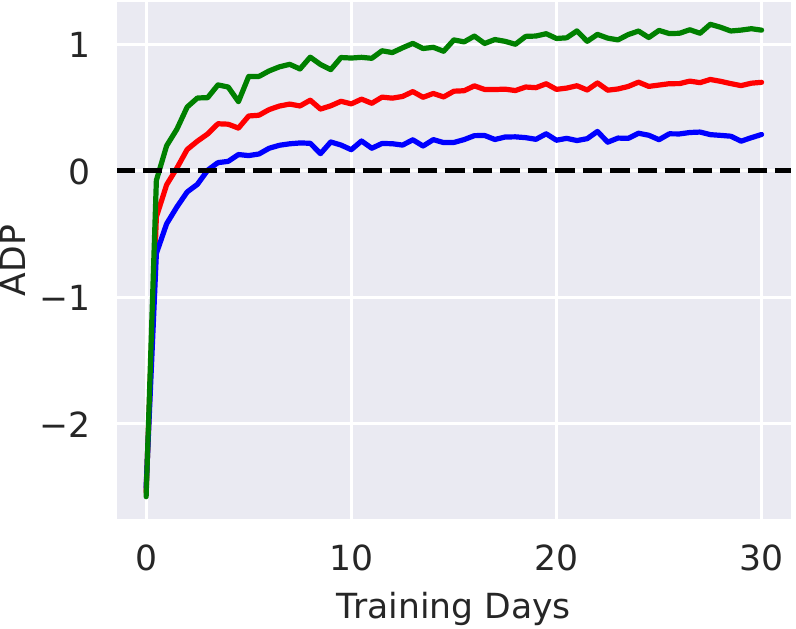}
    \caption{ADP against SL}
  \end{subfigure}%
  \begin{subfigure}[b]{0.25\textwidth}
    \centering
    \includegraphics[width=0.95\textwidth]{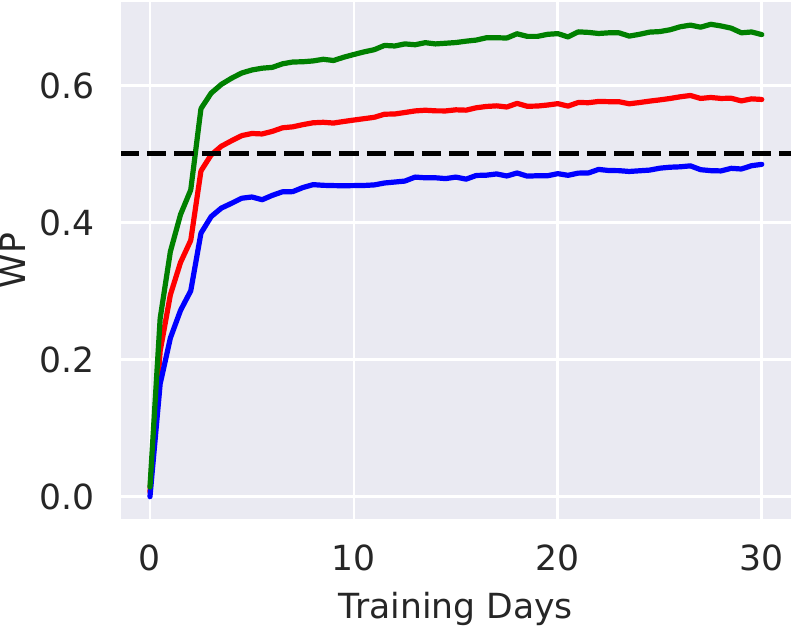}
    \caption{WP against DeltaDou}
  \end{subfigure}%
  \begin{subfigure}[b]{0.25\textwidth}
    \centering
    \includegraphics[width=0.95\textwidth]{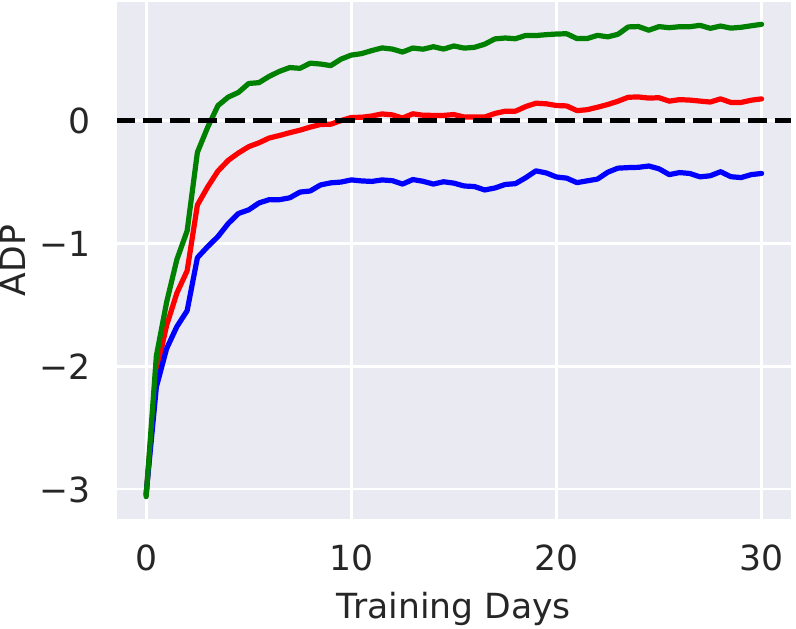}
    \caption{ADP against DeltaDou}
  \end{subfigure}%
  \vspace{-8pt}
  \caption{WP and ADP of $\DAI$ against SL and DeltaDou w.r.t. the number of training days. $\DAI$ outperforms SL with 2 days of training, i.e., the overall WP is larger than the threshold of 0.5 and the overall ADP is larger than the threshold of 0, and surpasses DeltaDou within 10 days, using a single server with four 1080 Ti GPUs and 48 processors.}
\end{figure}

\begin{figure}[H]
  \centering
  \begin{subfigure}[b]{0.40\textwidth}
    \includegraphics[width=1.0\textwidth]{imgs/learning_curves/legend.pdf}
  \end{subfigure}

  \begin{subfigure}[b]{0.25\textwidth}
    \centering
    \includegraphics[width=0.95\textwidth]{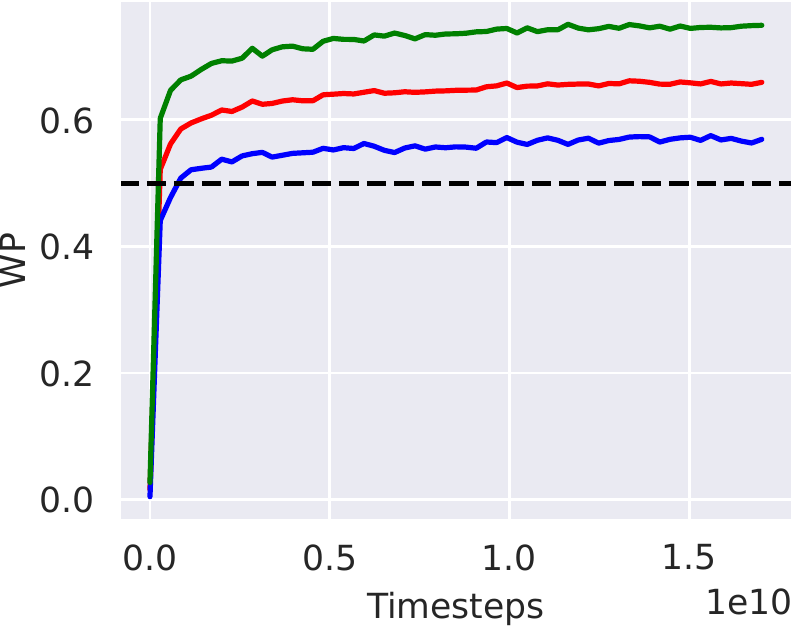}
    \caption{WP against SL}
  \end{subfigure}%
  \begin{subfigure}[b]{0.25\textwidth}
    \centering
    \includegraphics[width=0.95\textwidth]{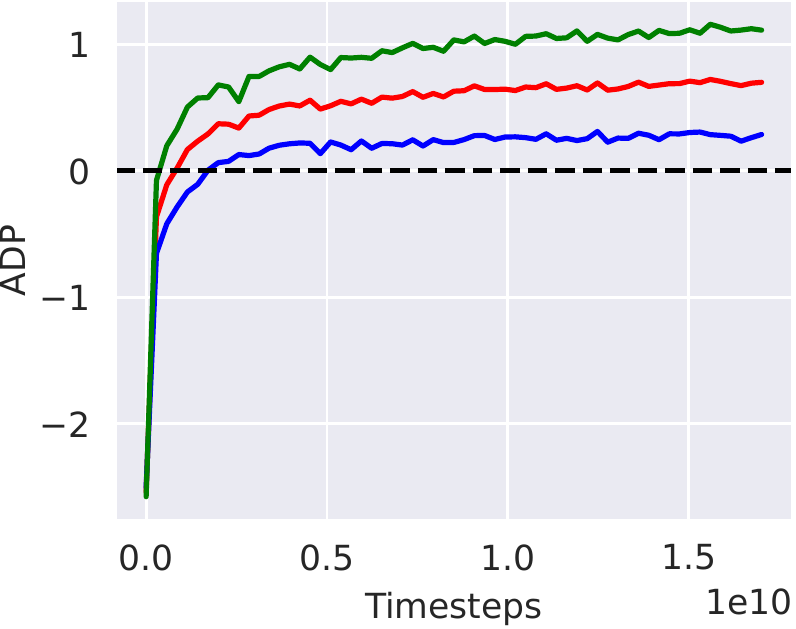}
    \caption{ADP against SL}
  \end{subfigure}%
  \begin{subfigure}[b]{0.25\textwidth}
    \centering
    \includegraphics[width=0.95\textwidth]{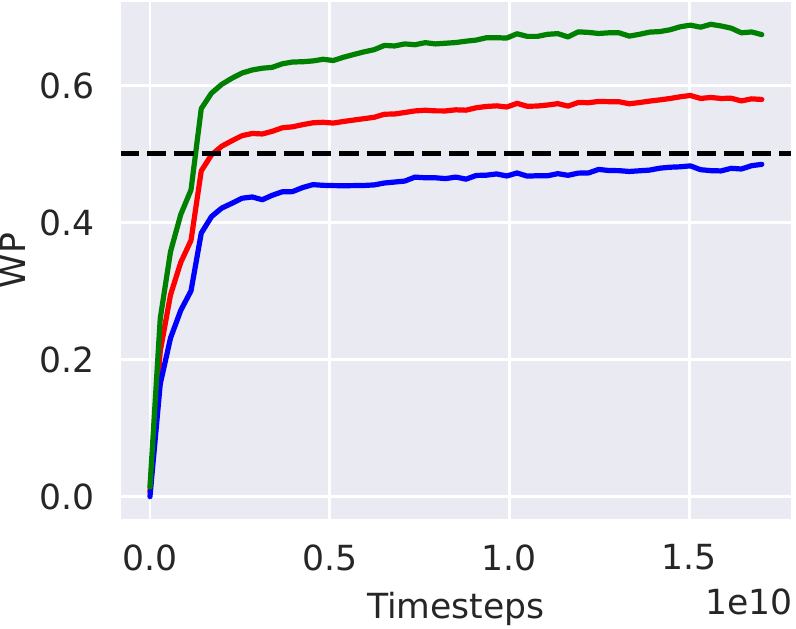}
    \caption{WP against DeltaDou}
  \end{subfigure}%
  \begin{subfigure}[b]{0.25\textwidth}
    \centering
    \includegraphics[width=0.95\textwidth]{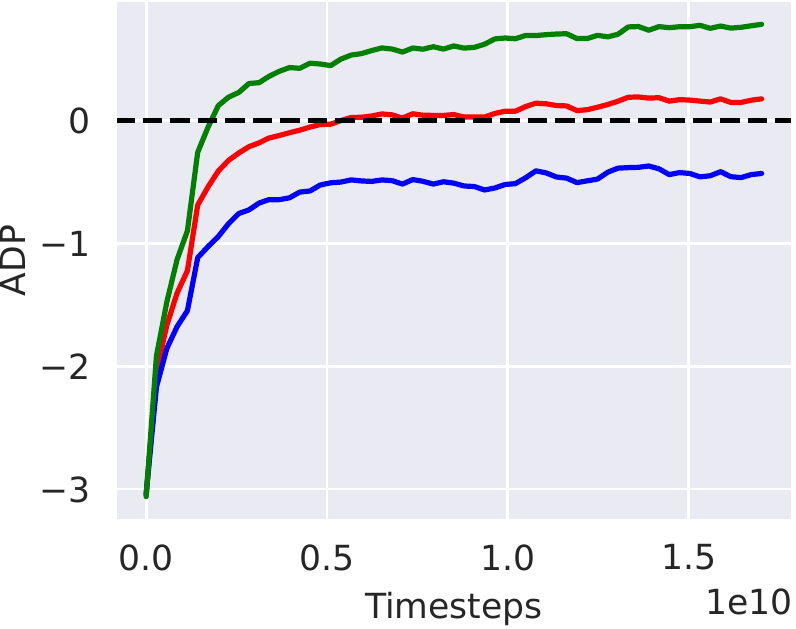}
    \caption{ADP against DeltaDou}
  \end{subfigure}%
  \caption{WP and ADP of $\DAI$ against SL and DeltaDou w.r.t. the number of training timesteps, i.e., the number of actions played by the agent. $\DAI$ outperforms SL with around $5 \times 10^8$ training timesteps, i.e., the overall WP is larger than the threshold of 0.5 and the overall ADP is larger than the threshold of 0, and surpasses DeltaDou within $5 \times 10^9$ training timesteps, using a single server with four 1080 Ti GPUs and 48 processors.}
\end{figure}

\subsection{Full Results of $\DAI$ on Expert Data}
\label{sec:D4}

\begin{figure}[H]
  \centering

  \begin{subfigure}[b]{0.33\textwidth}
    \centering
    \includegraphics[width=0.95\textwidth]{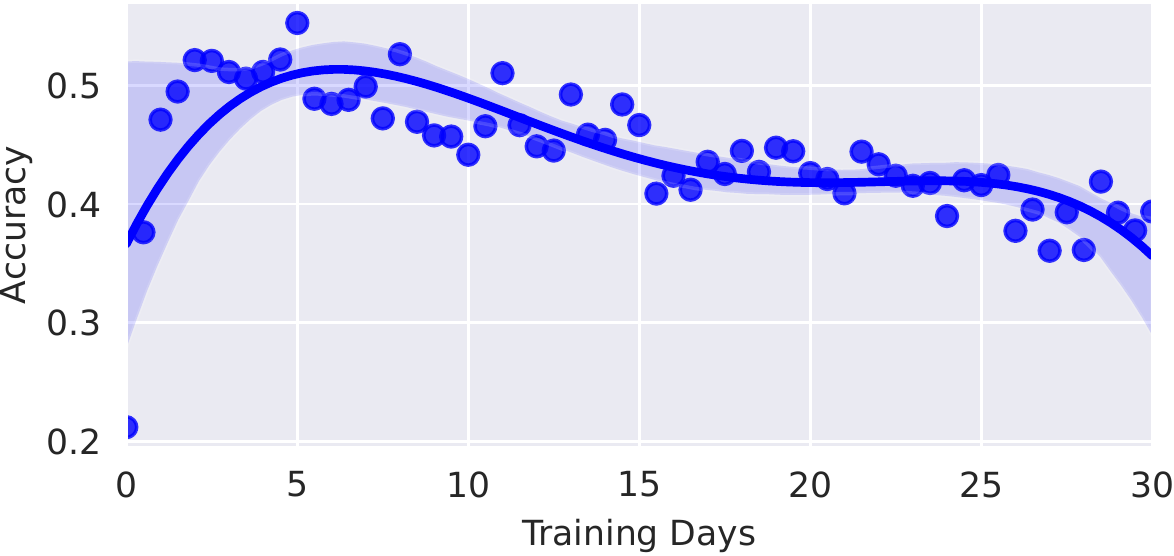}
    \caption{Landlord}
  \end{subfigure}%
  \begin{subfigure}[b]{0.33\textwidth}
    \centering
    \includegraphics[width=0.95\textwidth]{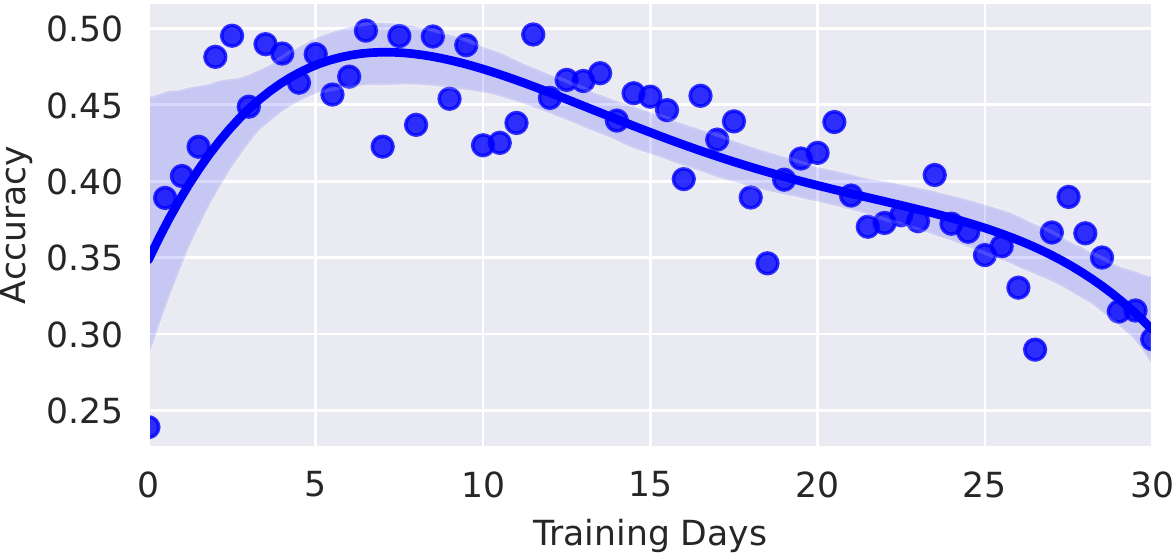}
    \caption{LandlordUp}
  \end{subfigure}%
  \begin{subfigure}[b]{0.33\textwidth}
    \centering
    \includegraphics[width=0.95\textwidth]{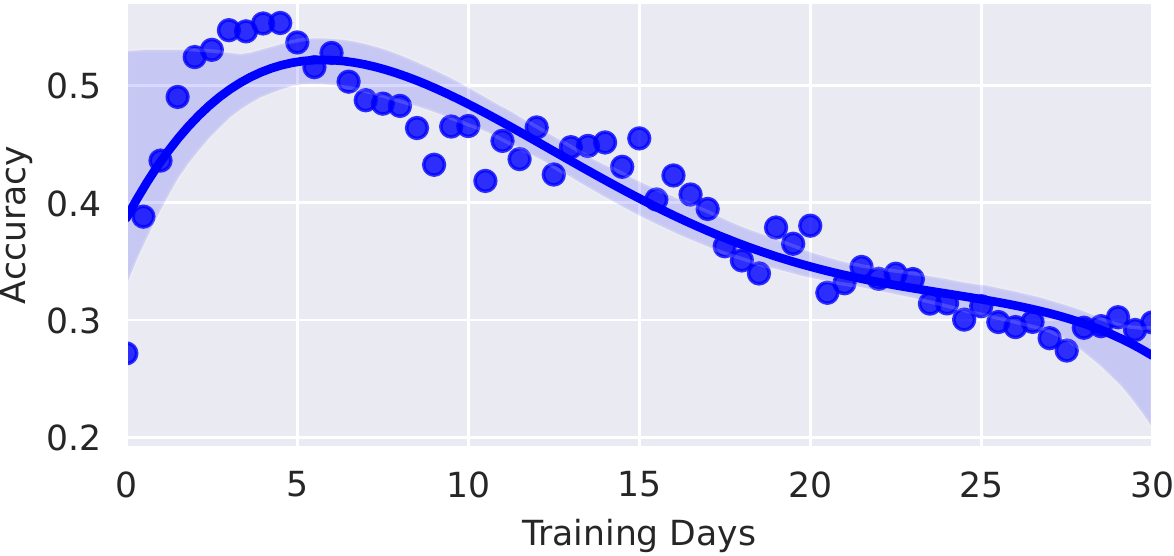}
    \caption{LandlordDown}
  \end{subfigure}%
  \vspace{-5pt}
  \caption{Accuracy for the three positions on the human data w.r.t. the number of training days for $\DAI$. We fit the data points with a polynomial with four terms for better visualizing the trend. LandlordUp stands for the Peasant that moves before the Landlord. LandlordDown stands for the Peasant that moves after the Landlord. The accuracies of SL for the Landlord, LandlordUp, LandlordDown are $83.3\%$, $86.1\%$, $83.1\%$, respectively. $\DAI$ aligns with human expertise in the beginning training stages but discovers novel strategies beyond human knowledge in the later training stages. }
  \label{fig:humancurveall}
\end{figure}

\subsection{Training Curves of $\DAI$}
\label{sec:D5}

In this work, we train three $\DAI$ agents with self-play for all the positions (i.e., the Landlord and the two Peasants) without considering the bidding phase. For each episode, a deck will be randomly generated. Then the three agents will perform self-play with partially-observed states. The generated episodes will be passed to the learner process to update the three $\DAI$ agents. In what follows, we show the self-play rewards and the losses throughout the training process.

\begin{figure}[H]
  \centering
  \begin{subfigure}[b]{0.30\textwidth}
    \includegraphics[width=1.0\textwidth]{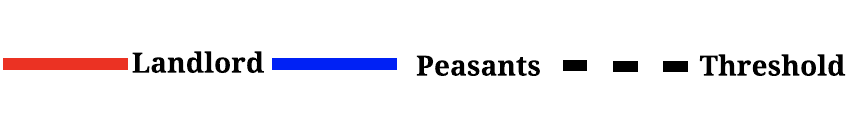}
  \end{subfigure}

  \begin{subfigure}[b]{0.25\textwidth}
    \centering
    \includegraphics[width=0.95\textwidth]{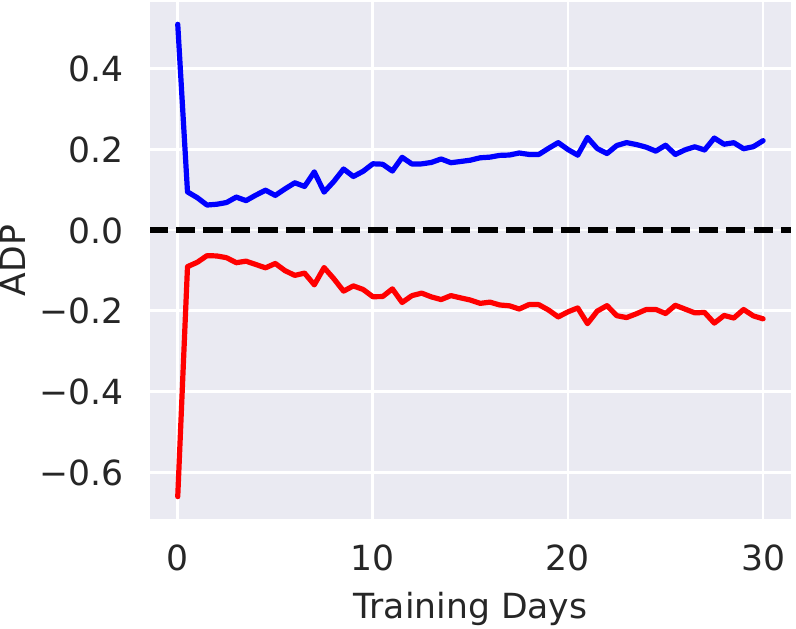}
    \caption{ADP w.r.t. training days}
  \end{subfigure}%
  \begin{subfigure}[b]{0.25\textwidth}
    \centering
    \includegraphics[width=0.95\textwidth]{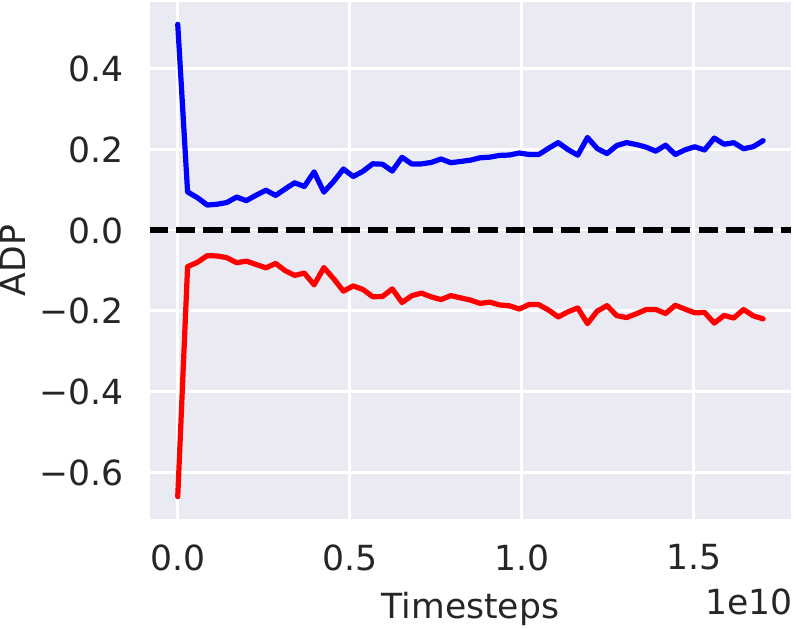}
    \caption{ADP w.r.t. timesteps}
  \end{subfigure}%
  \begin{subfigure}[b]{0.25\textwidth}
    \centering
    \includegraphics[width=0.95\textwidth]{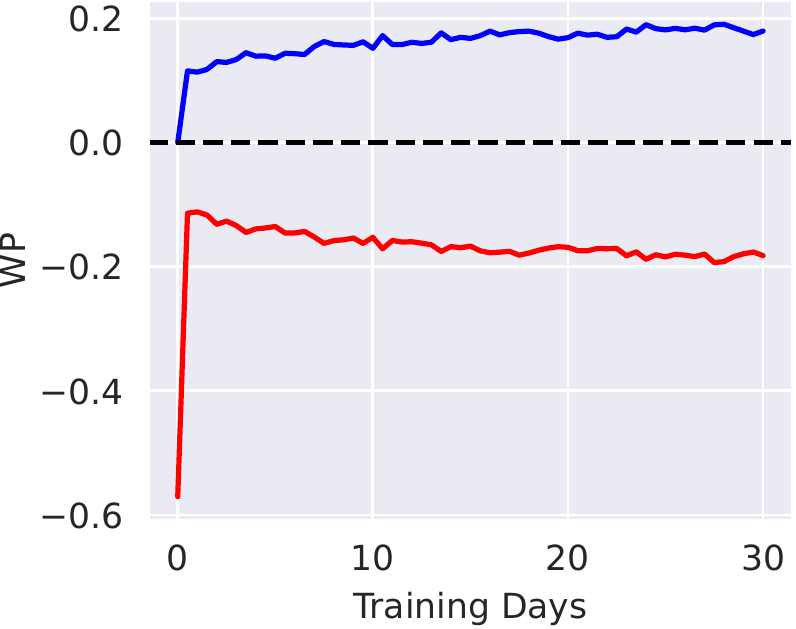}
    \caption{WP w.r.t. training days}
  \end{subfigure}%
  \begin{subfigure}[b]{0.25\textwidth}
    \centering
    \includegraphics[width=0.95\textwidth]{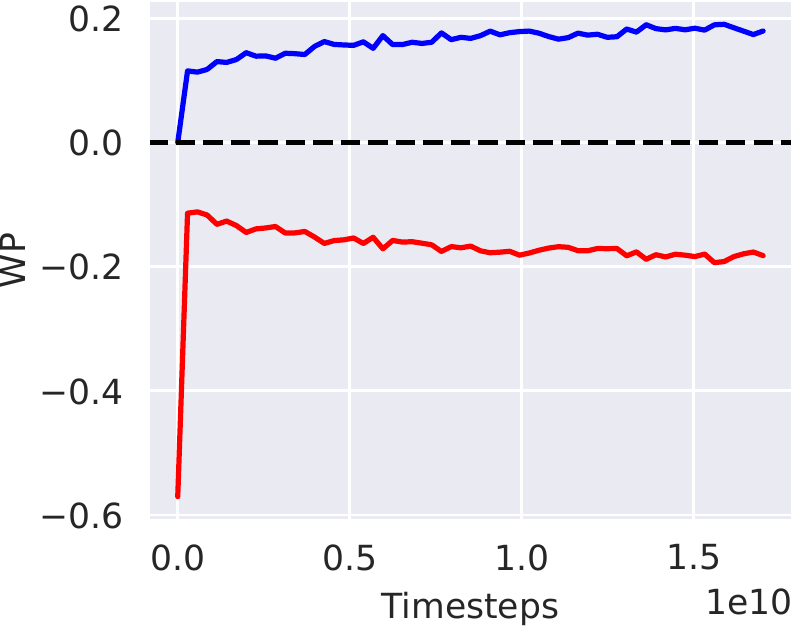}
    \caption{WP w.r.t. timesteps}
  \end{subfigure}%
  \caption{ADP and WP w.r.t. training days and the number of timesteps for the Landlord and Peasants of $\DAI$ during the training progress. At the early stage, the Peasants win the Landlord by a small margin. The peasants become stronger and stronger compared with the Landlord in the later training stages.}
\end{figure}

\begin{figure}[H]
  \centering
  \begin{subfigure}[b]{0.40\textwidth}
    \includegraphics[width=1.0\textwidth]{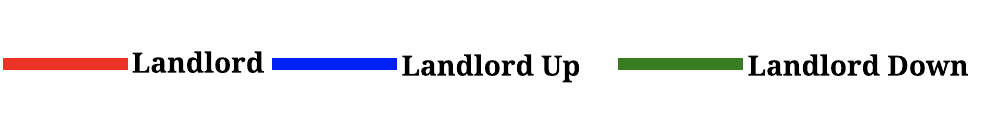}
  \end{subfigure}

  \begin{subfigure}[b]{0.30\textwidth}
    \centering
    \includegraphics[width=0.95\textwidth]{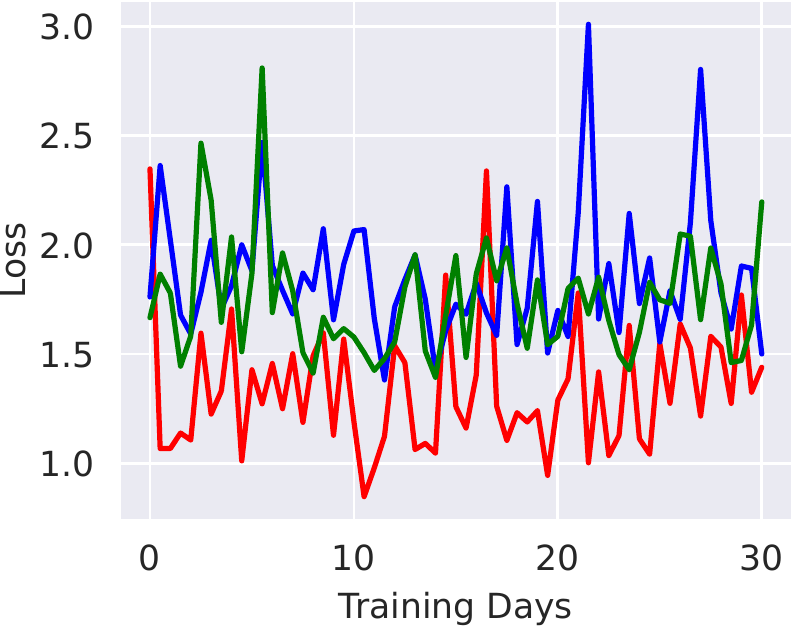}
    \caption{Loss w.r.t. training days}
  \end{subfigure}%
  \begin{subfigure}[b]{0.30\textwidth}
    \centering
    \includegraphics[width=0.95\textwidth]{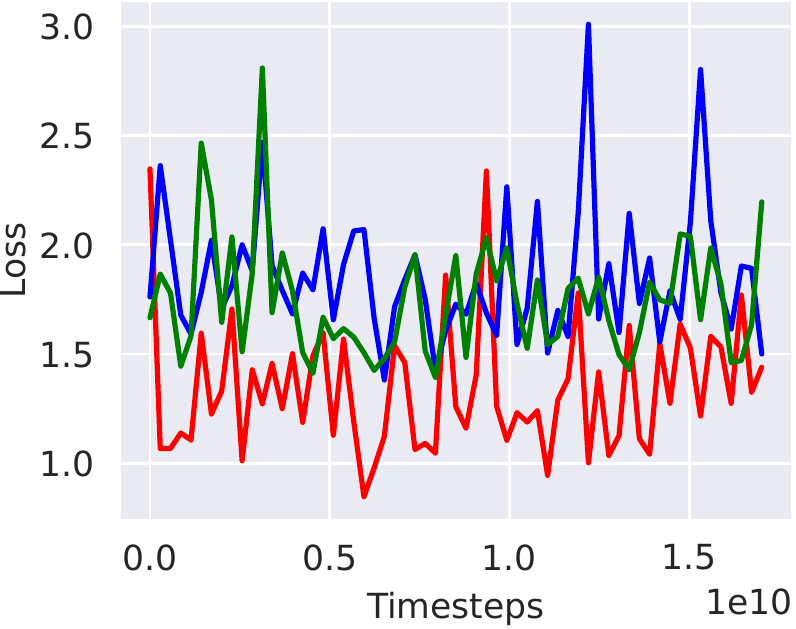}
    \caption{Loss w.r.t. timesteps}
  \end{subfigure}%
  \caption{Losses for different positions for $\DAI$ trained with ADP as rewards. LandlordUp stands for the Peasant that moves before the Landlord. LandlordDown stands for the Peasant that moves after the Landlord. }
\end{figure}

\begin{figure}[H]
  \centering
  \begin{subfigure}[b]{0.40\textwidth}
    \includegraphics[width=1.0\textwidth]{imgs/learning_curves/legend3.pdf}
  \end{subfigure}

  \begin{subfigure}[b]{0.30\textwidth}
    \centering
    \includegraphics[width=0.95\textwidth]{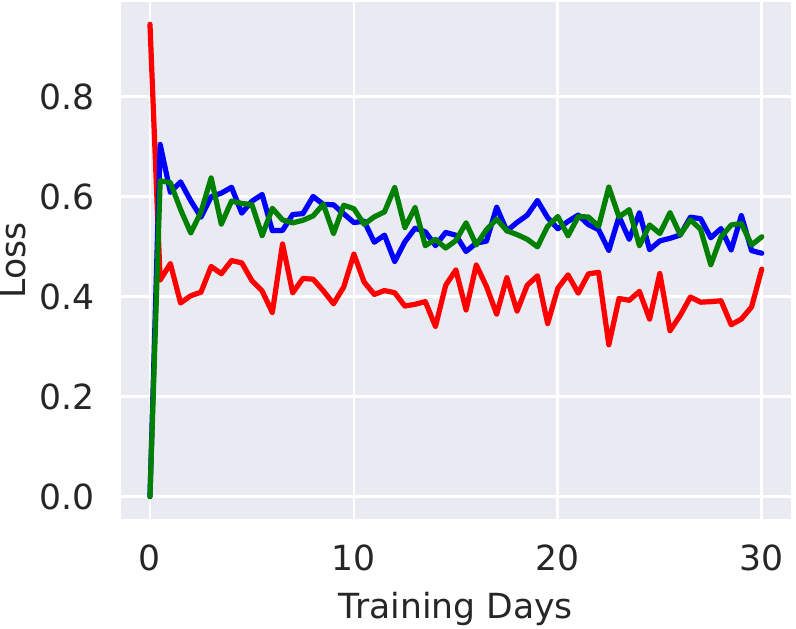}
    \caption{Loss w.r.t. training days}
  \end{subfigure}%
  \begin{subfigure}[b]{0.30\textwidth}
    \centering
    \includegraphics[width=0.95\textwidth]{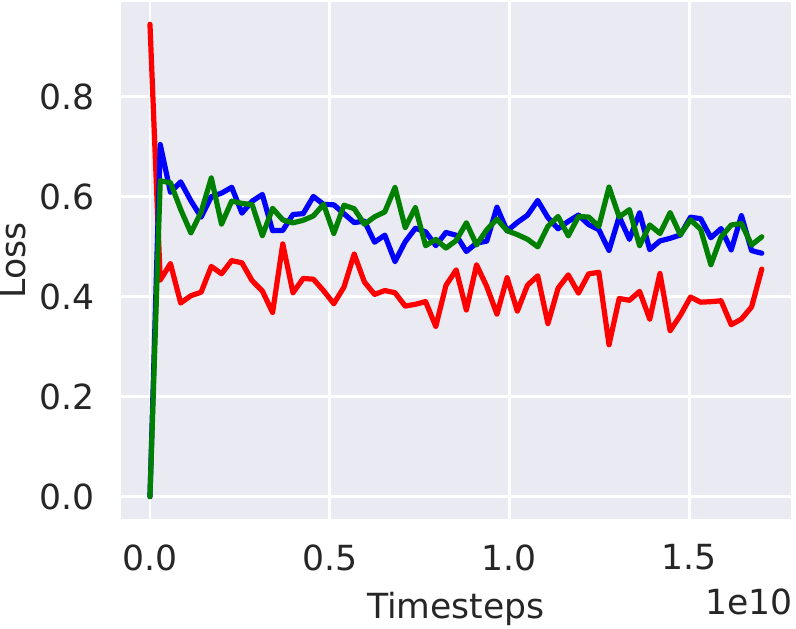}
    \caption{Loss w.r.t. timesteps}
  \end{subfigure}%
  \caption{Losses for different positions for $\DAI$ trained with WP as rewards. LandlordUp stands for the Peasant that moves before the Landlord. LandlordDown stands for the Peasant that moves after the Landlord. }
\end{figure}

\cleardoublepage

\section{More Details of Botzone}
\label{sec:E} 
Botzone is a comprehensive multi-game, multi-agent online game AI platform hosted by AILab, Peking University\footnote{\url{https://wiki.botzone.org.cn/index.php?title=\%E9\%A6\%96\%E9\%A1\%B5/en}}.
Besides DouDizhu, Botzone supports more than 20 games, including Go, Mahjong, and Ataxx, to name a few. Botzone currently has more than 3,500 users and has hosted various games, from in-class and campus contests within Peking University to nation- and worldwide game AI competitions such as the recent IJCAI 2020 Mahjong AI competition, which attracted top AI researchers worldwide.

On the Botzone platform, users upload their bot program as a virtual agent to compete with other bots in a selected game. In order to do so, a user can either nominate opponents and manually start a game or add their bots to the Botzone Elo system, where games among bots will be scheduled automatically. A bot added to the Botzone Elo system will also be associated with the so-called Elo rating score and will be added to the rank list (leaderboard) of the game she plays. 
Botzone assigns an initial Elo rating score of $1000$ to a new bot in the Elo, and the score is updated after each time the bot played an Elo rating game, as long as the bot remains in the Elo system.

\subsection{Interacting with Botzone}
Botzone provides a ``Judge" program that runs in the background and interacts with bot programs of users. A bot receives a request to act from the ``Judge" every time he has to take an action, i.e., it is her turn to play cards. In DouDizhu, the input sent to the bot contains three parts of information: her own hand cards, the public card, and a sequence of cards played by each player. This is consistent with the incomplete information known to a human player in a common DouDizhu game setting. 

The platform also sets constraints on the file size, memory, and running time of bots. Each decision made by the bot has to be completed within 1 second (or 6 seconds for a Python program) with no more than 256 MB of memory. The model size is limited to 140 MB.

\subsection{Botzone Ranking Rules}
Botzone maintains a leaderboard for each game, which ranks all the bots in the Botzone Elo system by their Elo rating scores in descending order. Every five minutes, Botzone schedules match with randomly selected bots, with priority given to new bots with recent updates. The Elo rating score of participating bots will be updated after the match. 

In the Botzone Elo of DouDizhu (named ``FightTheLandlord" on the Botzone platform), each game is played by two bots, with one bot acting as the Landlord and the other as Peasants. A pair of games are played simultaneously, in which the two bots will play different roles; that is, the bot who plays the Landlord in one game will play the Peasants in another, and vice versa. The two games form a match, and the Elo rating of each bot is updated according to the match outcome, as well as their relative ratings.

The game score of a bot is mainly determined by whether she wins or loses a game. The winning bot receives a score of two points, while the losing bot receives a score of zero\footnote{\url{https://wiki.botzone.org.cn/index.php?title=FightTheLandlord}}. To encourage more complicated card-playing strategies, Botzone associates small points with categories of the cards played by a bot and adds that to the game score. 
The total game score is thus the \textit{winning point} (2 for the winner and 0 for the loser) plus the \textit{card-playing advantage}, the sum of weights assigned according to categories of cards (Table \ref{tab:sumweightsbotzone}) played by a bot in the entire game divided by 100. Since by convention, two Peasants always receive the same game score, Peasants' game score is the average of their individual scores.

The match score is determined by the sum of game scores of the two games in the round, which further determines how the Elo rating will change for each player. If the match score of one bot is higher than the other, then the bot is considered the winner of this match. The winning bot will receive an increase in its Elo rating, while the same amount of rating points will be taken off from the losing bot. 


\vspace{-5pt}
\begin{table}[h!]
    \centering
    \caption{Summary of weights assigned to each category in Botzone.}
    \label{tab:sumweightsbotzone}
    \begin{tabular}{l|l}
    \toprule
    \textbf{Action Type} & \textbf{Weight} \\
    \midrule
     Solo    &  $1$\\
     Pair    &  $2$\\
     Trio (or with Solo / Pair)    &  $4$\\
     Chain of Solo & $6$\\
     Chain of Pair & $6$\\
     Chain of Trio & $8$\\
     Plane with Solo / Pair & $8$\\
     Quad with Solo / Pair & $8$\\
     Space shuttle A (2 consecutive Quad) & $10$\\
     Bomb & $10$\\
     Rocket & $16$\\
     Space shuttle B (more than 3 consecutive Quads) & $20$\\
     Pass & $0$\\
     \bottomrule
    \end{tabular}
\end{table}

\subsection{Discussion of Ranking Stability}
Although Elo rating is generally considered a stable measurement of relative strength among a pool of players in games like Chess and Go, DouDizhu Elo ranking on Botzone suffers from some fluidity. This could be attributed to the nature of the high variance of the game and also the design of Botzone Elo. Firstly, the game outcomes of DouDizhu relies on the luck of initial hand cards. In particular, if a player with bad hand cards is playing the Landlord, he will have a low chance to win a game. In practice, the bidding phase could compensate for this randomness, such that the player with bad initial hand cards can choose not to bid for the Landlord. However, Botzone does not incorporate the bidding phase into the game playing; rather, the Landlord position is specified even before the dealing phase happened. Secondly, although two bots exchange roles between games in each match, these two games are not initialized with the same hand cards. It is not rare to see that one bot was assigned bad initial hand cards in both games, making it infeasible for her to win the match. Finally, Elo rating games are not scheduled as frequently on Botzone, potentially due to limited server resources. We observe that, on average, $\DAI$ has the chance to play one Elo rating game about every 2 hours. As such, it might take a long time for a bot to achieve a stable ranking. With bots continuously added to or leaving the Elo system, it might be just impossible to observe absolute stable ranking. Nonetheless, since ranked top on Botzone for the first time on October 30, 2020, $\DAI$ has remained in the top-5 most of the time (at the time of submission deadline, $\DAI$ was still ranked first with around 1600 points). While the rank of $\DAI$ is impacted by the high variance of the BotZone platform, $\DAI$ has maintained an Elo rating score of at least 1480 points during the months between October 30, 2020, to the ICML submission deadline, suggesting that $\DAI$ has at least $95\%$ chance of winning in a match with an average bot.


%

\cleardoublepage

\section{Additional Case Studies}
\label{sec:F}
In this section, we conduct case studies for \DAI. We show both per-step decision with figures and the logs of full games. For simplicity, we use ``T" to denote ``10",``P" to denote ``PASS", ``B" to denote Black Joker, and ``R" to denote Red Joker. Each move is represented as ``position:move", where position can be ``L" for Landlord, ``U" for LandlordUp (i.e., the Peasant that moves before the Landlord), ``D" for LandlordDown (i.e., the Peasant that moves after the Landlord). For example, ''L:3555" means Landlord plays 3555, and ``U:T" means LandlordUp plays 10. The initial hands are represented as ``H:Landlord Hand; LandlordDown Hand; LandlordUp Hand". The moves and the hands are separated by ``,". \textbf{Note that except Section~\ref{sec:f4}, we focus on the agent with WP as objective. Thus, the agents tend to ambitiously play bombs even when they will lose, and will not try to play more bombs when they think they will win.}

\subsection{Strategic Thinking of the Agents}
\begin{figure}[H]
  \centering
    \includegraphics[width=0.5\textwidth]{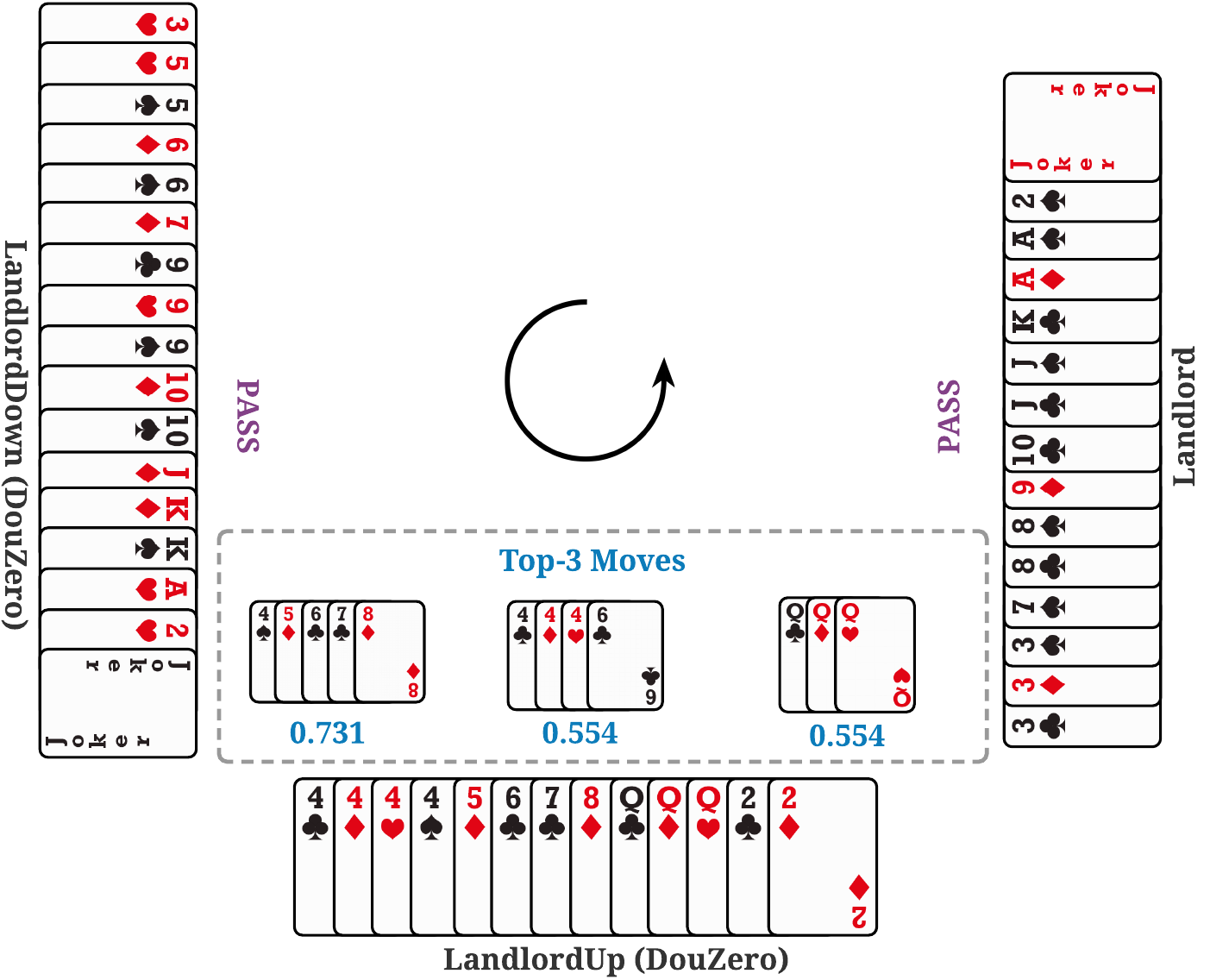}
  \vspace{-10pt}
  \caption{Case 1: Strategic thinking (turn 7). The LandlordUp has 4444 in hand. However, $\DAI$ strategically chooses to break the bomb and play 45678. This is because playing this Chain of Solo can empty the hand more quickly. \textbf{Full game:} H:333456778889TJJKAA2R; 355667999TTJKKA2B; 4445678TJQQQQKA22, L:45678, D:P, U:TJQKA, L:P, D:P, U:45678, L:789TJ, D:P, U:P, L:3338, D:999J, U:4QQQ, L:P, D:P, U:22, L:P, D:P, U:4.}
\end{figure}

\begin{figure}[H]
  \centering
    \includegraphics[width=0.5\textwidth]{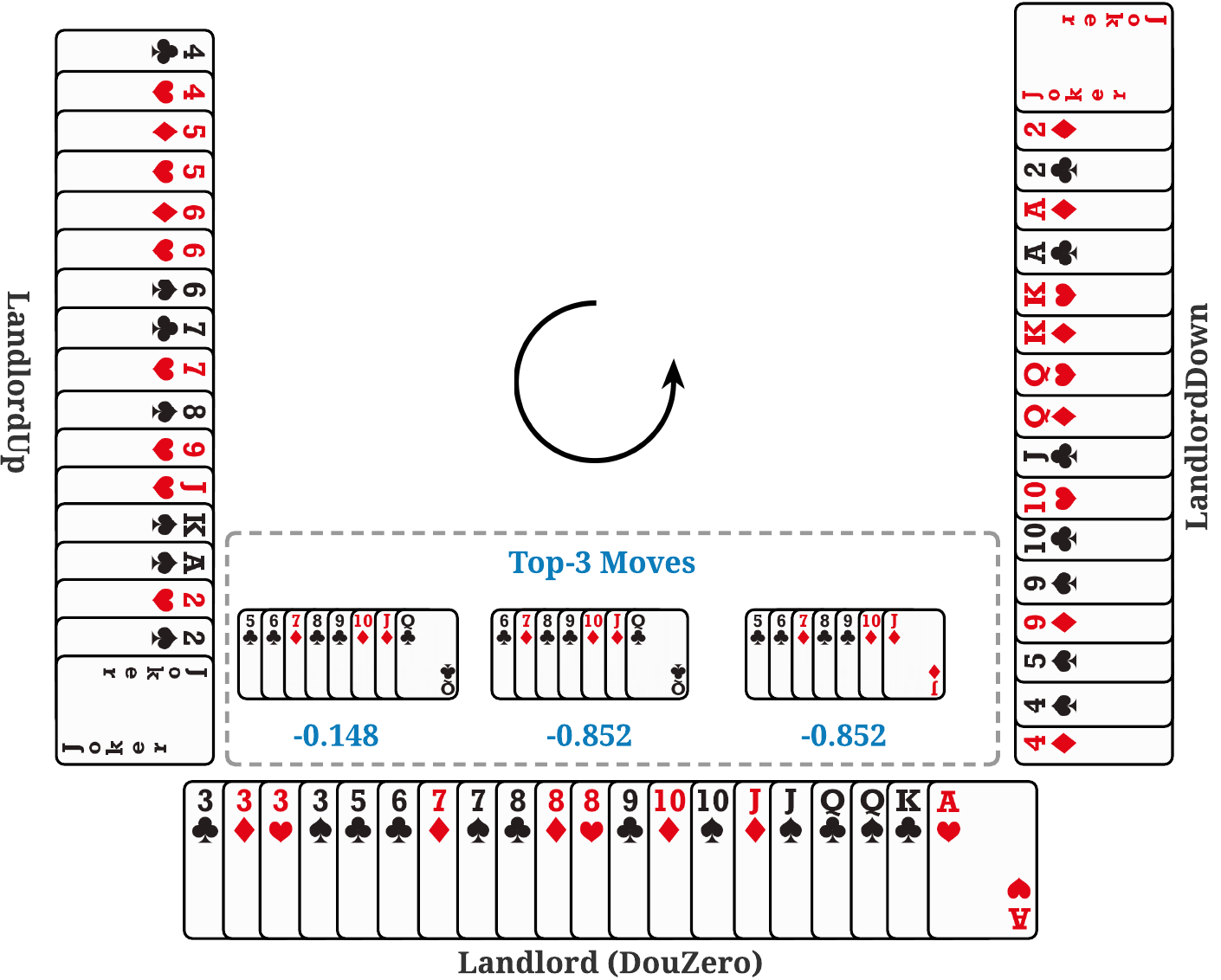}
  \vspace{-10pt}
  \caption{Case 2: Strategic thinking (turn 1). In this case, 56789TJQ is a very good move because the agent can play another Chain of Solo afterwards, i.e., TJQKA. \textbf{Full game:} H:333356778889TTJJQQKA; 44599TTJQQKKAA22R; 44556667789JKA22B, L:56789TJQ, D:P, U:P, L:TJQKA, D:P, U:P, L:7, D:J, U:K, L:P, D:P, U:44, L:88, D:99, U:22, L:3333.}
\end{figure}

\begin{figure}[t!]
  \centering
    \includegraphics[width=0.5\textwidth]{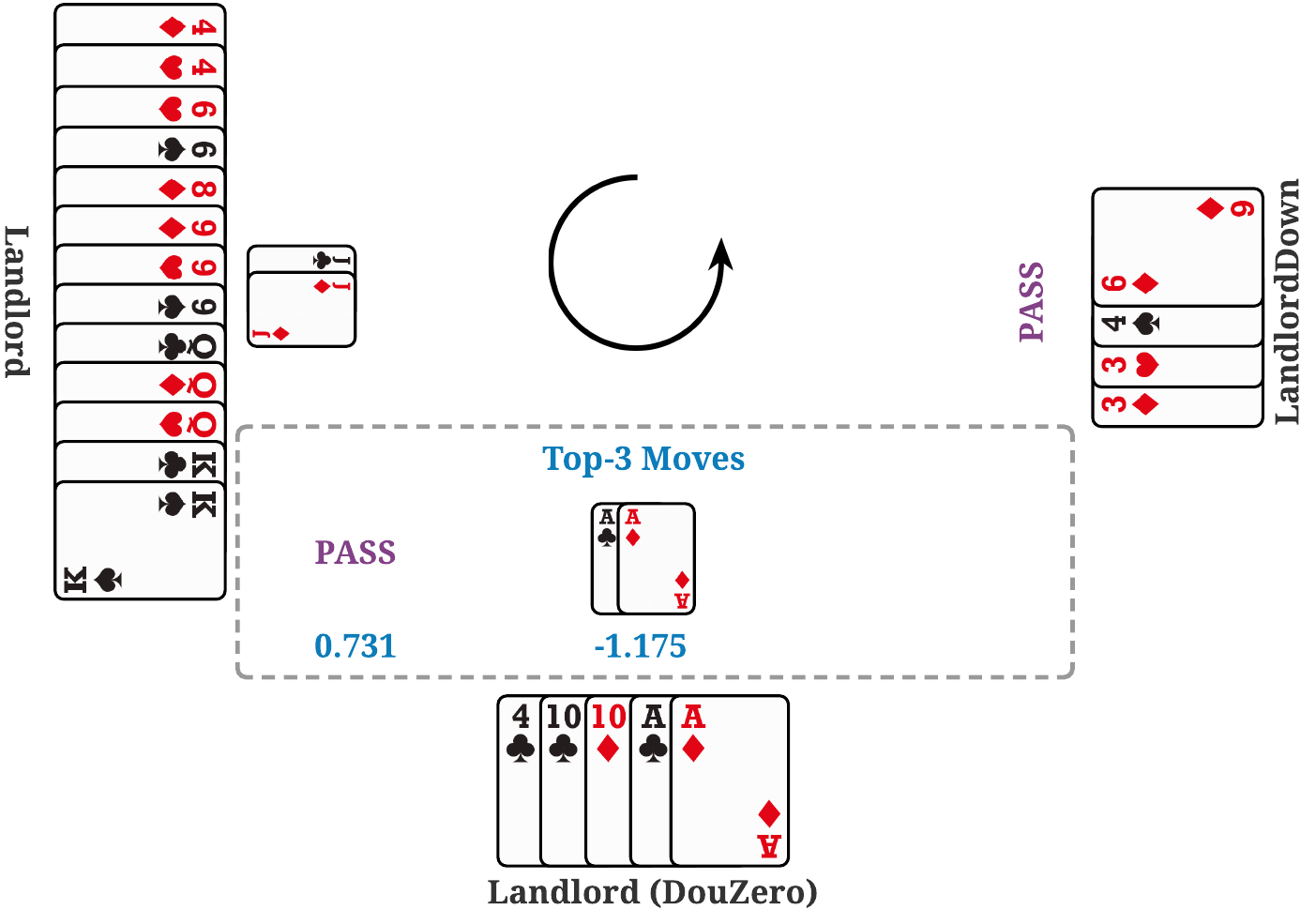}
  \vspace{-5pt}
  \caption{Case 3: Strategic thinking (turn 31). The Landlord has two pairs (AA and TT) and a small Solo in hand. $\DAI$ chooses not to play AA in this turn. This is a nice move because there is a pair of K out there. If the Landlord plays AA, then the TT will be dominated by the KK in later turns. $\DAI$ patiently chooses PASS and wins the game eventually. \textbf{Full game:} H:33455556889TTJAA22BR; 334677778TTJQKA22; 44668999JJQQQKKKA, L:33, D:22, U:P, L:BR, D:P, U:P, L:6, D:8, U:A, L:2, D:7777, U:P, L:P, D:TJQKA, U:P, L:5555, D:P, U:P, L:9, D:T, U:P, L:J, D:P, U:K, L:2, D:P, U:P, L:88, D:P, U:JJ, L:P, D:P, U:44999, L:P, D:P, U:66QQQ, L:P, D:P, U:KK, L:AA, D:P, U:P, L:TT, D:P, U:P, L:4.}
\end{figure}

\begin{figure}[H]
  \centering
    \includegraphics[width=0.5\textwidth]{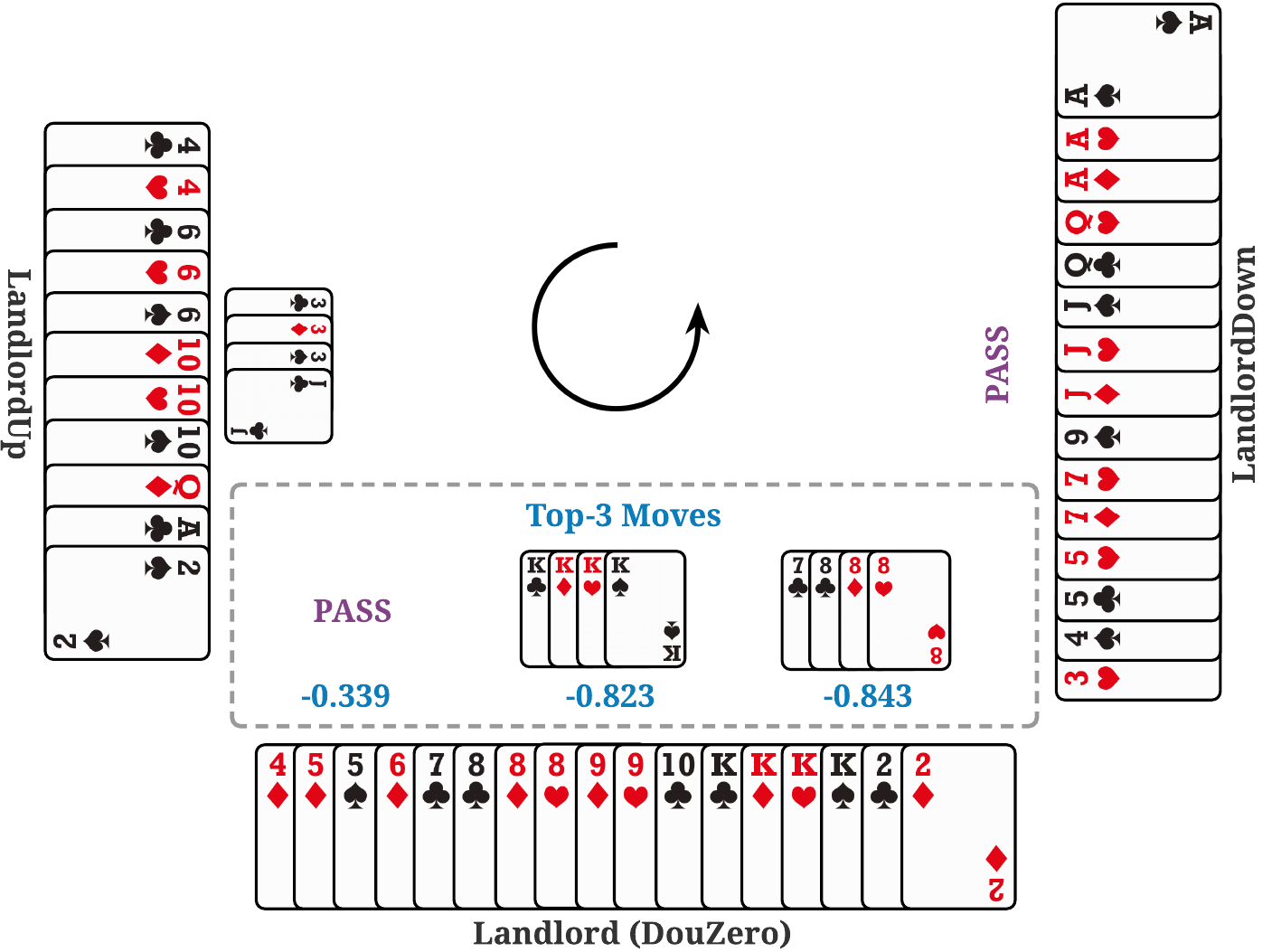}
  \vspace{-5pt}
  \caption{Case 4: Strategic thinking (turn 13). When the LandlordUp plays 333J, $\DAI$ chooses not to play 888. This is a nice move because playing 888 will break a Chain of Solo, i.e., 456789T. \textbf{Full game:} H:45567788899TQKKKK22B; 34557789JJJQQAAA2; 333446669TTTJQA2R, L:7, D:8, U:9, L:Q, D:2, U:P, L:B, D:P, U:R, L:P, D:P, U:333J, L:P, D:P, U:666Q, L:P, D:P, U:TTTA, L:P, D:P, U:44, L:55, D:77, U:P, L:88, D:QQ, U:P, L:22, D:P, U:P, L:49KKKK, D:P, U:P, L:6789T.}
\end{figure}

\newpage

\begin{figure}[H]
  \centering
    \includegraphics[width=0.5\textwidth]{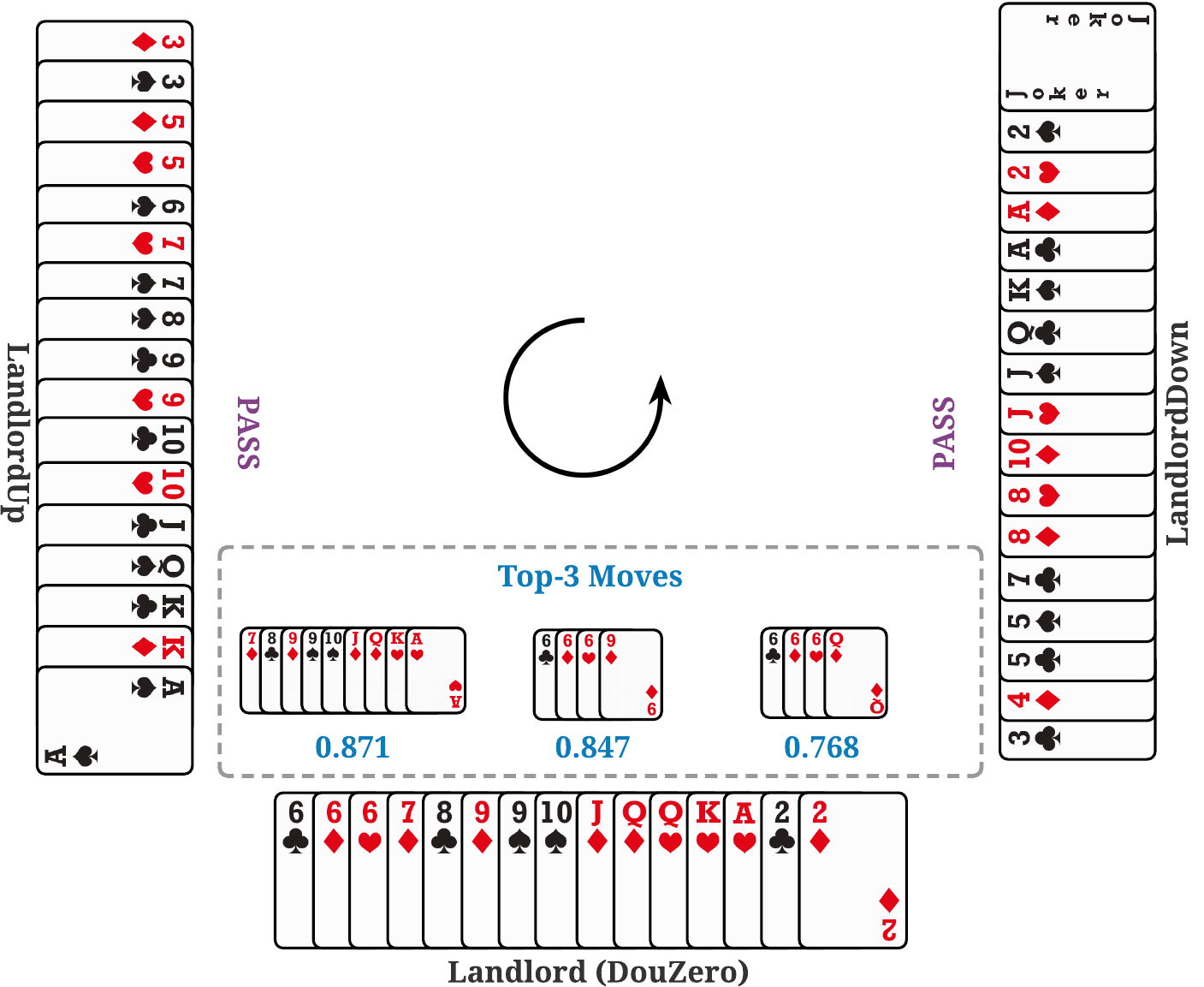}
  \vspace{-5pt}
  \caption{Case 5: Strategic thinking (turn 4). $\DAI$ can choose a move from many possible legal moves. The top-1 move is nice because it is a very long Chain of Solo. The second and the third moves are also very good because they choose a good kicker. \textbf{Full game:} H:34446667899TJQQKA22R; 3455788TJJQKAA22B; 3355677899TTJQKKA, L:3444, D:P, U:P, L:789TJQKA, D:P, U:P, L:6669, D:P, U:P, L:22, D:P, U:P, L:Q, D:B, U:P, L:R.}
\end{figure}

\subsection{Cooperation of Peasants}

\begin{figure}[H]
  \centering
    \includegraphics[width=0.5\textwidth]{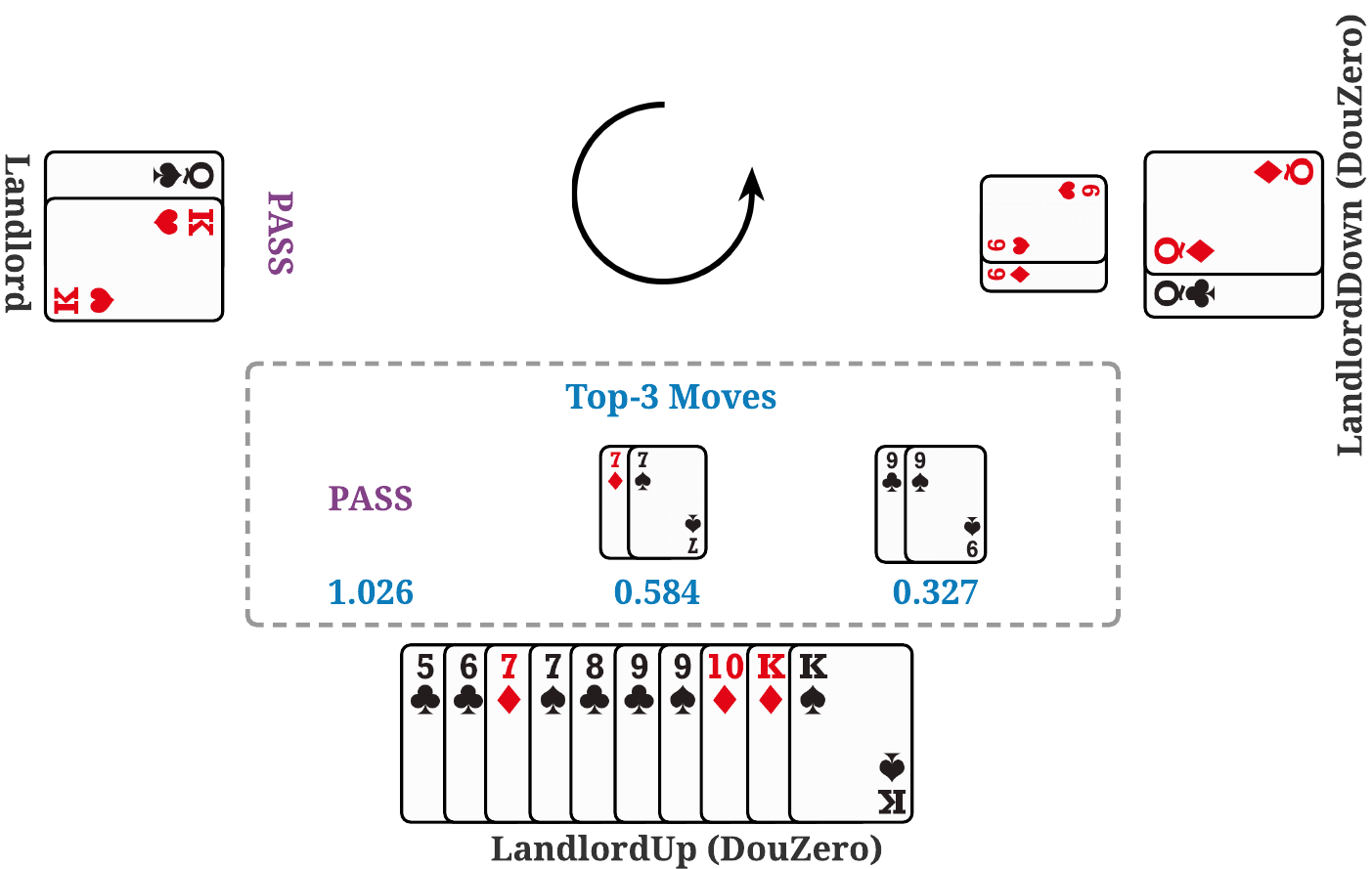}
  \vspace{-5pt}
  \caption{Case 1: Cooperation of Peasants (turn 29). Although the Landlord has much better hand than the Peasants (a Bomb plus a Rocket), the Peasants manage to win the game with cooperation. In this turn, the LandlordUp chooses PASS so that the LandlordDown can empty her hand. $\DAI$ has learned not to fight against the teammate. \textbf{Full game:} H:3335556799JJJJQK22BR; 44445677899TQKKAA; 3667888TTTQQKAA22, L:3336, D:P, U:3888, L:JJJJ, D:P, U:P, L:5557, D:P, U:7TTT, L:BR, D:P, U:P, L:99, D:AA, U:22, L:P, D:P, U:AA, L:22, D:4444, U:P, L:P, D:Q, U:K, L:P, D:P, U:66, L:P, D:P, U:QQ.}
\end{figure}

\newpage

\begin{figure}[H]
  \centering
    \includegraphics[width=0.5\textwidth]{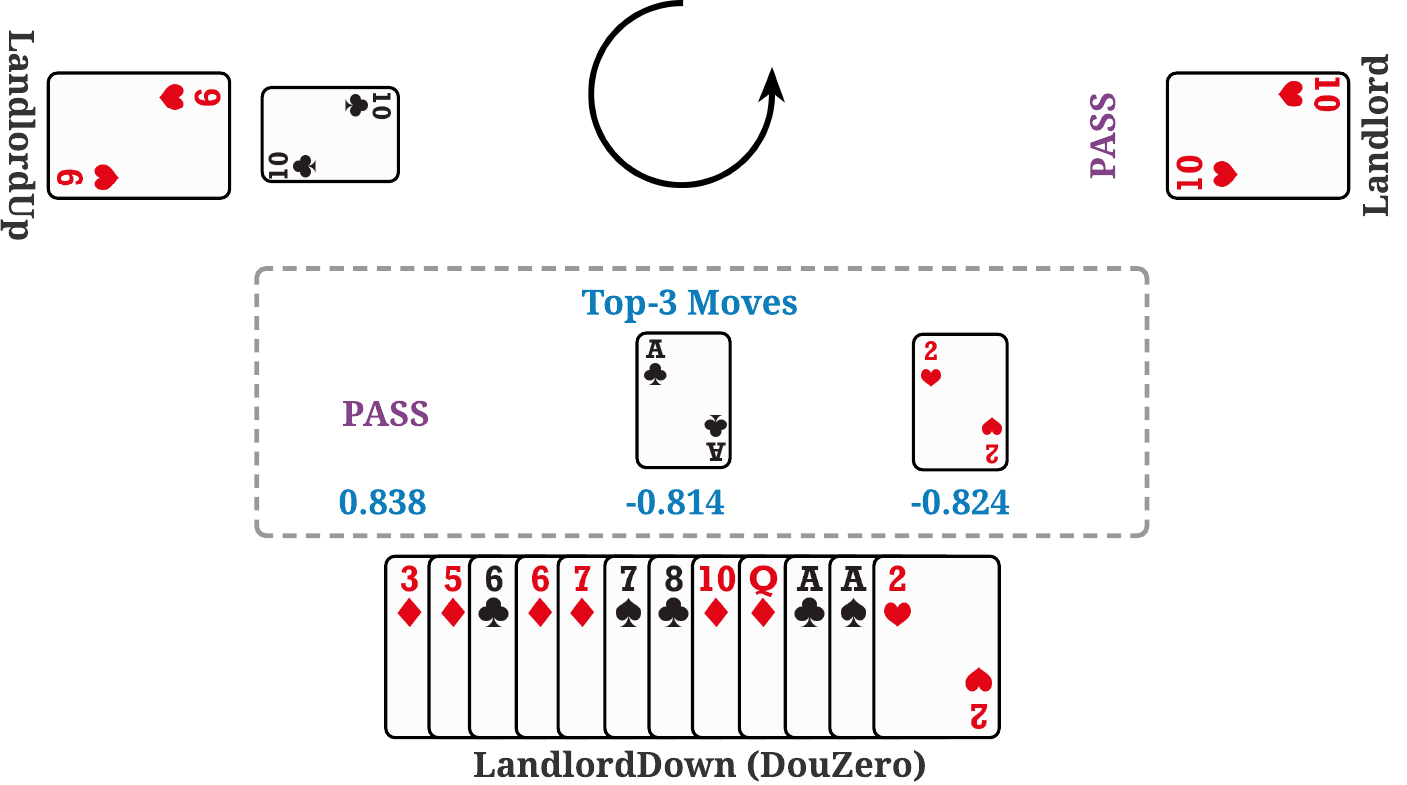}
  \vspace{-5pt}
  \caption{Case 2: Cooperation of Peasants (turn 33). The LandlordUp plays a T. $\DAI$ chooses PASS because there is no card out there larger than T in rank. This suggests that $\DAI$ has learned to reason about the cards that have not been played. \textbf{Full game:} H:334444556689TTJJJQ2R; 3577889TQQKKKKAA2; 356677899TJQAA22B, L:33444455, D:7788KKKK;, U:P, L:P, D:3, U:J, L:2, D:P, U:B, L:R, D:P, U:P, L:89TJQ, D:P< U:P, L:66, D:QQ, U:P, L:P, D:5, U:2, L:P, D:P, U:99, L:JJ, D:AA, D:P, L:P, D:2, U:P, L:P, D:T, U:P, L:P, D:9.}
\end{figure}

\begin{figure}[H]
  \centering
    \includegraphics[width=0.5\textwidth]{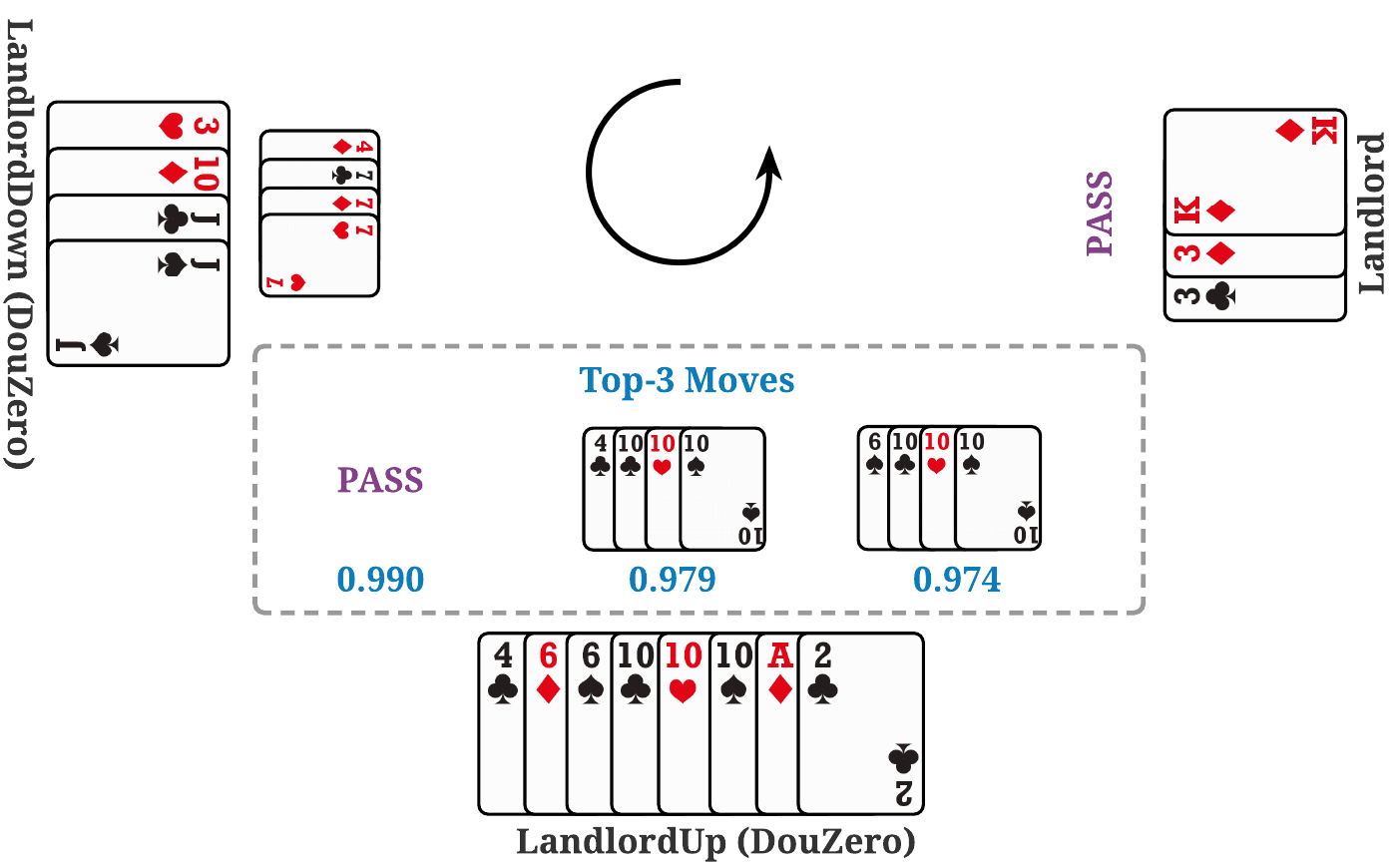}
  \vspace{-5pt}
  \caption{Case 3: Cooperation of Peasants (turn 36). The LandlordDown plays a Trio with Solo. While LandlordUp has a very good hand and can win the game by itself, $\DAI$ chooses PASS to let her teammate win. \textbf{Full game:} H:3344566788999JQK222; 34577788TJJQQQKAA; 3455669TTTKKAA2BR; L:45678, D:P, U:P, L:4999, D:5QQQ, U:P, L:6222, D:P, U:BR, L:P, D:P, U:55, L:JJ, D:P, U:KK, L:P, D:P, U:9, L:Q, D:K, U:A, L:P, D:P, U:3, L:8, D:A, U:P, L:P, D:88, U:P, L:P, D:A, U:P, L:P, D:4777, U:P, L:P, D:3, U:T, L:K, D:P, U:A, L:P, D:P, U:T, L:P, D:P, U:T, L:P, D:J, U:P, L:P, D:J, U:P, L:P, D:T.}
\end{figure}

\subsection{Comparison of the Models in Early Stage and Later Stage}
\begin{table}[H]
    \centering
    \caption{Case 1: Comparison of $\DAI$ in early stage and later stage. $\DAI$ plays the Landlord position on the same deck with the same opponents. $\DAI$ loses in the early stage but wins the game in the later stage. $\DAI$ in the later stage tends to perform a better planning of the hand. Specifically, $\DAI$ in the later stage tends to first play small pairs, such as 88 and TT, so that it can easily empty the hand later.}
    \begin{tabular}{l|l}
    \toprule
     & Logs  \\
    \midrule
     Early stage & \makecell[l]{H:455557777889TTKKAA22; 3446668999QQKA2BR; 333468TTJJJJQQKA2, L:88, D:QQ, U:P, L:KK,\\ D:P, U:P, L:455559, D:P, U:46JJJJ, L:P, D:P, U:3338, L:P, D:3666, U:P, L:P, D:44999, U:P, L:P, D:8, U:K,\\ L:A, D:2, U:P, L:P, D:K, U:P, L:A, D:BR, U:P, L:P, D:A}  \\
     \midrule
     Later stage & \makecell[l]{H:455557777889TTKKAA22; 3446668999QQKA2BR; 333468TTJJJJQQKA2, L:TT, D:QQ, U:P, L:KK,\\ D:P, U:P, L:88, D:99, U:TT, L:AA, D:P, U:P, L:477779, D:P, U:46JJJJ, L:P, D:P, U:3338, L:5555, D:BR,\\ U:P, L:P, D:3666, U:P, L:P, D:8, U:K, L:2, D:P, U:P, L:2} \\
     \bottomrule 
    \end{tabular}

\end{table}

\newpage

\begin{table}[t]
    \centering
    \caption{Case 2: Comparison of $\DAI$ in early stage and later stage. $\DAI$ plays the Landlord position on the same deck with the same opponents. In the early stage, $\DAI$ keeps playing PASS because it does not want to break the Rocket. As result, $\DAI$ loses the game. In the later stage, $\DAI$ smartly breaks the Rocket to dominate other Solos, and eventually wins the game.}
    \begin{tabular}{l|l}
    \toprule
     & Logs  \\
    \midrule
     Early stage & \makecell[l]{H:3355556778889JJQKABR; 344446679JJQQKA22; 367899TTTTQKKAA22, L:9, D:K, U:P, L:A, D:2,\\ U:P, L:P, D:3, U:9, L:K, D:A, U:P, L:P, D:4, U:Q, L:P, D:P, U:6789T, L:P, D:P, U:3TTT, L:P, D:P, U:KK,\\ L:P, D:P, U:AA, L:P, D:P, U:22}  \\
     \midrule
     Later stage & \makecell[l]{H:3355556778889JJQKABR; 344446679JJQQKA22; 367899TTTTQKKAA22, L:33, D:66, U:KK, L:P,\\ D:P, U:6789T, L:P, D:P, U:3TTT, L:P, D:P, U:9, L:Q, D:K, U:P, L:A, D:2, U:P, L:P, D:3, U:Q, L:K, D:A,\\ U:P, L:B, D:P, U:P, L:6, D:7, U:A, L:R, D:P, U:P, L:555577JJ, D:P, U:P, L:8889} \\
     \bottomrule 
    \end{tabular}

\end{table}

\begin{table}[H]
    \centering
    \caption{Case 3: Comparison of $\DAI$ in early stage and later stage. $\DAI$ plays the Landlord position on the same deck with the same opponents. In the early stage, the first move of $\DAI$ is 9. However there is a 4 in the hand, which causes troubles in later turns. In the later stage, $\DAI$ plays in a different style by starting with 33 and finally wins the game. Although playing 9 seems to be not bad, it may lead to losing the game later.}
    \begin{tabular}{l|l}
    \toprule
     & Logs  \\
    \midrule
     Early stage & \makecell[l]{H:33455556889TTJAA22BR; 334677778TTJQKA22; 446688999JJQQQKKKA, L:9, D:T, U:A, L:2,\\ D:7777, U:P, L:P, D:TJQKA, U:P, L:5555, D:P, U:P, L:6, D:8, U:P, L:J, D:2, U:P, L:R, D:P, U:P, L:88,\\ D:P, U:JJ, L:AA, D:P, U:P, L:33, D:P, U:66, L:TT, D:P, U:KK, L:P, D:P, U:8999, L:P, D:P, U:QQQK, L:P,\\ D:P, U:44}  \\
     \midrule
     Later stage & \makecell[l]{H:33455556889TTJAA22BR; 334677778TTJQKA22; 446688999JJQQQKKKA, L:33, D:22, U:P, L:BR,\\ D:P, U:P, L:6, D:8, U:A, L:2, D:7777, U:P, L:P, D:TJQKA, U:P, L:5555, D:P, U:P, L:9, D:T, U:P, L:J, D:P,\\ U:K, L:2, D:P, U:P, L:88, D:P, U:JJ, L:P, D:P, U:44999, L:P, D:P, U:66QQQ, L:P, D:P, U:KK, L:AA, D:P, U:P,\\ L:TT, D:P, U:P, L:4} \\
     \bottomrule 
    \end{tabular}

\end{table}

\begin{table}[H]
    \centering
    \caption{Case 4: Comparison of $\DAI$ in early stage and later stage. $\DAI$ plays the LandlordUp and LandlordDown positions on the same deck with the same opponent. While $\DAI$ wins both games. $\DAI$ in the later stage looks more reasonable. In turn 16, $\DAI$ in the early stage chooses to break a pair of 2, which is hard to be explained. $\DAI$ in the later stage tends to play more smoothly and wins the games quickly.}
    \begin{tabular}{l|l}
    \toprule
     & Logs  \\
    \midrule
     Early stage & \makecell[l]{H:3455567788TJJQKKKKA2; 33446678899JQQQA2; 3456799TTTJAA22BR, L:345678, D:P, U:P,\\ L:TJQKA, D:P, U:P, L:55KKK, D:P, U:P, L:7, D:P, U:R, L:P, D:P, U:2, L:P, D:P, U:34567, L:P, D:P, U:J,\\ L:2, D:P, U:B, L:P, D:P, U:2, L:P, D:P, U:AA, L:P, D:P, U:99TTT}  \\
     \midrule
     Later stage & \makecell[l]{H:3455567788TJJQKKKKA2; 33446678899JQQQA2; 3456799TTTJAA22BR, L:345678, D:P, U:P,\\ L:TJQKA, D:P, U:P, L:55KKK, D:P, U:BR, L:P, D:P, U:99TTT, L:P, D:P, U:34567, L:P, D:P,\\ U:AA, L:P, D:P, U:22, L:P, D:P, U:J} \\
     \bottomrule 
    \end{tabular}

\end{table}

\newpage

\subsection{Comparison of Using WP and ADP as Objectives}
\label{sec:f4}

\begin{figure}[H]
  \centering

  \begin{subfigure}[b]{0.5\textwidth}
    \centering
    \includegraphics[width=0.95\textwidth]{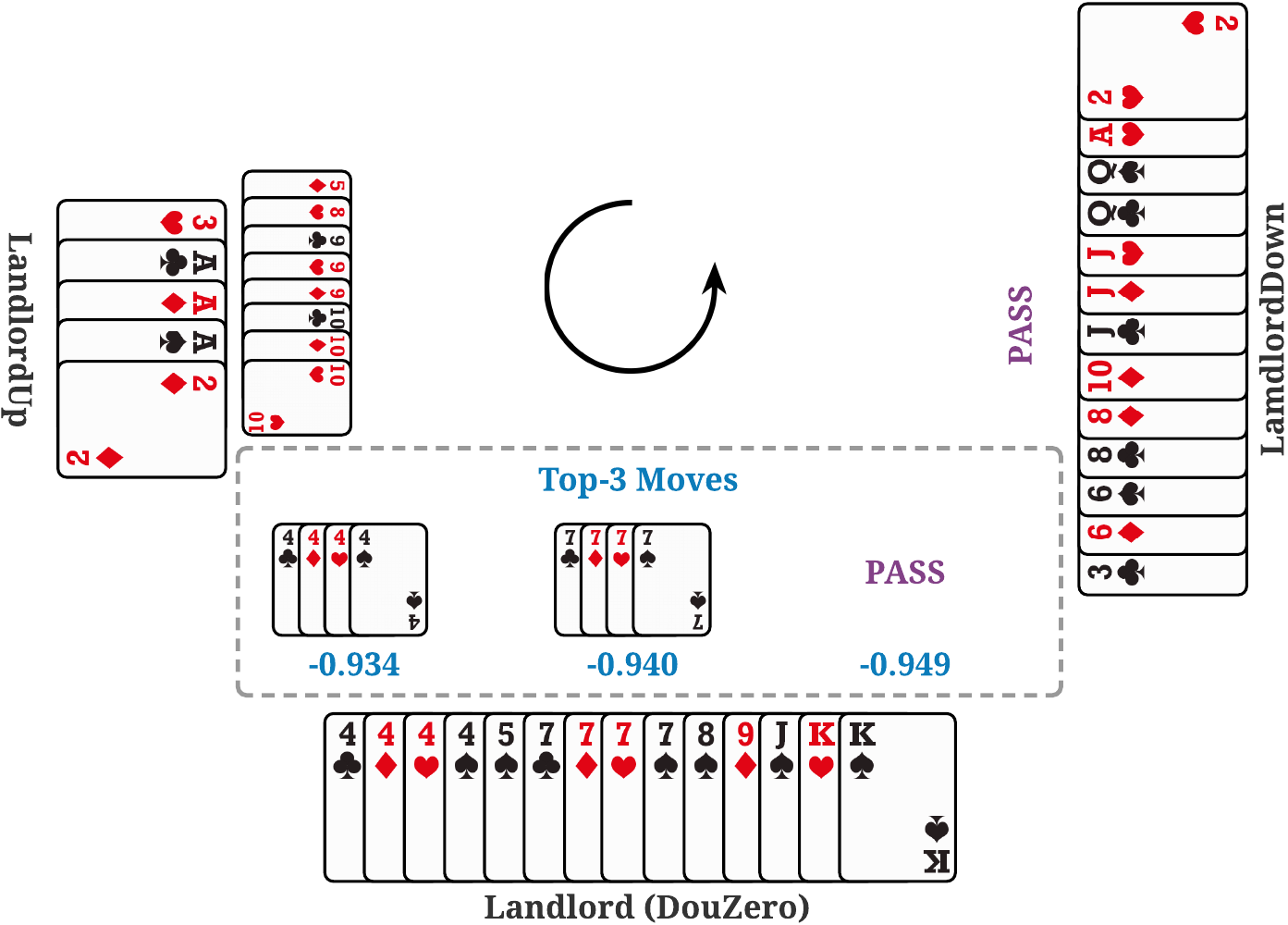}
    \caption{WP (turn 13)}
  \end{subfigure}%
  \begin{subfigure}[b]{0.5\textwidth}
    \centering
    \includegraphics[width=0.95\textwidth]{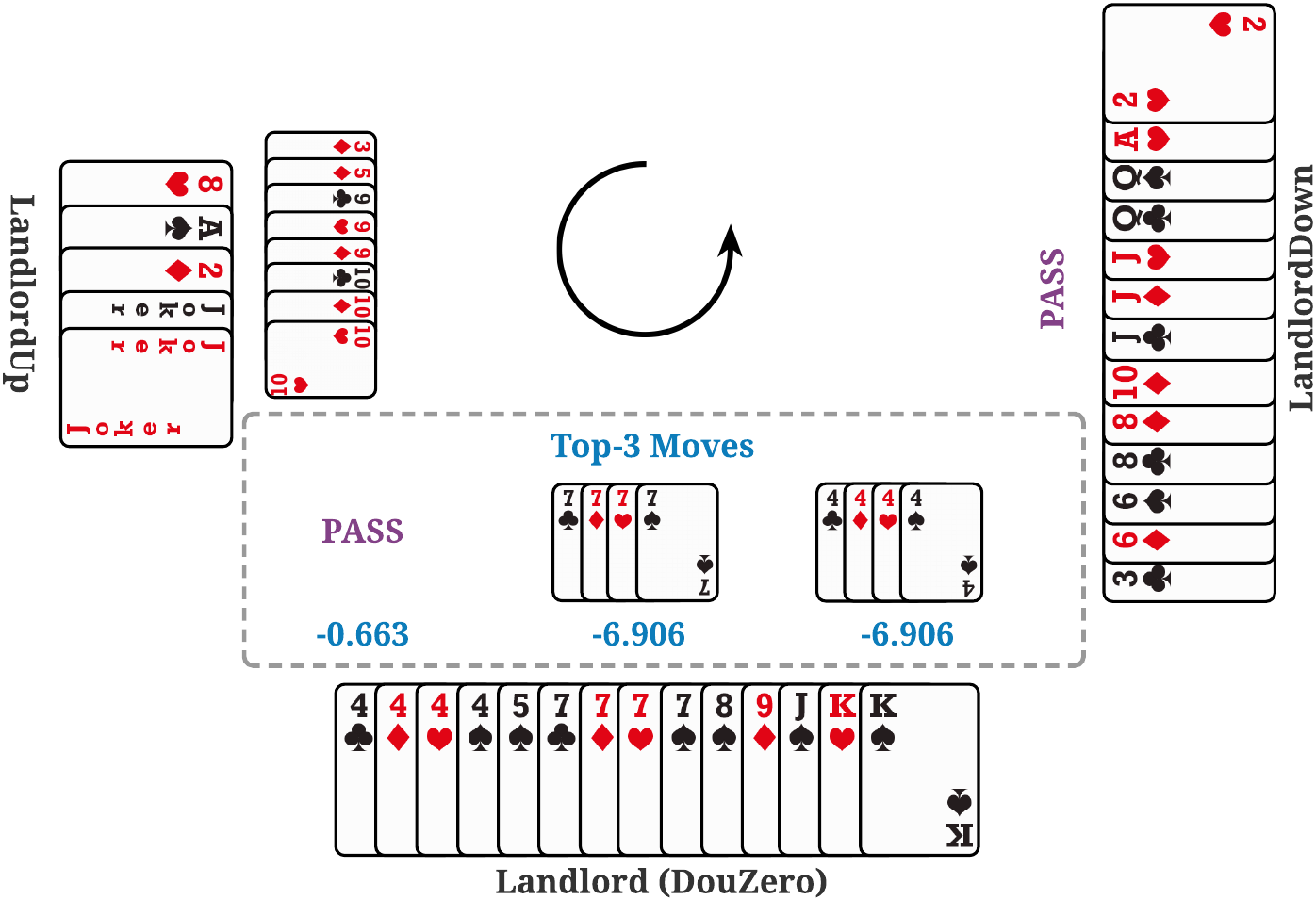}
    \caption{ADP (turn 10)}
  \end{subfigure}%
  \caption{Case 1: Comparison of using WP and ADP as objectives. The two agents play the same deck twice. It is difficult for the Landlord to win this game since the LandlordUp has a very good hand. When LandlordUp plays a Plane with Solo, WP agent tends to ambitiously play bombs because playing bombs has no cost. However, ADP agent tends to be very cautious of playing bombs since it may lead to a larger loss of ADP. \textbf{Full game of (a):} H:3344445666689JQQKK22; 3556688TJJJQQKKA2; 35668999TTTAAA2BR, L:33, D:55, U:66, L:QQ, D:KK, U:P, L:22, D:P, U:BR, L:P, D:P, U:58999TTT, L:7777, D:P, U:P, L:8, D:T, U:2, L:P, D:P, U:3AAA. \textbf{Full game of (b):} H:3344445666689JQQKK22; 3556688TJJJQQKKA2; 35668999TTTAAA2BR, L:33, D:55, U:66, L:KK, D:P, U:AA, L:P, D:P, U:35999TTT, L:P, D:P, U:8, L:J, D:A, U:2, L:P, D:P, U:BR, L:P, D:P, U:A. }
\end{figure}

\begin{figure}[H]
  \centering

  \begin{subfigure}[b]{0.5\textwidth}
    \centering
    \includegraphics[width=0.95\textwidth]{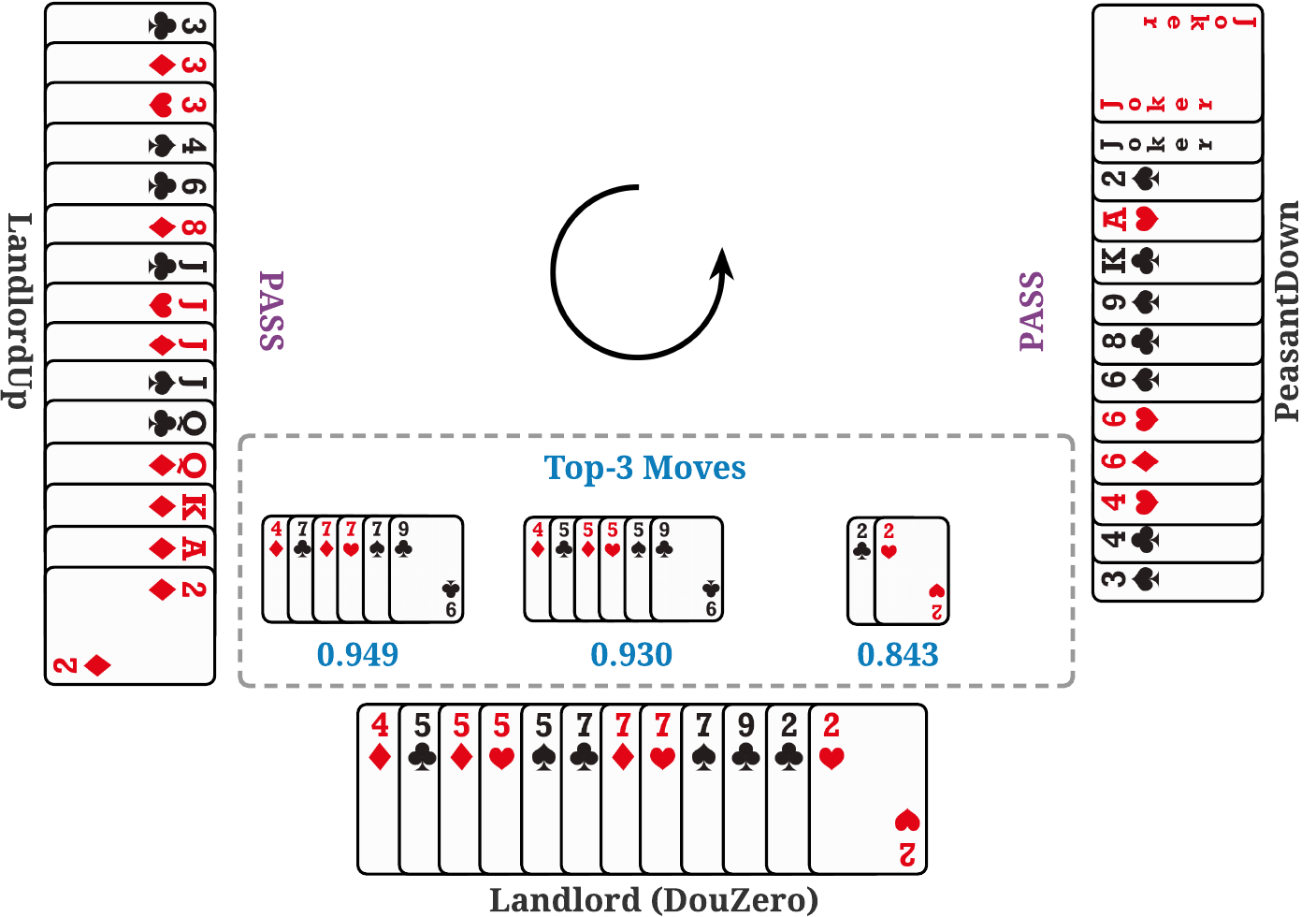}
    \caption{WP (turn 13)}
  \end{subfigure}%
  \begin{subfigure}[b]{0.5\textwidth}
    \centering
    \includegraphics[width=0.95\textwidth]{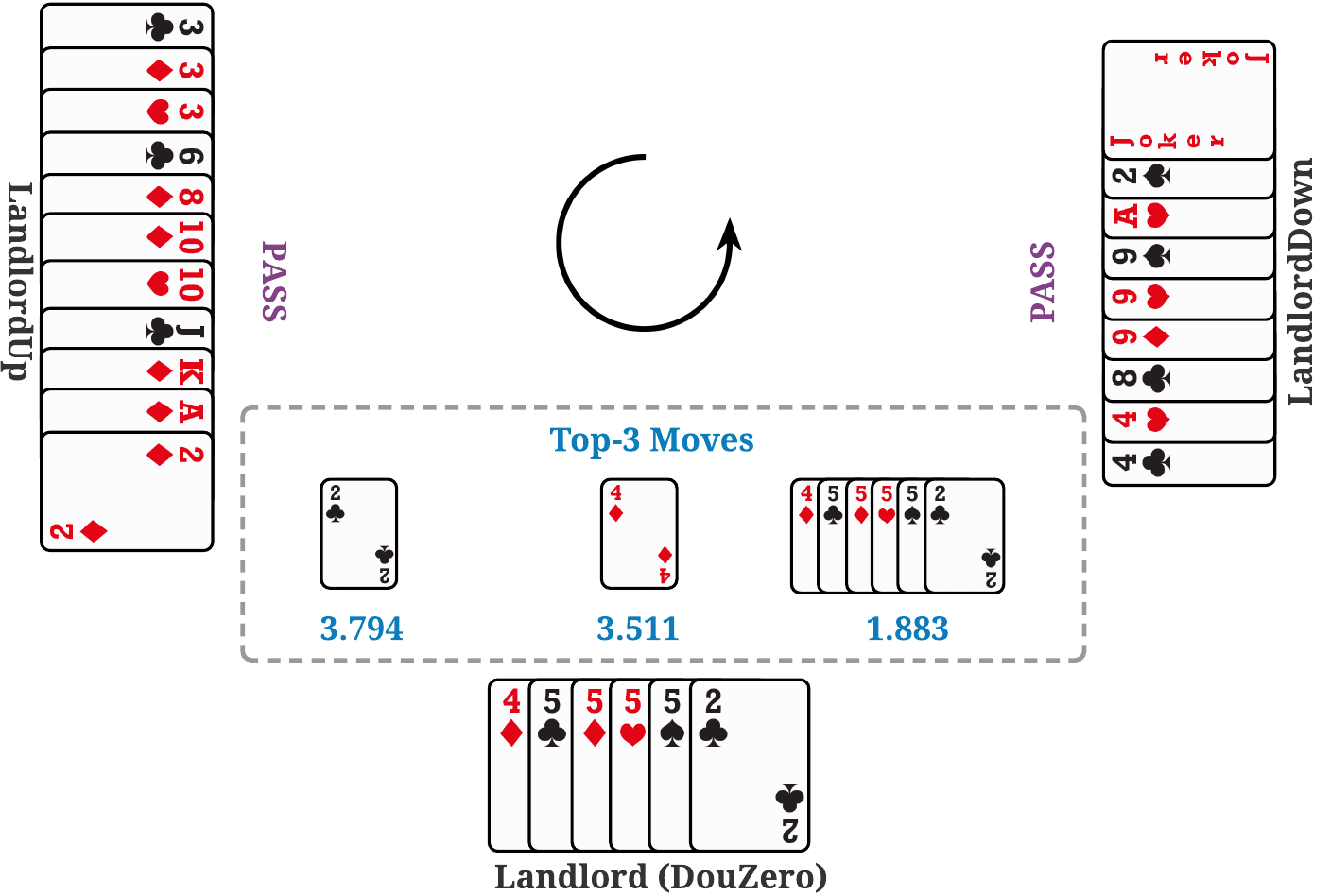}
    \caption{ADP (turn 25)}
  \end{subfigure}%
  \caption{Case 2: Comparison of using WP and ADP as objectives. The two agents play the same deck twice. While both agents win this game, they have different styles. The WP agent plays Quad with Solo instead of a bomb because it can empty the hand more quickly. This is reasonable since playing one more bomb will not double WP. In contrast, the ADP agent first plays a 2 so that it can play a bomb later. The ADP agent will try to play every bomb when it thinks it can win the game. \textbf{Full game of (a):} H:455557777889TTKKAA22; 3446668999QQKA2BR; 333468TTJJJJQQKA2, L:TT, D:QQ, U:P, L:KK, D:P, U:P, L:88, D:99, U:TT, L:AA, D:P, U:P, L:477779, D:P, U:46JJJJ, L:P, D:P, U:3338, L:5555, D:BR, U:P, L:P, D:3666, U:P, L:P, D:8, U:K, L:2, D:P, U:P, L:2. \textbf{Full game of (b):} H:455557777889TTKKAA22; 3446668999QQKA2BR; 333468TTJJJJQQKA2, L:88, D:QQ, U:P, L:KK, D:P, U:P, L:TT, D:P, U:QQ, L:AA, D:P, U:P, L:9, D:K, U:P, L:2, D:B, U:P, L:P, D:3666, U:4JJJ, L:7777, D:P, U:P, L:2, D:R, U:P, L:5555, D:P, U:P, L:4.}
\end{figure}

\newpage

\begin{figure}[H]
  \centering

  \begin{subfigure}[b]{0.5\textwidth}
    \centering
    \includegraphics[width=0.95\textwidth]{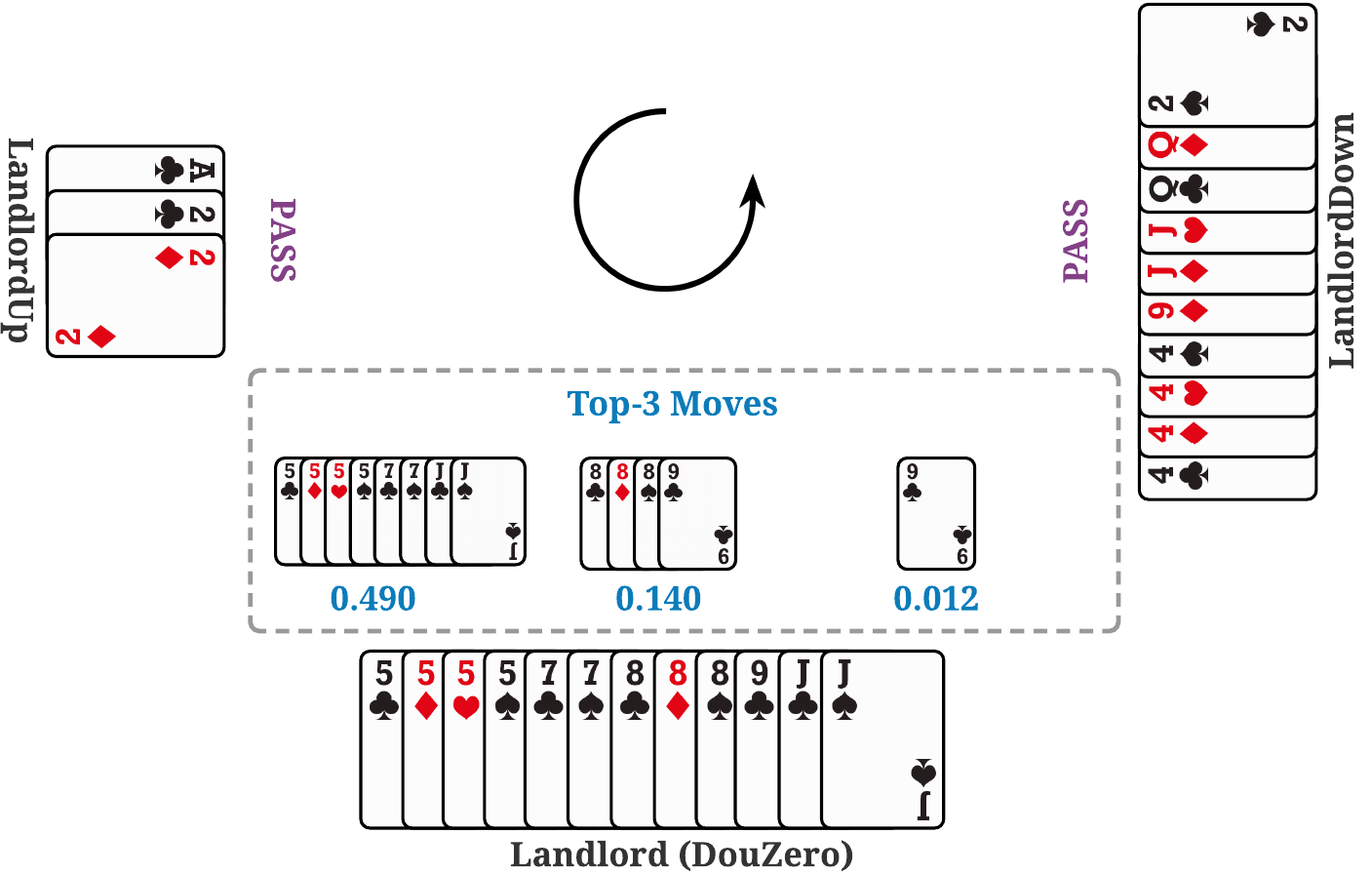}
    \caption{WP (turn 34)}
  \end{subfigure}%
  \begin{subfigure}[b]{0.5\textwidth}
    \centering
    \includegraphics[width=0.95\textwidth]{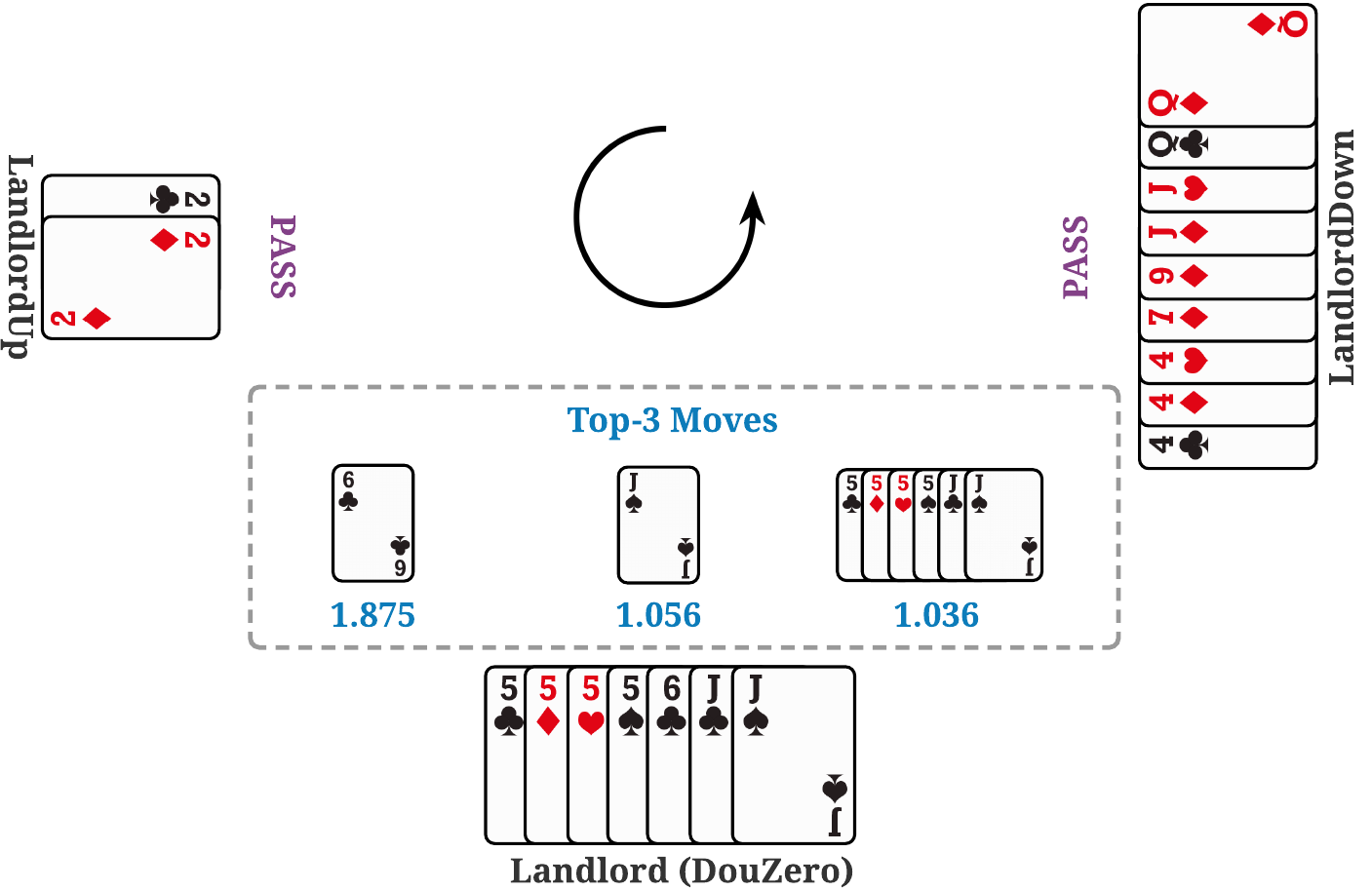}
    \caption{ADP (turn 43)}
  \end{subfigure}%
  \caption{Case 3: Comparison of using WP and ADP as objectives. The two agents play the same deck twice. In this game, the Landlord is very likely to win. The WP agent chooses a more conservative move by playing Quad with Pair to quickly empty the hand. In contrast, the ADP agent plays a small Solo first so that it can play 5555 later to double the ADP. \textbf{Full game of (a):} H:3355556778889JJQKABR; 344446679JJQQKA22; 367899TTTTQKKAA22, L:33, D:66, U:KK, L:P, D:P, U:6789T, L:P, D:P, U:3TTT, L:P, D:P, U:9, L:Q, D:K, U:P, L:A, D:2, U:P, L:P, D:3, U:Q, L:K, D:A, U:P, L:B, D:P, U:P, L:6, D:7, U:A, L:R, D:P, U:P, L:555577JJ, D:P, U:P, L:8889. \textbf{Full game of (b):} H:3355556778889JJQKABR; 344446679JJQQKA22; 367899TTTTQKKAA22, L:33, D:66, U:KK, L:P, D:P, U:6789T, L:P, D:P, U:3TTT, L:P, D:P, U:9, L:Q, D:K, U:P, L:A, D:2, U:P, L:P, D:3, U:Q, L:K, D:A, U:P, L:P, D:4, U:A, L:P, D:P, U:A, L:R, D:P, U:P, L:9, D:2, U:P, L:B, D:P, U:P, L:77888, D:P, U:P, L:6, D:7, U:2, L:5555, D:P, U:P, L:JJ.}
\end{figure}

\subsection{Bad Cases}

\begin{figure}[H]
  \centering
    \includegraphics[width=0.5\textwidth]{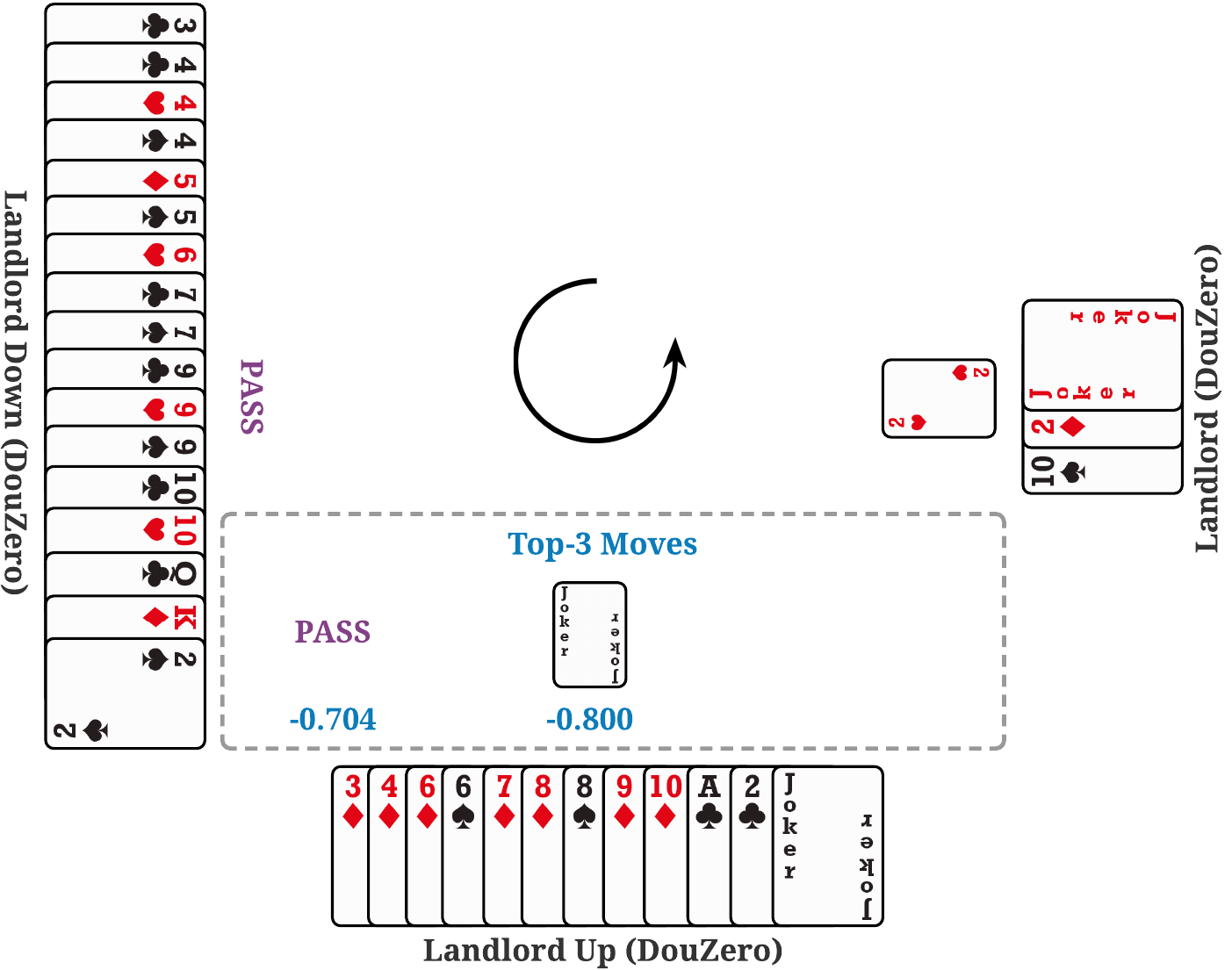}
  \caption{Case 1: Bad case (turn 21). The Landlord plays a 2 with only 3 cards left. While the LandlordUp has Black Joker in hand, $\DAI$ chooses not to play it. Although the Peasants will lose whatever the LandlordUp plays in this specific case, playing the Black Joker should have a larger chance to win (if the Red Joker happens to be in the hand of the LandlordDown). Thus, it is worth a try to play Black Joker. \textbf{Full game:} H: 3556788TQQQKKKAAA22R; 344455677999TTQK2; 334667889TJJJJA2B; L:55, D:P, U:P, L:88, D:P, U:P, L:3AAA, D:P, U:P, L:7KKK, D:P, U:P, L:6QQQ, D:P, U:JJJJ, L:P, D:P, U:3, L:2, D:P, U:P, L:T, D:K, U:A, L:2, D:P, U:B, L:R.}
\end{figure}

\newpage

\begin{figure}[H]
  \centering
    \includegraphics[width=0.5\textwidth]{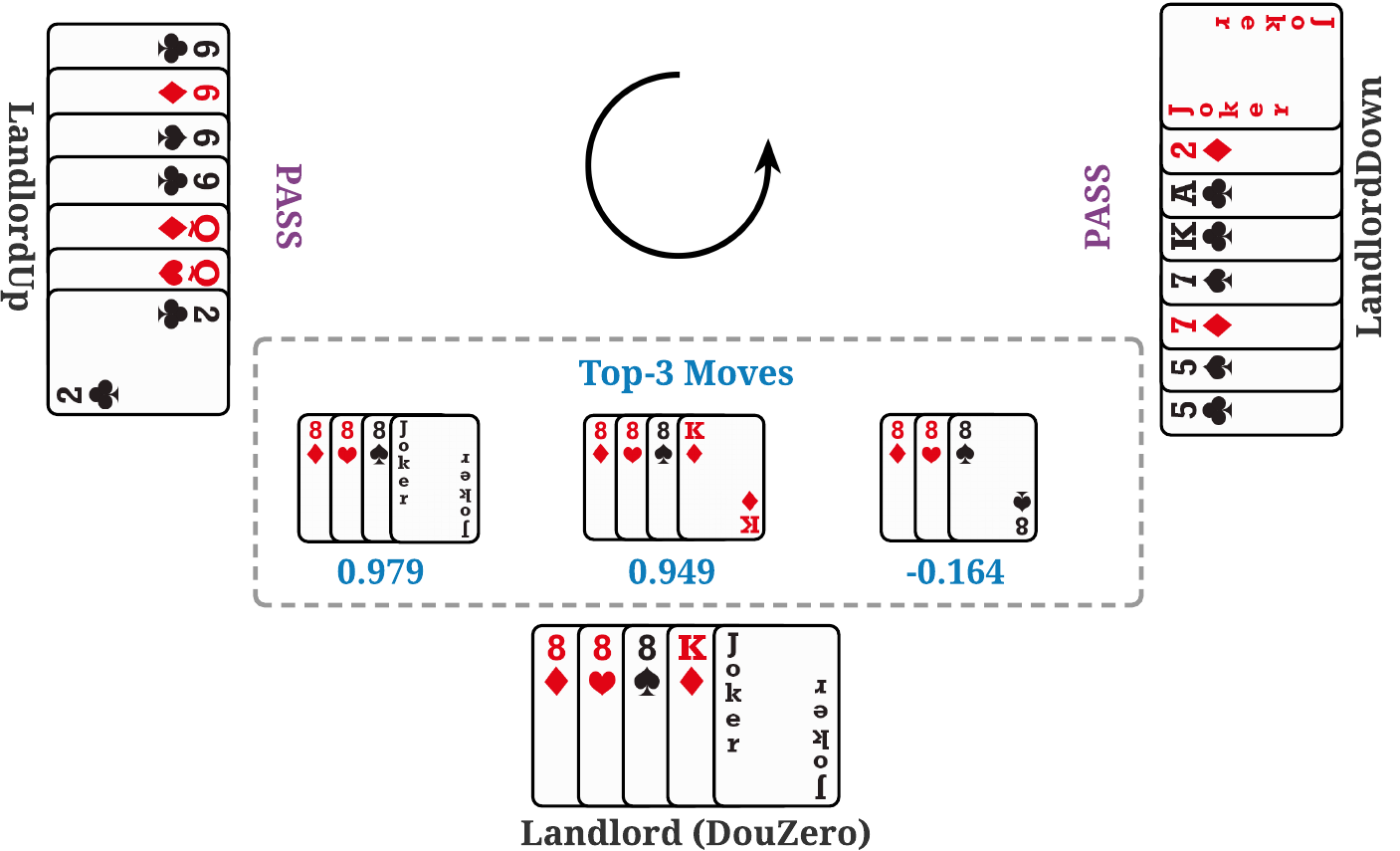}
  \caption{Case 2: Bad case (turn 22). This could be a bad case. $\DAI$ aggressively chooses Black Joker as the kicker instead of K. While it is true that $\DAI$ can win whichever kicker it chooses, choosing Black Joker is risky because this knowledge could be generalized to other cases with neural networks and results in losing a game. In fact, choosing K as the kicker will always be at least as good as Black Joker given the current hand. \textbf{Full game:} H:333578889TJQKKAAA22B; 444455779JJJQKA2R; 3566667899TTTQQK2, L:3335, D:9JJJ, U:P, L:7AAA, D:4444, U:P, L:P, D:Q, U:K, L:P, D:P, U:3TTT, L:P, D:P, U:56789, L:9TJQK, D:P, U:P, L:22, D:P, U:P, L:888B, D:P, U:P, L:K.}
\end{figure}

\newpage

\subsection{Randomly Selected (rather than Cherry-Picked) Full Games}
\begin{table}[H]
    \centering
    \caption{Case 1: Randomly selected (rather than cherry-picked) full self-play games of $\DAI$. $\DAI$ plays the Landlord, LandlordUP and LandlordDown positions. All $\DAI$ agents were trained using WP as objectives. }
    \begin{tabular}{l}
    \toprule
    Logs  \\
      \midrule 
 \makecell[l]{H:335556788899TTJKKA2R; 6677789TTJJQQKA2B; 334444569JQQKAA22, L:33, D:66, U:QQ, L:KK,\\ D:P, U:AA, L:P, D:P, U:9, L:T, D:A, U:P, L:2, D:B, U:P, L:P, D:9, U:J, L:A, D:2, U:P, L:R, D:P, U:P, L:5559,\\ D:777J, U:P, L:P, D:J, U:K, L:P, D:P, U:5, L:6, D:8, U:4444, L:P, D:P, U:33, L:88, D:TT, U:22, L:P, D:P, U:6}  \\ 
 \midrule 
 \makecell[l]{H:34455777889JQQKA222B; 345668899TTTJQKAA; 33456679TJJQKKA2R, L:44, D:P, U:JJ, L:QQ,\\ D:AA, U:P, L:P, D:3, U:T, L:J, D:P, U:Q, L:K, D:P, U:A, L:P, D:P, U:9, L:A, D:P, U:2, L:B, D:P, U:P, L:3,\\ D:4, U:6, L:9, D:Q, U:P, L:2, D:P, U:R, L:P, D:P, U:34567, L:P, D:P, U:KK, L:22, D:P, U:P, L:77788, D:66TTT, U:P, L:P,\\ D:J, U:P, L:P, D:9, U:P, L:P, D:9, U:P, L:P, D:8, U:P, L:P, D:5, U:P, L:P, D:K, U:P, L:P, D:8}  \\ 
 \midrule 
 \makecell[l]{H:34455666789TTJQQKA2R; 335688899TJQA222B; 34457779TJJQKKKAA, L:445566, D:P, U:P, L:6789T,\\ D:89TJQ, U:9TJQK, L:TJQKA, D:P, U:P, L:Q, D:A, U:P, L:2, D:B, U:P, L:R, D:P, U:P, L:3}  \\ 
 \midrule 
 \makecell[l]{H:334447779TTTJQKKK22R; 3345566678TJQKA2B; 556888999JJQQAAA2, L:33444, D:55666, U:888JJ, L:TTT22,\\ D:P, U:55AAA, L:P, D:P, U:QQ, L:P, D:P, U:6999, L:QKKK, D:P, U:P, L:R, D:P, U:P, L:777J, D:P, U:P, L:9}  \\ 
 \midrule 
 \makecell[l]{H:455577899TTJJKKAA22B; 3346789TJJQQAA22R; 334456667889TQQKK, L:4, D:P, U:K, L:A,\\ D:2, U:P, L:P, D:6789TJ, U:P, L:P, D:4, U:Q, L:A, D:2, U:P, L:B, D:R, U:P, L:P, D:33, U:88, L:TT, D:QQ, U:P, L:KK,\\ D:AA, U:P, L:22, D:P, U:P, L:JJ, D:P, U:P, L:77, D:P, U:P, L:99, D:P, U:P, L:5558}  \\ 
 \midrule 
 \makecell[l]{H:3345666889JQQQKAA22B; 4457777899TJQAA2R; 33455689TTTJJKKK2, L:5, D:9, U:J, L:K,\\ D:2, U:P, L:B, D:R, U:P, L:P, D:89TJQ, U:P, L:P, D:44, U:P, L:88, D:AA, U:P, L:22, D:7777, U:P, L:P, D:5}  \\ 
 \midrule 
 \makecell[l]{H:334679999TTQQKKKAA2B; 3455677888JQKA22R; 344556678TTJJJQA2, L:33, D:55, U:TT, L:QQ,\\ D:P, U:P, L:TT, D:P, U:JJ, L:AA, D:22, U:P, L:P, D:4, U:Q, L:2, D:P, U:P, L:469999, D:P, U:P, L:7KKK, D:P, U:P, L:B}  \\ 
 \midrule 
 \makecell[l]{H:3455667899TTTJJQQKAB; 33567788JJQQKA22R; 3444567899TKKAA22, L:6, D:K, U:A, L:B,\\ D:P, U:P, L:34567, D:P, U:56789, L:89TJQ, D:P, U:P, L:5, D:6, U:A, L:P, D:P, U:T, L:A, D:2, U:P, L:P, D:5, U:9, L:J,\\ D:P, U:K, L:P, D:P, U:K, L:P, D:P, U:22, L:P, D:P, U:3444}  \\ 
 \midrule 
 \makecell[l]{H:345578TTJJJQKKKA2222; 3334566667789QQKR; 445788999TTJQAAAB, L:3, D:8, U:Q, L:2,\\ D:P, U:B, L:P, D:P, U:J, L:Q, D:K, U:P, L:A, D:P, U:P, L:4, D:9, U:T, L:P, D:P, U:88, L:TT, D:QQ, U:P, L:P,\\ D:4, U:9, L:2, D:6666, U:P, L:P, D:5, U:9, L:K, D:R, U:P, L:P, D:33377}  \\ 
 \midrule 
 \makecell[l]{H:3334577889TTJQKAAA2B; 345566789TJQQKKA2; 445667899TJJQK22R, L:5, D:6, U:K, L:P,\\ D:A, U:P, L:2, D:P, U:P, L:T, D:K, U:P, L:B, D:P, U:R, L:P, D:P, U:456789TJQ, L:P, D:P, U:4, L:K, D:2, U:P, L:P,\\ D:3, U:9, L:A, D:P, U:2, L:P, D:P, U:J, L:A, D:P, U:2, L:P, D:P, U:6}  \\ 
     \bottomrule 
    \end{tabular}
\end{table}

\newpage

\begin{table}[H]
    \centering
    \caption{Case 2: Randomly selected (rather than cherry-picked) full games of $\DAI$ played with other agents in Botzone. $\DAI$ plays the position as noted in the left column. All $\DAI$ agents were trained using WP as objectives. }
    \begin{tabular}{l|l}
    \toprule
    Position & Logs  \\
      \midrule 
Landlord & \makecell[l]{H:33455778999TTTQKKAA2; 34567788TJQKKA2BR; 344566689JJJQQA22, L:48999TTT, D:BR,\\ U:P, L:P, D:34567, U:P, L:P, D:TJQKA, U:P, L:P, D:88, U:QQ, L:KK, D:P, U:22, L:P, D:P, U:3666,\\ L:P, D:P, U:5JJJ, L:P, D:P, U:44, L:AA, D:P, U:P, L:2, D:P, U:P, L:77, D:P, U:P, L:33,\\ D:P, U:P, L:55, D:P, U:P, L:Q}  \\ 
 \midrule 
Peasants & \makecell[l]{H:34456779TTJKKKAAAA22; 34455668888JQQQ2R; 335677999TTJJQK2B, L:34567, D:P,\\ U:9TJQK, L:P, D:P, U:T, L:J, D:Q, U:P, L:A, D:2, U:P, L:P, D:445566, U:P, L:P, D:38888Q, U:P,\\ L:P, D:Q, U:P, L:P, D:R, U:P, L:P, D:J}  \\ 
 \midrule 
Landlord & \makecell[l]{H:334455668899TJQQAA2B; 34567777TJJKKKAA2; 34568899TTJQQK22R, L:33445566, D:P,\\ U:P, L:9, D:T, U:P, L:Q, D:A, U:P, L:P, D:777JJ, U:P, L:P, D:34567, U:P, L:89TJQ, D:P, U:9TJQK,\\ L:P, D:P, U:3, L:8, D:A, U:P, L:2, D:P, U:R, L:P, D:P, U:88, L:AA, D:P, U:22, L:P,\\ D:P, U:4, L:B}  \\ 
 \midrule 
Peasants & \makecell[l]{H:34455677889TTJJQKK2R; 33557899JQQQKKA22; 344666789TTJAAA2B, L:345678, D:P,\\ U:6789TJ, L:789TJQ, D:P, U:P, L:4, D:7, U:A, L:2, D:P, U:B, L:R, D:P, U:P, L:5, D:A, U:P,\\ L:P, D:JQQQ, U:P, L:P, D:55, U:P, L:KK, D:22, U:P, L:P, D:K, U:A, L:P, D:P, U:A, L:P,\\ D:P, U:3, L:T, D:P, U:2, L:P, D:P, U:66, L:P, D:P, U:44, L:P, D:P, U:T}  \\ 
 \midrule 
Peasants & \makecell[l]{H:3355667789TJJQA222BR; 334446788TQQKKAA2; 455678999TTJJQKKA, L:33, D:88,\\ U:TT, L:P, D:P, U:45678, L:89TJQ, D:P, U:P, L:556677, D:QQKKAA, U:P, L:P, D:4, U:A, L:2, D:P, U:P,\\ L:J, D:P, U:Q, L:A, D:P, U:P, L:2, D:P, U:P, L:2, D:P, U:P, L:BR}  \\ 
 \midrule 
Landlord & \makecell[l]{H:334566778899TTQQKA2B; 3455689TJJQKKAA22; 344567789TJJQKA2R, L:456789T, D:89TJQKA,\\ U:P, L:P, D:3, U:4, L:K, D:A, U:2, L:P, D:P, U:34567, L:6789T, D:P, U:P, L:33, D:55, U:P,\\ L:QQ, D:22, U:P, L:P, D:4, U:J, L:B, D:P, U:R, L:P, D:P, U:789TJQKA}  \\ 
 \midrule 
Landlord & \makecell[l]{H:33455668TTTJJJQKAA22; 3445677789QQKKA2R; 3456788999TJQKA2B, L:4, D:A,\\ U:P, L:P, D:4, U:9, L:Q, D:K, U:2, L:P, D:P, U:89TJQKA, L:P, D:P, U:B, L:P, D:P, U:3456789}  \\ 
     \bottomrule 
    \end{tabular}

\end{table}

\end{document}